\pdfoutput=1

\documentclass[opre,nonblindrev]{informs3mod}

\RequirePackage{tgtermes}
\RequirePackage{newtxtext}
\RequirePackage{newtxmath}
\RequirePackage{bm}

\usepackage{xurl}
\usepackage{graphicx}
\usepackage{amsmath}
\usepackage{amsfonts}
\usepackage{amssymb}
\usepackage{xcolor}
\definecolor{pumpkin}{HTML}{E67E22}
\definecolor{belize}{HTML}{2980b9}
\usepackage{algorithmicx}
\usepackage{algorithm}
\usepackage{algpseudocode}
\usepackage{subfigure}
\usepackage[suppress]{color-edits}
\usepackage{caption}
\usepackage{tikz-network}
\usepackage[hidelinks]{hyperref}
\usepackage[noabbrev, capitalise]{cleveref}
\usepackage{paralist}
\usepackage{booktabs}

\addauthor{mp}{red}
\addauthor{ar}{blue}

\newcommand{\one}{\bm 1}

\newcommand{\linktoproof}[1]{\begin{center} \hyperref[#1]{\texttt{[Link to Proof]}}\end{center}}
\newcommand{\linktostatement}[1]{\begin{center} \hyperref[#1]{\texttt{[Link to Statement]}}\end{center}}
\newcommand{\ev}[2]{\mathbb E_{#1}\left [ #2 \right ]}

\renewcommand{\Pr}{\mathbb P}
\newcommand{\var}[2]{\mathbb V_{#1} \left [ #2 \right ]}

\newcommand{\cA}{\mathcal{A}}

\newcommand{\cD}{\mathcal{D}}
\newcommand{\cE}{\mathcal{E}}

\newcommand{\cG}{\mathcal{G}}

\newcommand{\cM}{\mathcal{M}}
\newcommand{\cN}{\mathcal{N}}
\newcommand{\cO}{\mathcal{O}}

\newcommand{\cR}{\mathcal{R}}
\newcommand{\cS}{\mathcal{S}}
\newcommand{\cT}{\mathcal{T}}
\newcommand{\cU}{\mathcal{U}}

\newcommand{\Nbb}{\mathbb{N}}

\newcommand{\Rbb}{\mathbb{R}}

\newcommand{\trt}{\mathrm{trt}}
\newcommand{\ctrl}{\mathrm{ctrl}}
\newcommand{\compTheta}[0]{\bar {\Theta}}
\newcommand{\kldiv}[2]{D_{KL} \left ( #1 | #2 \right )}

\renewcommand{\Pr}{\mathbb P}
\newcommand{\ovec}[1]{\overline {\vec #1}}
\newcommand{\Otheta}[1]{\ensuremath {\cO_{n, \eps, \eta} \left ( #1 \right )}}
\newcommand{\Onn}[1]{\ensuremath{\cO_{|\Theta|, \eps, \eta} \left ( #1 \right )}}
\newcommand{\Opriv}[1]{\ensuremath{\cO_{|\Theta|, n} \left ( #1 \right )}}
\newcommand{\thres}[0]{\mathrm{thres}}
\newcommand{\tauthres}[1]{\tau^{\thres, #1}}
\newcommand{\hattauthres}[1]{{\hat \tau}^{\thres, #1}}

\def\eps{{\varepsilon}}
\renewcommand{\vec}{\bm}
\newcommand{\AM}[0]{\mathrm{AM}}
\newcommand{\GM}[0]{\mathrm{GM}}
\newcommand{\intt}[0]{\mathrm{OL}}

\renewcommand{\var}[2]{\mathbb V_{#1} \left [ #2 \right ]}
\renewcommand{\Pr}{\mathbb P}
\newcommand{\pfsketch}[1]{}

\OneAndAHalfSpacedXII % Current default line spacing
\makeatletter
\renewcommand\Dskips[4]{%
  \abovedisplayskip 0.5em \@plus 0.2em \@minus 0.2em
  \abovedisplayshortskip \z@ \@plus 0.2em
  \belowdisplayshortskip 0.3em \@plus 0.2em \@minus 0.2em
  \belowdisplayskip \abovedisplayskip}%
\makeatother
\normalsize

\usepackage{tikz}
\usepackage{natbib}
 \bibpunct[, ]{(}{)}{,}{a}{}{,}%
 \def\bibfont{\small}%
\EquationsNumberedThrough    % Default: (1), (2), ...
\TheoremsNumberedThrough     % Preferred (Theorem 1, Lemma 1, Theorem 2)
\MANUSCRIPTNO{}

\begin{document}
%%%%%%%%%%%%%%%%

%\setcounter{page}{1}
%\DOI{10.1287/xxxx.0000.0000}%

\RUNAUTHOR{Papachristou and Rahimian}

% Title or shortened title suitable for running heads.
\RUNTITLE{Differentially Private Distributed Inference for Multicenter Clinical Studies}

\TITLE{Differentially Private Distributed Inference for Multicenter Clinical Studies}

\ARTICLEAUTHORS{%
\AUTHOR{Marios Papachristou}
\AFF{W. P. Carey School of Business,
Arizona State University, \EMAIL{mpapachr@asu.edu}}

\AUTHOR{M. Amin Rahimian}
\AFF{Department of Industrial Engineering,
University of Pittsburgh, \EMAIL{rahimian@pitt.edu}}
} % end of the block

\ABSTRACT{%
Extracting reliable conclusions from data distributed across institutions is a core problem in healthcare, as pooling patient records across centers would improve inference; however, privacy regulations and
the lack of a trusted central authority pose significant challenges that frequently delay or prevent multicenter studies. 
In this work, we develop a framework for differentially private distributed
inference in which institutions repeatedly exchange log belief-ratio statistics with other participating institutions, subject to differential privacy (DP). We show that with arithmetic and geometric averaging of beliefs, we can control the false-negative and false-positive rates as functions of the privacy budget, communication rounds, and statistical separation between the hypotheses; exposing a three-way trade-off among accuracy, communication, and privacy. We derive finite-sample bounds on the Type I and Type II error probabilities that allow us to perform distributed hypothesis tests at a target significance level and show that the Laplace mechanism minimizes convergence time subject to differential privacy guarantees. For distributed online learning from data streams (e.g., in epidemiological surveillance or in studies with rolling recruitment), we show that privacy noise vanishes asymptotically and online learning admits similar finite-sample convergence guarantees. On simulated multicenter survival analyses on data from the AIDS Clinical Trials Group and an advanced-cancer cohort, and on a simulated genetic association study over New York City hospitals, our method approaches the non-private baseline at a small privacy budget between $1$ and $10$, and it runs $10\times$ to $1000\times$ faster than homomorphic-encryption methods, and incurs up to $100\times$ lower error than first-order private convex optimization methods. Finally, the level of aggregation is a primary design choice: federating at the organizational rather than the hospital level simultaneously strengthens privacy, raises statistical power, and lowers communication and administrative burden, so the priority should be to pool data within organizations before setting up federated analytics frameworks.

}%

%\HISTORY{This version: \today}

\maketitle
%%%%%%%%%%%%%%%%%%%%%%%%%%%%%%%%%%%%%%%%%%%%%%%%%%%%%%%%%%%%%%%%%%%%%%

% Text of the paper here (shared with ijds.tex)
\section{Introduction} \label{sec:introduction}

Clinical trials represent the gold standard for generating medical evidence and depend on recruiting a sufficient number of patients in order to perform valid inferences. However, patient recruitment, both in terms of the total number of patients as well as recruiting patients from underrepresented minorities, faces significant challenges. One of the most challenging obstacles that results in centers having lower sample sizes or samples that are not representative of the general population is patient privacy and regulatory compliance. The privacy risk is not hypothetical, even when only aggregate statistics are shared:  from published allele frequencies alone, membership-inference attacks can determine whether a given individual contributed to a study \citep{homer2008resolving}; a finding that prompted the withdrawal of summary genomic data from open access and continues to impede collaboration \citep{zerhouni2008protecting,erlich2014routes}.

A potential solution to circumvent this problem is establishing multicenter studies where several participating institutions opt in to share their data, subject to data usage agreements (DUAs). Centralized trial networks such as the AIDS Clinical Trials Group or the UK Biobank have demonstrated the value of pooled multicenter recruitment \citep{hammer1996trial, campbell2012efficacy, sudlow2015uk}, but they depend on coordination infrastructures that are beyond the operational capacity of many institutions, particularly smaller centers whose patient populations are often the most clinically relevant and demographically diverse. 

Geographically separated institutions often hold patient data that, if jointly analyzed, would yield statistically reliable conclusions about treatment efficacy, prognostic factors, genetic associations, and disease mechanisms. Performing valid inferences collectively by pooling such data would therefore be of great value, were it not for the strict privacy regulations that constrain it. The downstream consequences of these regulations, namely HIPAA (1996) in the United States and GDPR (2018) in the European Union, are well-documented for multicenter collaborations \citep{ness2007influence, iom2009beyond,kao2026data,clarke2019gdpr}: such collaborations are routinely delayed or abandoned due to regulatory barriers, hindering discovery and innovation. %The risk is not hypothetical even when only aggregate statistics are shared: membership-inference attacks can determine whether an individual contributed to a study from published allele frequencies alone \citep{homer2008resolving}, a finding that led to the withdrawal of summary genomic data from open access and continues to impede collaboration \citep{zerhouni2008protecting,erlich2014routes}.

\noindent These barriers motivate the central question that this paper addresses:

\begin{quote}
\emph{How can geographically distributed institutions jointly analyze sensitive patient data to support reliable statistical inference, without any institution relinquishing control of its own records or violating the confidentiality obligations it owes to its patients?}
\end{quote}

\noindent We answer this question by developing a framework for differentially private distributed inference in which multiple centers exchange privacy-protected beliefs rather than patient records and perform social learning \citep{bullo2009distributed,Jackson2008,GolubSadler,rahimian2016distributed}. Differential privacy (DP) provides the formal guarantee: a mechanism satisfying $\varepsilon$-DP ensures that the probability of any observable output changes by at most a factor of $e^\varepsilon$ when any single patient's data is added or removed, providing each patient with a quantifiable and formally verifiable form of plausible deniability \citep{dwork2014algorithmic}. This guarantee is not merely contractual but mathematical, and it is auditable by regulators and patients alike \citep{dwork2019differential}. By incorporating DP into iterative belief exchange protocols, our framework enables institutions to collectively reach statistically valid conclusions, such as performing hypothesis tests at a prescribed significance level, while mathematically preserving privacy and without the need for a centralized coordinating authority or specialized cryptographic infrastructure.

The practical relevance of this approach is underscored by recent policy developments. The U.S.\ Department of Health and Human Services' Advancing 
Clinical Trial Readiness initiative, launched by \citet{arpaH_ACTR2024}, seeks to establish a decentralized, on-demand clinical trial infrastructure accessible to $90\%$ of eligible participants within thirty minutes of their home, 
automating data extraction and synchronization across electronic health records at distributed sites. Our framework addresses the core statistical challenge such an infrastructure must solve: how to extract reliable collective inference from distributed, privacy-constrained institutional data without centralized 
access to patient records.

% \subsection{Main Contributions}

\noindent \textbf{Main Contributions.} We propose a differentially private distributed inference framework for 
information aggregation among a network of $n$ institutions, each holding 
private patient data and engaging in iterative belief exchange to collectively 
identify the most likely alternatives (e.g., to choose from possible diagnoses or treatment options), test a 
hypothesis at a prescribed significance level (e.g., in genetic association and survival analyses), or learn the true state from 
a continuous stream of observations (e.g., for epidemiological surveillance). Privacy is enforced through calibrated 
Laplace noise injection into the belief statistics exchanged between 
neighboring institutions, ensuring that no institution's communications 
reveal individual patient records.

We establish explicit finite-sample bounds on Type-I and Type-II error probabilities as functions of the privacy budget $\eps$, the number of communication rounds $K$, the number of iterations per round $T$, and the statistical divergence between competing hypotheses, showing a three-way tradeoff among inference accuracy, communication complexity, and privacy level. To manage this tradeoff in practice, we develop two complementary aggregation strategies: arithmetic averaging, which controls 
Type-II error and is suited to screening contexts where missed detections carry the greater clinical cost, and geometric averaging, which controls 
Type-I error and is suited to regulatory contexts where false positives carry legal and clinical liability. A two-threshold algorithm provides 
simultaneous control over both error types at a modest increase in communication complexity. These guarantees extend naturally to formal hypothesis testing at a prescribed significance level $\alpha$, and accommodate both simple and composite hypotheses including the generalized likelihood ratio tests used in survival analysis. For prospective multicenter trials and ongoing epidemiological surveillance, we further extend the framework to continuous observation streams, establishing that privacy noise vanishes asymptotically and providing 
finite-sample convergence guarantees characterizing the communication cost of privacy. In both the finite-sample and online settings, the Laplace mechanism is shown to be optimal for minimizing convergence time.

The most closely related prior work is
\citet{papachristou2023differentially}, which incorporates differential privacy into the distributed learning framework of  \citet{rahimian2016distributed} for continuous parameter estimation, providing convergence guarantees in expectation on the $\ell_2$ norm of
the estimation error. Our setting differs along three fundamental dimensions: inference in discrete parameter spaces introduces a failure probability bounded away from zero as iterations grow, requiring institutions to repeat deliberations across multiple rounds and aggregate results under an entirely new analytical framework; our finite-sample
guarantees are expressed as explicit Type-I and Type-II error bounds rather than the expected $\ell_2$ error, the latter being insufficient to
support hypothesis testing at a prescribed significance level; and the clinical applications we develop, multicenter survival analysis
and differentially private genetic association analysis, are not addressed by \citet{papachristou2023differentially} and require
problem-specific sensitivity analyses and algorithmic adaptations. The optimality
of the Laplace mechanism for minimizing convergence time, which also arises in the continuous-estimation setting of \citet{papachristou2023differentially}, thus persists across two structurally distinct inference settings, providing independent validation of this design choice. 

The complementary works of \citet{rizk2023enforcing} and \citet{cyffers2024differentially} provide first-order methods for distributed optimization and learning subject to DP. \citet{cyffers2024differentially} analyzes the random-walk first-order stochastic gradient descent method under pairwise network DP. Additionally, \citet{rizk2023enforcing} provides consensus algorithms and introduces the graph-homomorphic noise model to achieve DP. We note that both \citet{cyffers2024differentially} and \citet{rizk2023enforcing}, provide convergence guarantees to the optimum compared to our method, which gives Type-I and Type-II error guarantees. Later in the paper, we show that our method achieves significantly lower error than first-order methods, as our methods rely on adding noise only once to the statistics before propagation, where, in contrast, first-order methods need to repeatedly add noise to the gradient.  

We validate the framework on three healthcare applications of increasing scale and complexity: multicenter survival analysis on data from ACTG Study~175 under the Cox proportional hazards 
model \citep{cox1972regression}; survival analysis of advanced cancer patients from \citet{samstein2019tumor} assessing the prognostic value of tumor mutational burden; and a large-scale simulation of differentially private genetic region study on a federated network derived from the New York City hospital system, comprising $84$ hospitals across $16$ parent organizations with $N = 25{,}000$ simulated individuals. Our framework is able to achieve statistical
performance approaching the centralized baseline under practically deployable privacy budgets, at runtimes between $10\times$ and $1{,}000\times$  faster than encryption-based baselines \citep{froelicher2021truly, geva2023collaborative}, and with substantially  lower error than existing first-order differentially private optimization methods \citep{rizk2023enforcing}. Analysis of the New York City hospital  network further reveals that organizational-level federation simultaneously  dominates hospital-level federation in privacy strength, statistical power, and communication complexity providing a principled quantitative basis for governance decisions which, historically, are made without formal analytical grounding, with concrete managerial implications.

\subsection{Motivating Example: Federated Survival Analysis for Multicenter Clinical Trials} \label{sec:federation}

Survival analysis is widely used to examine how patients' survival outcomes (i.e., timing of events such as death, relapse, remission, recovery, etc.) vary in response to specific treatments or other health factors. Importantly, survival analysis accounts for censoring, which refers to the existence of data points where events of interest, such as death, hospitalization, or remission, have not occurred by the end of the study period, yet these data points are informative about the effect of treatments. Survival models account for censoring to give more accurate and unbiased estimates. This approach typically involves key metrics such as survival functions, hazard rates, and median survival times, allowing researchers to compare the effectiveness of different treatments or identify prognostic factors. Standard nonparametric and semiparametric models, such as the Kaplan-Meier curve \citep{kaplan1958nonparametric} and the Cox proportional hazards model \citep{cox1972regression}, allow us to analyze patient survival data with great flexibility in various clinical and experimental settings. By modeling time-to-event data, survival analysis plays a crucial role in clinical trials, epidemiological studies, and personalized medicine, helping clinicians and researchers make informed decisions about treatment strategies and patient care. Our simulation study in \cref{sec:sim} consists of survival analysis on a data set from the AIDS Clinical Trials Group (ACTG) network \citep{atcg_study175}, which has conducted several studies of HIV-infected and AIDS patients since the late 1980s \citep{hammer1996trial,campbell2012efficacy}.

Healthcare centers vary in the demographics of the patients that they serve, and they may even have different target groups (e.g., children's or women's hospitals). Although different centers can conduct clinical trials and hypothesis tests on their own, potential data limitations can yield false conclusions, for example, due to small sample sizes or higher prevalence of specific health conditions among some demographics that are under- or overrepresented in their patient population. For this reason, multiple healthcare centers can join multicenter trials to perform hypothesis tests with more accuracy, using larger and more diverse patient samples. To model multicenter trials, we divide patient data from the ACTG study 175 \citep{atcg_study175} equally at random between five centers and use the proportional hazards model \citep{cox1972regression} for survival analysis. For ease of exposition, we consider a simple hypothesis testing scenario to determine the efficacy and safety of a new treatment (ddI) against standard care (ZDV), which are antiretroviral HIV / AIDS medications in the ACTG study 175. For each center $i \in [n]$, we denote their data set of $n_i$ patients by $\vec s_i$. The patient $j$ in the center $i$ has the treatment variable $\vec x_{ij} \in \{0, 1\}$ which indicates if they have received the new treatment (ddI), $\vec x_{ij} = 1$, or standard care (ZDV), $\vec x_{ij} = 0$, and control variables $\vec z_{ij}$. The effect of the treatment is parametrized by $\theta_{\trt}$. The survival event for each patient is denoted by $\vec \delta_{ij} \in \{0, 1 \}$ with $\vec \delta_{ij} = 1$ corresponding to death (survival time is observed) and $\vec \delta_{ij} = 0$ corresponding to survival (survival time is censored). The survival or censoring time is measured in days and is indicated by $\vec t_{ij} \in \mathbb N$. Following the Cox proportional hazards model, the partial log-likelihood of the patient survival data for center $i$ is given by:
{
\begin{equation}
    \log \ell_i \left ( \vec s_i | \theta_\trt, \theta_\ctrl \right ) = \sum_{j: \vec \delta_{ij} = 1} \left (\theta_\trt \vec x_{ij}  + \vec \theta_\ctrl^T \vec z_{ij} - \log \sum_{j' : \vec t_{ij'} \ge \vec t_{ij}} e^{\theta_{\trt} \vec x_{ij'} + \theta_\ctrl^T \vec z_{ij'}} \right ). \label{eq:partial-liklihood}
\end{equation}}Note that using $\ell_i$ in \eqref{eq:partial-liklihood} is with some abuse of notation because it does not represent the full likelihood of the observed data; however, under Cox's proportional hazards model we can reliably infer the effect of the new treatment and other risk factors that can be included as covariates along with the treatment variable in \cref{eq:partial-liklihood}, from the partial likelihoods without specifying the full likelihoods or the functional form of the baseline hazard rate corresponding to $\vec x_{ij} =0$. We are interested in rejecting the simple null hypothesis that $\theta_\trt = 0$, i.e., that the two survival curves are the same, which, in the absence of other covariates, is equivalent to a nonparametric log-rank test of survival times between the treatment and control groups. For concreteness, we use $\theta_\trt = \theta_1 = -\log(2)$ as the simple alternative, which corresponds to patients being twice as likely to survive under the new treatment than in standard care ($e^{\theta_1}$ is the relative risk for the new treatment under the proportional hazards model). %  For the alternative hypothesis $\theta = \theta_1$, we are primarily interested in the cases where $\theta_1 < 0$, that is, the treatment increases the likelihood of survival (see, for example, \cref{subfig:log_hazard_sec2}). \arcomment{testing for $\theta= 0$ is equivalent to log-rank test - can include risk factors as patient covariates}

\begin{figure}[t]
    \centering
    %\subfigure[Centralized\label{subfig:hiv_curves_centralized}]{\includegraphics[width=0.3\linewidth]{log-linear/figures/survival_curves_hiv.pdf}}
    \subfigure[One hospital\label{subfig:hiv_curves_decentralized}]{\includegraphics[width=0.25\linewidth]{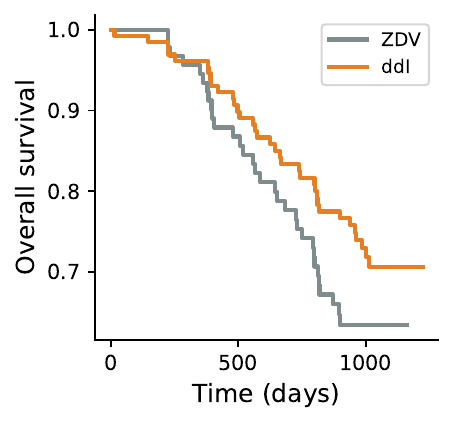}}
    \subfigure[Centralized\label{subfig:hiv_curves_centralized}]{\includegraphics[width=0.25\linewidth]{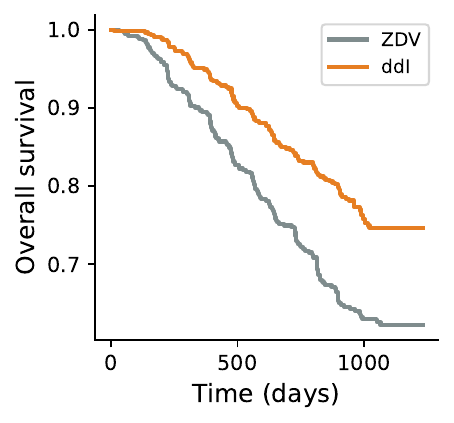}}
    \subfigure[Adjusted log hazard ratios\label{subfig:log_hazard_sec2}]{\includegraphics[width=0.25\linewidth]{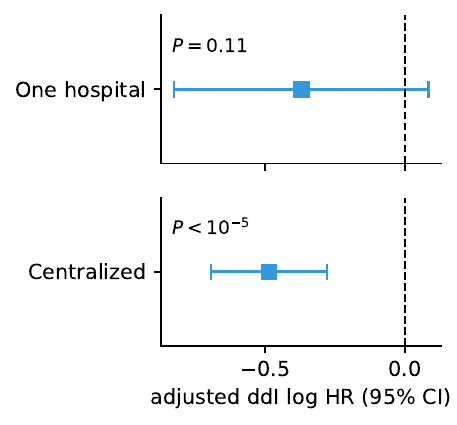}}
    \caption{Survival analysis for ACTG study 175 \citep{atcg_study175} comparing the ZDV and ddI treatments. ZDV stands for zidovudine, and ddI stands for didanosine, and the data is split equally among five hospitals. \textbf{Left and middle:} Kaplan-Meier survival curves for one hospital and for all the data pooled (centralized). \textbf{Right:} ddI log hazard ratios with 95\% confidence intervals from a proportional hazards model {adjusted} for baseline covariates (age, CD4 count, Karnofsky score, weight, prior antiretroviral exposure, race, and sex), fitted on all the data (centralized) and on one hospital. A single hospital cannot resolve the effect whereas the pooled analysis is highly significant.}
    \label{fig:hiv_curves}
    \vspace{-10pt}
\end{figure}

\cref{subfig:hiv_curves_decentralized} shows that the sample size of Kaplan-Meier survival curves calculated by one of the hospitals is not large enough to detect a difference between the two treatments, and, in fact, fitting the proportional hazards model using data from one hospital yields a $p$-value of $0.11$ in \Cref{subfig:log_hazard_sec2}. However, if centers were to pool their data (ignoring privacy concerns) through a centralized authority, then the centralized authority could perform the test at a larger sample size. \Cref{subfig:hiv_curves_centralized} shows the Kaplan-Meier survival curves are easily distinguishable in the centralized case. In \cref{subfig:log_hazard_sec2}, when all data are pooled together, the difference between $\log$-hazard ratios (log HR) from fitting proportional hazards models for the two treatments is statistically significant at $p<0.001$.

In a distributed setting where privacy concerns limit data sharing and preclude data pooling for centralized access, centers can exchange noisy information locally to achieve distributed hypothesis testing with privacy guarantees. In the private regime, inspired by the likelihood ratio test, we propose that each center calculate its local (partial) $\log$-likelihood ratio statistic $\log \left ( \frac {\ell_i(\vec s_i | \theta_\trt = \theta_1, \; \theta_\ctrl)} {\ell_i (\vec s_i | \theta_\trt = 0, \; \theta_\ctrl)} \right )$, add appropriately chosen Laplace noise $\vec d_i$, and then exchange the noisy statistic with its neighbors; see the simulation study in \Cref{sec:sim}. The methods we devise in \Cref{sec:prelim,sec:dp-algs} allow centers to form belief statistics locally, in a privacy-preserving manner, and use them for collective hypothesis testing with guarantees of false positive and false negative rates. 

\section{Belief Propagation for Differentially Private Distributed Inference}\label{sec:prelim} 

More generally, each center can have access to private observations with different likelihoods (for example, representing different populations of patients in different centers), which are parametrized by a common, finite set of alternatives. The goal of the centers is to exchange information to calculate the likelihoods of their collective observations and to choose a common set of maximum likelihood estimates (MLEs). To this end, centers exchange (non-Bayesian) belief statistics and decide on a common set of best alternatives (collective likelihood maximizers) based on their convergent beliefs. This MLE setup has natural extensions to hypothesis testing and online learning that we explore in \cref{sec:hypothesis_testing} and \cref{sec:app:dp-online-learning}, respectively. Before focusing on the theoretical performance and privacy guarantees of our distributed belief exchange algorithms in \cref{sec:dp-algs}, we introduce our information environment and the setup of distributed inference in \cref{sec:formulation}. Our belief exchange rules have a $\log$-linear format, which has normative foundations in decision analysis \citep{abbas2009kullback}, and can also be justified on the basis of their convergence properties \citep{olshevsky2017linear}. Due to length constraints, we elaborate on these connections in \cref{sec:app:normative-foudations}.

\subsection{Problem Formulation: Distributed Inference \& Learning in Discrete Spaces}\label{sec:formulation}

A collection of $n$ centers, denoted by the set $[n] = \{ 1, \dots, n \}$, wants to select an alternative from a finite set $\Theta$. Each center $i$ has its own data, which are fixed at the beginning or arrive in an online stream over time. The data follow a parametric model given by its likelihood function $\ell_i(\cdot | \theta), \theta \in \Theta$. The goal of the centers is to exchange information to select a set of alternatives that best describes their data collectively, in a maximum likelihood sense. Information exchange between centers (e.g., collaborating hospitals) is constrained by organizational and legal barriers, as well as \emph{privacy regulations} (e.g., HIPAA or GDPR) that limit who can exchange what information with whom. In our framework, organizational and legal barriers are captured by the structure of the communication network that limits information exchange to permissible local neighborhoods, and protection of private data (e.g., protected health information at each hospital) is achieved through differential privacy.

Centers are connected according to an undirected graph $\cG([n], \cE)$ on $n$ nodes indexed by $[n]$ with an edge set $\cE$ that contains no self-loops. The neighborhood of center $i$ in $\cG$ is denoted by $\cN_i$ and corresponds to all centers with whom center $i$ can exchange information locally. The graph is associated with an irreducible and aperiodic (primitive), doubly stochastic adjacency matrix $A = \left [ a_{ij} \right ]_{i, j \in [n]}$ with weights $a_{ij} = 0$ whenever $(i, j) \notin \cE$ and weight $a_{ij} > 0$ for all $(i, j) \in \cE$, and moreover, $\sum_{i \in [n]} a_{ij} = \sum_{j \in [n]} a_{ij} = 1$. The $n$ eigenvalues of the adjacency matrix $A$ are ordered by their modulus and denoted by $0<|\lambda_n(A)| \le \dots \leq |\lambda_2(A)| < \lambda_1(A) = 1$, with their associated set of bi-orthonormal eigenvectors, $\{ \vec l_i, \vec r_i \}_{i \in [n]}$, satisfying $\| \vec l_i \|_2 = \| \vec r_i \|_2 = \vec l_i^T \vec r_i = 1$ for all $i \in [n]$ and $\vec l_i^T \vec r_j = 0$ for all $i \neq j$. We use $\ev{}{\cdot}$ and $\var{}{\cdot}$ to denote expectation and variance operators. 

\noindent \textbf{The Privacy Notion.} Differential privacy restrictions dictate the protection of individual data, that is, ensuring that information exchange mechanisms between centers satisfy the following criteria for $\eps$-DP.
\begin{definition}\label{def:DP}
    A mechanism $\cM_i: \cS \to \cR$, acting on private patient datasets (or signals), is said to be $\eps$-DP with respect to the patient dataset $\vec s \in \cS \subset \mathbb{R}^d$ if and only if for all $X \subseteq \cR$ we have that $\Pr  [ \cM_i (\vec s) \in X  ] \le e^\eps {\Pr  [ \cM_i (\vec s') \in X ]}$ for any pair of adjacent patient datasets $\vec s \sim \vec s'$ that differ in at most one patient's data.
\end{definition}
% A mechanism $\cM: \cS \to \cR$ acting on the centers' signals, $\vec s \in \cS \subseteq  \mathbb R^d$, is $(\eps, \delta)$-DP if $\Pr [\cM(\vec s) \in X] \le e^\eps \Pr [\cM(\vec s') \in X] + \delta$ for any pair of adjacent observations $\vec s, \vec s'$ and any subsets $X$ of the range space $\cR$. 
The $\eps$-DP constraint on $\cM_i$ ensures that as $\eps\to 0$, no information can be inferred about whether $\vec s$ or any of its adjacent points (i.e., $\vec s' \in \cS: \vec s' \sim \vec s$) are the input that generates the observed output $\cM_i(\vec s)$. Our primary interest in this work is in belief exchange mechanisms for which the range space $\cR$ is the probability simplex of all distributions over the state space $\Theta$, denoted by $\mathrm {Simplex} (\Theta)$. Defining the appropriate notion of adjacency allows us to control sensitive information leaks. %When working with study data, we declare two observations $\vec s$ and $\vec s'$ adjacent if, and only if, they differ in the data of one patient. 

\noindent \textbf{The Distributed Inference Problem.} We consider a distributed information environment, where the goal is to estimate the best alternative or perform a hypothesis test in a distributed manner based on a set of initial observations. %; in the second, 
In \cref{sec:app:dp-online-learning},  we consider an online learning extension where the goal is to learn the true alternative online by repeatedly observing data streams over time. %In the sequel, we present these two settings. The corresponding nonprivate algorithms are provided in \cref{sec:app:learning-non-priv} and serve as benchmarks to evaluate our DP designs. 
In multicenter studies, the online learning setup corresponds to centers recruiting patients over time. The centers can rely on each other's observational capabilities to recruit a diverse patient demographic, which they may otherwise lack access to, compromising their ability to detect effects in certain populations.

%\noindent \textbf{Distributed MLE.} 
In the distributed maximum likelihood estimation task (MLE), there is a set of likelihood maximizers $\Theta^\star \subset \Theta$, and centers aim to determine $\Theta^\star$ collectively by combining their private patient datasets while communicating within their local neighborhoods. Specifically, each center has access to a dataset $\vec s_i \in \cS$ of $n_i$ patients that obeys a likelihood function $\ell_i (\vec s_i | \hat \theta)$ for all $\hat \theta \in \Theta$. The centers' task is, given patient datasets $\vec s_i$ for each center $i \in [n]$, to find the set of likelihood maximizers, i.e., 
{\begin{equation*}
  \Theta^\star = \argmax_{\hat \theta \in \Theta} \Lambda(\hat \theta), \text{ where } \Lambda(\hat \theta) = \sum_{i = 1}^n \log \left ( \ell_i(\vec s_i | \hat \theta) \right ).
\end{equation*}} We define $\compTheta = \Theta \setminus \Theta^\star$ as the set of non-optimal states. We let $f^\star$ represent the proportion of MLE states, i.e., $f^\star = |\Theta^\star| / |\Theta|$. \citet{rahimian2016distributed} give a non-private algorithm for belief exchange, described in \Cref{sec:app:learning-non-priv-mle}, that is able to recover $\Theta^\star$ asymptotically.

In the multicenter clinical trial example of \cref{sec:federation}, the state space is $\Theta = \{ 0, \theta_1 \}$, where $\theta=0$ corresponds to the null hypothesis that the two treatments are statistically indistinguishable, and $\theta = \theta_1 < 0$ corresponds to the hypothesis that ddI improves patient survival compared to ZDV, as is the case in \cref{fig:hiv_curves}; hence, $\Theta^\star = \left \{ \theta_1 \right \}$. The goal of the centers is to collectively infer that $\theta = \theta_1$ is the best state (MLE) given the data of the patients because $\Lambda (\theta_1) > \Lambda (0)$. In \cref{sec:hypothesis_testing}, we show how distributed MLE algorithms can be modified for distributed hypothesis testing with statistical significance guarantees.   

\subsection{Performance Analysis Framework}\label{sec:performance-metrics}

Our proposed log-linear update rules rely on adding zero-mean Laplace noise to the $\log$-likelihood statistics of the private data. Ideally, by adding zero-mean noise in the MLE problem, we would anticipate that log-linear updates should recover MLE states $\Theta^\star$ in expectation. However, the added noise makes our algorithm output random, so we need a systematic way to control the error of the output and achieve a desired accuracy level by repeating the algorithm. Thus, providing high-probability guarantees for our algorithms comes at a cost: The algorithms need to be repeated in sufficiently many rounds, and their outputs combined across these rounds. 

\noindent{\bf Type I and Type II error rates.} To control the output uncertainty resulting from privacy-preserving randomization noise, our distributed MLE algorithms, in addition to respecting a set DP budget $\eps > 0$, also admit two types of error guarantees, which we refer to as Type I and Type II error probabilities and denote by $\alpha$ and $1 - \beta$, following the hypothesis testing nomenclature, where $\beta$ denotes the power of the test. In fact, in \cref{sec:hypothesis_testing} (with proofs and convergence analysis in \cref{sec:app:hypothesis_testing}), we show that our distributed MLE algorithms and guarantees can be adapted to perform a distributed hypothesis test at the significance level $\alpha$. For a distributed MLE algorithm $\cA$ that returns a set of maximum likelihood estimators $\hat \Theta^\cA \subseteq \Theta$, we define the \emph{Type I} error rate $\alpha$ as the probability of the event that $\hat \Theta^\cA$ includes non-MLE states, i.e., $\left \{ \Theta^\star \not \supseteq \hat \Theta^\cA \right \}$, which can be controlled by bounding $\Pr \left [ \Theta^\star \supseteq \hat \Theta^\cA \right ] \ge 1 - \alpha$. Similarly, we define the \emph{Type II} error rate $1 - \beta$ as the probability of the event that $\hat \Theta^\cA$ fails to include some MLE states, i.e., $\left \{ \Theta^\star \not \subseteq \hat \Theta^\cA \right \}$, which can be controlled by bounding $\Pr \left [ \Theta^\star \subseteq \hat \Theta^\cA \right ] \ge \beta$. %A Type I error occurs when the algorithm detects a superset of the MLE set $\Theta^\star$. 
A Type I error occurs whenever $\Theta^\star \not \supseteq \hat \Theta^\cA $, for example, when the algorithm outputs too many states, including some that do not belong to $\Theta^\star$. A Type II error occurs whenever $\Theta^\star \not \subseteq \hat \Theta^\cA $, for example, when the algorithm filters too many states, missing some MLE states in its output.%, thus generating a subset of the MLE set $\Theta^\star$. 

Returning to our examples from \cref{sec:introduction}, when selecting among treatments with severe side effects, we want to limit the possibility of selecting ineffective treatments by ensuring a small Type I error rate $\alpha$ so that $\Pr \left [ \Theta^\star \supseteq \hat \Theta^\cA \right ] \ge 1 - \alpha$. Similarly, when conducting a clinical trial or a genome-wide association study, we need to ensure the efficacy of treatments in clinical trials or the discovery of significant single nucleotide polymorphisms (SNPs) associations at a desired level of statistical significance (e.g., $\alpha = 0.05$). On the other hand, when developing a new cancer screening tool, we want MLE states to be included in the algorithm output ($\{ \Theta^\star \subseteq \hat \Theta^{\mathcal A} \}$) with high probability, that is, for $\beta$ to be high (e.g., $\beta = 0.8$). This will ensure that we can detect all patients who are at high risk of cancer ($\Theta^\star$). In this case, it is important to control the Type II error rate so that $\Pr \left [\Theta^\star \not\subseteq \hat \Theta^{\mathcal A} \right ] \le 1 - \beta$ for a large enough $\beta$, but it may be acceptable to misidentify some patients as high risk (i.e., for $\Theta^{\mathcal A}$ to include some non-MLE states so that $\hat \Theta^\cA \not \subseteq \Theta^\star$). In \cref{sec:dp-dist-mle}, we show that arithmetic or geometric averaging of beliefs across the rounds may each be suitable, depending on the type of error that one needs to preclude or control. 
% \mar{add genomics example as a case of Type I error rate control}

\noindent{\bf Communication complexity.} To achieve guarantees on the two error probabilities, $\alpha$ and $1-\beta$, subject to a DP budget $\eps$, we must run our algorithms in $K\left (\eps, \eta, \Theta, \cG, \left \{ \ell_i(\cdot|\cdot) \right \}_{i \in [n]} \right )$ rounds and with $T \left (\eps, \eta, \Theta, \cG, \left \{ \ell_i(\cdot|\cdot) \right \}_{i \in [n]} \right )$ iterations in each round, where $\eta$ corresponds to either $\alpha$ or $1 - \beta$, or the maximum of the two (depending on the algorithm and the error controls that it provides). We define $$\text{Communication Complexity} = K\left (\eps, \eta, \Theta, \cG, \left \{ \ell_i(\cdot|\cdot) \right \}_{i \in [n]} \right ) \cdot T \left (\eps, \eta, \Theta, \cG, \left \{ \ell_i(\cdot|\cdot) \right \}_{i \in [n]} \right ),$$ as the total number of belief updates. Beyond the privacy budget $\eps$, the error probability $\eta$, the number of centers $n$, and the algorithm design parameters, the communication complexity bounds (see also \cref{sec:app:summary} for a summary of the results) depend on the following statistical properties of the information environment:

\begin{itemize}
    \item \emph{\textbf{Private signal structures:}} In the MLE case, our bounds depend on the largest $\log$-likelihood magnitude: 
    { \begin{equation}
    \Gamma_{n, \Theta} = \max_{i \in [n], \hat \theta \in \Theta} |\log \vec \gamma_{i}(\hat \theta)|, \label{eq:Gamma}
    \end{equation}} the minimum divergence between any non-MLE state and an MLE state:
    {\begin{equation} l_{n, \Theta} = \min_{\bar \theta \in \compTheta, \theta^\star \in \Theta^\star} \left | \sum_{i = 1}^n \kldiv {\ell_i(\cdot | \bar \theta)} {\ell_i(\cdot | \theta^\star)} \right |\label{eq:l}
    \end{equation}}and the maximum standard deviation of the $\log$-likelihood ratio statistics: 
    {\begin{equation}
    Q_{n, \Theta} = \max_{i \in [n], \bar \theta \in \compTheta, \theta^\star \in \Theta^\star} \sqrt {\var {} {\log \left ( \frac {\ell_i(\vec s_i | \bar \theta)} {\ell_i(\vec s_i | \theta^\star)} \right )}} \label{eq:Q}
    \end{equation}}
    \item \emph{\textbf{Sensitivity:}} To satisfy the $\eps$-DP requirement in \cref{def:DP}, the noise $\vec d_i$ can be set according to the Laplace mechanism \citep[Section 3.3]{dwork2014algorithmic}, which adds Laplace distribution noise in the $\log$ space: $\vec d_i \sim \cD = \mathrm {Lap} \left ( \frac {K|\Theta|\Delta_{i, \Theta}} {\eps} \right )$, where ${K|\Theta|\Delta_{i, \Theta}}/{\eps}$ is the scale parameter of the Laplace distribution and the global $\ell_1$-sensitivity, $\Delta_{i, \Theta}$, is defined as: 
{\begin{equation}
    \Delta_{i, \Theta} = \max_{\hat \theta \in \Theta} \left \{ \max_{\vec s_i  \sim \vec s'_i} \big \| \log \ell_i(\vec s_i | \hat \theta) -  \log \ell_i(\vec s'_i | \hat \theta)\big \|_1 \right \}. \label{eq:sensitivity}
\end{equation}}

    Intuitively, when $\log$-likelihoods are highly sensitive to signal values more noise is required to mask the input signals at the $\eps$-DP level. Here we have used $\eps/(K|\Theta|)$ privacy budget per state per round to ensure that the initial beliefs are $\eps$-DP overall by the composition property of DP \citep[Theorem 3.14]{dwork2014algorithmic}, after computing its value at all states $\hat{\theta}\in\Theta$. We state the non asymptotic bounds in terms of the maximum of the sensitivities $\Delta_\Theta^n = \max_{i \in [n]} \Delta_{i, \Theta}$.
    %\item \emph{\textbf{Network structure:}} 
\end{itemize}

\section{Performance and Privacy Guarantees for Distributed Inference} \label{sec:dp-algs}

\subsection{Differentially Private Distributed MLE}\label{sec:dp-dist-mle}

\begin{algorithm}[t]
\footnotesize
\captionsetup{font=footnotesize}
\caption{Private Distributed MLE (AM/GM)} \label{alg:private_distributed_mle}

\begin{flushleft}
\noindent \textbf{Inputs:} Privacy budget $\eps$, Error probabilities $\alpha, 1 - \beta$, Log-belief Thresholds $\varrho^\AM, \varrho^\GM > 0$.

\noindent \textbf{Initialization:} Set the number of iterations $T$ and the number of rounds $K$ as indicated by \cref{theorem:non_asymptotic_private_distributed_mle} (for the non-asymptotic case), and $\tau^\AM = 1 / (1 + e^{\varrho^\AM})$ (resp. for $\tau^\GM$). The DP noise distribution for protecting beliefs is $\cD_{i} (\hat \theta; \eps)$ which can be set optimally according to \cref{theorem:non_asymptotic_private_distributed_mle}, independently and identically across the states $\hat \theta \in \Theta$.

\noindent \textbf{Procedure:} The following is repeated in $K$ rounds indexed by $k \in [K]$. In each round $k$, centers begin by forming noisy likelihoods $\vec \sigma_{i, k}(\hat \theta) = e^{\vec d_{i, k} (\hat \theta)} \vec \gamma_i(\hat \theta)$, where $ \vec\gamma_{i}(\hat{\theta}) = \ell_i(\vec{s}_{i}|\hat{\theta})$ and $\vec d_{i, k} (\hat \theta)  \sim \cD_{i} (\hat \theta; \eps)$ independently across centers $i \in [n]$ and states $\hat \theta \in \Theta$. The centers initialize their beliefs to ${\vec\nu}_{i,k,0}(\hat{\theta})$ $ =$ $  \vec\sigma_{i, k}(\hat{\theta}) /\sum_{\tilde{\theta}\in\Theta} \vec\sigma_{i, k}(\tilde{\theta})$, and over the next $T$ iterations, they communicate with their neighbors and update their beliefs accordingly:  
\begin{equation*}
\hspace{-10pt}{\vec\nu}_{i,k,t}(\hat{\theta})
=  \frac{{\vec\nu}^{1+a_{ii}}_{i, k, t-1}(\hat{\theta}) \prod\limits_{j\in\mathcal{N}_i }{{\vec\nu}^{a_{ij}}_{j, k,t-1}(\hat{\theta})} }{\sum\limits_{\tilde{\theta} \in \Theta}{\vec\nu}_{i,k, t-1}^{1+a_{ii}}(\tilde{\theta}) \prod\limits_{j\in\mathcal{N}_i} {{\vec\nu}^{a_{ij}}_{j, k,t-1}(\tilde{\theta})} } \quad \text{for every $\hat \theta \in \Theta$ and $t \in [T]$; repeated in rounds $k \in [K]$.} \end{equation*} 

After $T$ iterations in $K$ rounds, the centers aggregate the outcome of the $K$ rounds:
\begin{equation*}
    {\vec \nu}_{i, T}^{\AM}(\hat \theta)  = \frac {\sum_{k \in [K]} \vec \nu_{i, k, T} (\hat \theta)} {K} \mathpunct{\raisebox{0.5ex}{,}} \quad {\vec \nu}_{i, T}^{\GM}(\hat \theta)  = \frac {\prod_{k \in [K]} \vec \nu_{i, k, T} (\hat \theta)^{1/K}} {\sum_{\tilde \theta \in \Theta} \prod_{k \in [K]} \vec \nu_{i, k, T} (\tilde \theta)^{1/K}} \quad \text{for every $\hat \theta \in \Theta$.}
\end{equation*}

\noindent \textbf{Outputs:} Return 
\begin{equation*}
    \hat \Theta^\AM_{i, T}   = \left \{ \hat \theta \in \Theta : \vec \nu_{i, T}^\AM (\hat \theta) \ge \tau^\AM \right \} \mathpunct{\raisebox{0.5ex}{,}} \quad \hat \Theta^\GM_{i, T}  = \left \{ \hat \theta \in \Theta : \vec \nu_{i, T}^\GM (\hat \theta) \ge \tau^\GM \right \} \cdot    
\end{equation*}

\end{flushleft}

\end{algorithm}

The first idea behind the MLE algorithm is to introduce multiplicative noise subject to DP to the likelihood functions $\vec \gamma_i(\hat \theta)$. Multiplicative noise, which corresponds to additive noise in the log domain, ensures privacy. The centers then communicate their beliefs about the states with their local neighbors and form estimates of the true MLE over time. However, introducing noise makes the hypothesis selection randomized; therefore, the centers may misidentify the MLE states with a non-zero probability. To overcome this problem and guarantee Type I (resp. Type II) error at most $\alpha$ (resp. $1 - \beta$), we run the algorithm for $K$ independent rounds and combine these $K$ estimates to produce the final beliefs. In \cref{theorem:asymptotic_private_distributed_mle}, we show how an appropriate choice of $K$ allows us to control the algorithm's errors. 

To produce the final estimates, we rely on two ways of aggregating the $K$ rounds of the algorithm: The first approach is to take the arithmetic mean of the $K$ rounds to produce the final belief, which we call the AM estimator. The benefit of the AM estimator is that it can accurately, with high probability, identify all the MLE states; i.e., we can control its Type II error probability: $\Pr \left [ \Theta^\star \not\subseteq \hat \Theta^\AM_{i, T} \right ] \leq 1 - \beta$ for all centers $i\in[n]$ and $K$ and $T$ large enough. This estimator has high recall but low precision because it can recover a superset of $\Theta^\star$. The second way to combine the $K$ rounds is to take the geometric mean, which we call the GM estimator. For the GM estimator, we can control the Type I error probability: $\Pr \left [ \hat \Theta^\GM_{i, T} \not\subseteq \Theta^\star \right ] \leq \alpha$, by choosing $K$ and $T$ large enough to ensure that it recovers a subset of $\Theta^\star$ with high probability. We first start with a result on the asymptotic behavior of \cref{alg:private_distributed_mle}. Specifically, we show that we can recover the MLE with high probability by repeating the algorithm $K$ times for appropriately chosen $K$ (proved in \cref{sec:app:proof:theorem:asymptotic_private_distributed_mle}):

% \cref{fig:aggregate_mle} shows how repeating the algorithm and averaging the results over $K$ independent runs yields higher values for the belief of the MLE state compared to just applying DP in one round, as shown in \cref{fig:hiv_curves}. 

\begin{theorem} \label{theorem:asymptotic_private_distributed_mle}
    Consider \cref{alg:private_distributed_mle} run with state-independent noise distributions $\cD_{i} (\hat \theta; \eps) = \cD_i(\eps)$ that satisfy $\eps$-DP for a fixed $\eps > 0$. Then, for every center $i \in [n]$, the following hold:
    
    % as $\varrho^\AM \to \infty$ and $\varrho^\GM \to \infty$, 
    
    \begin{itemize}
        \item  \textbf{Type II error (AM estimator):} if $K \ge |\compTheta| \log \left (|\Theta^\star| / (1 - \beta) \right )$, then $\lim_{T \to \infty} \Pr \left [ \Theta^\star \subseteq \hat \Theta^\AM_{i, T} \right ] \ge \beta$; moreover, $\lim_{T \to \infty} \Pr \left [|\Theta \setminus \hat \Theta^\AM_{i, T}| = 1 \right ]  \ge \beta - e^{-K/|\compTheta|}$.
        \item \textbf{Type I error (GM estimator):} if $K \ge |\Theta^\star| \log \left (|\compTheta| / \alpha \right )$, then $\lim_{T \to \infty} \Pr \left [ \hat \Theta^\GM_{i, T} \subseteq \Theta^\star \right ] \ge 1 - \alpha$; moreover, $\lim_{T \to \infty} \Pr \left [ |\hat \Theta_{i, T}^\GM | = 1 \right ] \ge 1 - \alpha - |\Theta^\star| e^{-K/|\Theta^\star|}$.
        \item The beliefs exchanged and the resulting estimates are $\eps$-DP with respect to private signals.
    \end{itemize}

\end{theorem}

\pfsketch{Note that as $\varrho^\AM \to \infty$ in \cref{alg:private_distributed_mle}, $\tau^\AM \to 0$ (resp. for $\tau^\GM$).  At a high level for the AM estimator, we show that the probability that an MLE state $\theta^\star \in \Theta^\star$ ends up with a non-zero belief at a given round $k \in [K]$ is at least $1 / |\compTheta|$, and therefore we can show that the probability that the AM estimator gives a zero value to an MLE state is exponentially small, that is, at most $e^{-K/|\compTheta|}$, since the algorithm is repeated $K$ times independently. Therefore, the probability that any MLE state is misclassified by the AM estimator is at most $|\Theta^\star| e^{-K/|\compTheta|}$ by the union bound. Requiring this probability to be less than $1 - \beta$ yields the value of $K$. A similar argument also works for the GM estimator.}

\smallskip

Repeating \cref{alg:private_distributed_mle} for $K \ge |\Theta| \log (|\Theta| / \min \{ \alpha, 1 - \beta \} )$ iterations, one can assert $\hat \Theta^\GM_{i, T} \subseteq \Theta^\star \subseteq \hat \Theta^\AM_{i, T}$ as $T \to \infty$ with probability $\beta - \alpha$ without knowing $f^\star$. Also, it is interesting to point out that the above theorem holds regardless of the noise distribution, as long as the noise distribution does not depend on the state for a given center and it satisfies $\eps$-DP. 

In practice, the hyperparameters $K$ and $T$ can be set optimally using \cref{theorem:non_asymptotic_private_distributed_mle}, which provides explicit, closed-form expressions for both. Moreover, for the common case of hypothesis testing with two states, the GM estimators recover the correct hypothesis exactly, as confirmed by our theoretical and empirical results in \cref{sec:hypothesis_testing,sec:sim}, respectively. For richer state spaces, our two-threshold algorithm (deferred to \cref{app:private_mle_two_threshold}) enables 
simultaneous control of both Type-I and Type-II errors without this tension. Nevertheless, it is instructive to understand the role of $K$ in the AM/GM setting: the error probability decreases exponentially in $K$, while the noise standard 
deviation scales linearly, creating a bias--variance-like trade-off with increasing $K$. This is reflected in \cref{theorem:asymptotic_private_distributed_mle}: as 
$K, T \to \infty$, the GM estimator concentrates on a singleton (high precision, 
potentially lower recall), while the AM estimator expands toward the full set (high 
recall, potentially lower precision). Choosing $K$ as prescribed by \cref{theorem:non_asymptotic_private_distributed_mle}
optimally balances these effects and yields the target error rates $\alpha$ and $1 - \beta$.

% Finally, one may be tempted to set a high value for $K$; however, our results show that as $K, T \to \infty$ the GM estimator will return a singleton with high probability (at least $1 - \alpha$), potentially missing other MLE states from $\Theta^*$, i.e., yielding a high Type II error (low recall). Similarly, as $K, T \to \infty$, the AM estimator returns all but one element with high probability (at least $\beta$), potentially including many non-MLE states from $\Theta^*$, i.e., yielding a high Type I error (low precision). This is driven by a trade-off between reducing the error probability of a randomized algorithm by repeating it $K$ times, where the error decreases exponentially with $K$, independently, and the standard deviation of the total noise added, which scales linearly with $K$.   

The distributed MLE algorithm is repeated $K$ times for each of the $|\Theta|$ states, and an attacker would have more observations of the outputs from the same signals; therefore, we need to rescale the privacy budget by $K|\Theta|$ to account for the number of runs and states. Thus, choosing $\cD_i(\eps) = \mathrm {Lap} \left ( \frac {K |\Theta| \Delta_{i, \Theta}} {\eps} \right )$ satisfies $\eps$-DP and the assumptions of \cref{theorem:asymptotic_private_distributed_mle}. Our nonasymptotic analysis (presented next) shows that $\cD_i(\eps) = \mathrm {Lap} \left ( \frac {K |\Theta| \Delta_{i, \Theta}} {\eps} \right )$ is indeed optimal to minimize the required number of iterations per round. 

The previous analysis focused on the regime where $T \to \infty$. The next question that arises is to study the behavior of \cref{alg:private_distributed_mle} for finite $T$. As expected, the speed of convergence depends on the sum of standard deviations of the noise, that is, $V_{n, \eps, \Theta} = \sum_{i = 1}^n \sqrt {\var {\vec d_i \sim \cD_i(\eps)} {\vec d_i}}$. Our main result for private distributed MLE using the AM/GM method follows (proved in \cref{sec:app:proof:theorem:non_asymptotic_private_distributed_mle}): 
\begin{theorem} \label{theorem:non_asymptotic_private_distributed_mle}
   Let $a^\star_n = |\lambda_2(A + I)|/2$ be the second-largest eigenvalue modulus (SLEM) of $(A + I) / 2$. For \cref{alg:private_distributed_mle} with thresholds $\varrho^\AM, \varrho^\GM > 0$ and privacy budget $\eps > 0$, the following hold for every center $i \in [n]$:
   \vspace{1pt}
   \begin{itemize}
           \item For {\footnotesize $T \ge \max \left \{ \frac {\log \left ( \frac {2 \varrho^{\GM} n} {l_{n, \Theta}} \right )} {\log 2}, \frac {\log \left ( \frac {|\Theta|^2 (n-1) (n\Gamma_{n, \Theta} + V_{n, \eps, \Theta})} {2 \alpha \varrho^\GM \sqrt K} \right )} {\log (1 / a_n^\star)}\right \}$} and $K \ge |\Theta^\star| \log (|\compTheta| / \alpha)$, we have $\Pr [\hat \Theta^\GM_{i, T} \subseteq \Theta^\star] \ge 1 - 2 \alpha$. Moreover, $\Pr [\hat{\Theta}^{\mathrm{GM}}_{i,T} \neq \emptyset] \geq 1 - 2\alpha - |\Theta^\star|e^{-K/|\Theta^\star|}$.
           \item For {\footnotesize $T \ge \max \left \{ \frac {\log \left ( \frac {2 \varrho^{\AM} n} {l_{n, \Theta}} \right )} {\log 2}, \frac {\log \left ( \frac {|\Theta|^2 (n-1) K (n\Gamma_{n, \Theta} + V_{n, \eps, \Theta})} {2 \log (1 / (1 - \beta)) \varrho^\AM}  \right )} { \log (1 / a_n^\star)} \right \}$} and $K \ge |\compTheta| \log (|\Theta^\star| / (1 - \beta))$, we have $\Pr [\Theta^\star \subseteq \hat \Theta^\AM_{i, T}] \ge 2 \beta - 1$. Moreover, $\Pr [\hat{\Theta}^{\mathrm{AM}}_{i,T} \neq \Theta] \geq 2\beta - 1 - e^{-K/|\bar{\Theta}|}$.
           \item    The noise distributions that minimize the convergence time $T$ are the Laplace distributions $\cD_i^\star (\eps) = \mathrm{Lap} \left ( \frac {\Delta_{i, \Theta} K |\Theta|} {\eps} \right )$.
           \item The beliefs exchanged and the resulting estimates are $\eps$-DP with respect to private signals.
   \end{itemize}   
\end{theorem}

Recall that $\Gamma_{n, \Theta}$ and $l_{n, \Theta}$ that appear in the communication complexity bounds in \cref{theorem:non_asymptotic_private_distributed_mle} are properties of the signal structure defined in \cref{eq:Gamma,eq:l}. To minimize communication complexity, it suffices to pick the optimal threshold that makes the terms inside the maximum equal.

%, and, hence, the maximum is minimized, which corresponds to $\varrho^{\AM}_\star = \left ( \frac {|\Theta|^2 (n - 1) K (n \Gamma_{n, \Theta} + V_{n, \eps, \Theta})} {2 \log (1 / (1 - \beta))} \right )^{\frac {\log 2} {\log (2 / a_n^\star)}} \left ( \frac {l_{n, \Theta}} {2 n} \right )^{\frac {-\log a_n^\star} {\log (2 / a_n^\star)}},$ for AM and $\varrho^{\GM}_\star = \left ( \frac {|\Theta|^2 (n - 1) (n \Gamma_{n, \Theta} + V_{n, \eps, \Theta})} {2 \alpha \sqrt K} \right )^{\frac {\log 2} {\log (2 / a_n^\star)}} \left ( \frac {l_{n, \Theta}} {2 n} \right )^{\frac {-\log a_n^\star} {\log (2 / a_n^\star)}}$ for GM.\\

\noindent \textbf{A Two-threshold Algorithm for Distributed MLE.} Beyond the AM/GM algorithm, we develop a two-threshold algorithm that enables centers to jointly control both Type I and Type II errors in distributed maximum-likelihood estimation while preserving privacy. The method tracks how often each hypothesis’ belief exceeds two confidence thresholds across multiple communication rounds, effectively distinguishing MLE from non-MLE states. %Intuitively, the higher threshold ensures that only well-supported hypotheses are accepted, while the lower threshold safeguards against excluding true states that occasionally receive weaker evidence.
We show that, with appropriately chosen thresholds and a sufficient number of communication rounds, the algorithm identifies all MLE states while excluding non-MLE ones with high probability, achieving asymptotic recovery of the correct hypothesis set. Compared to single-threshold or averaging methods, the two-threshold approach offers finer control over false positive and false negative rates at a modest increase in communication complexity, as formalized in our asymptotic and non-asymptotic analyses. Due to length constraints, we defer the discussion of the two-threshold algorithm to \cref{app:private_mle_two_threshold}. \cref{sec:app:summary} presents a summary of the main results obtained from the two-threshold algorithm. 

\subsection{Differentially Private Distributed Hypothesis Testing} \label{sec:hypothesis_testing}

The results we devised above for the MLE have a natural application in the case of hypothesis testing with simple or composite hypotheses. Specifically, we can specialize our designs to decentralized hypothesis testing at a significance level $\alpha$ as follows: We run the GM algorithm for $T$ iterations and $K$ rounds and set a threshold $\varrho_d$ to reject the null hypothesis if the log-belief ratio exceeds it: $(n / 2^{T - 1}) \log (\vec \nu_{i, T}^\GM(1) / \vec \nu_{i, T}^\GM (0)) > \varrho_d$. For the case of a simple null and a simple alternative, the threshold $\varrho_d$ can be derived from the threshold $\varrho_c$ of the corresponding uniformly most powerful (UMP) centralized $\log$-likelihood ratio test (LRT) at level $\alpha / 2$. For sufficiently large $T$ and $K$, we can show that

\begin{proposition}[Simple Hypothesis Testing] \label{prop:hypothesis_testing}
    Let $\alpha \in (0, 1)$ be a significance level. Let $\varrho_c$ be the threshold of the UMP centralized log-likelihood ratio test at level $\alpha/2$, such that $\Pr [2 (\Lambda(1) - \Lambda(0)) \ge \varrho_c | \theta = 0] = \alpha/2$. Run the GM algorithm (\cref{alg:private_distributed_mle}) with a Type I error guarantee $\alpha/2$, with $T$ and $K$ as in \cref{theorem:non_asymptotic_private_distributed_mle} and the rejection threshold $\varrho_d =  \varrho_c - 1$. Then the decentralized test controls the Type I error at level $\alpha$, that is, {$$\Pr \left [   (n / 2^{T-1}) \log (\vec \nu_{i, T}^\GM(1) / \vec \nu_{i, T}^\GM (0)) > \varrho_d \big | \theta = 0\right ] \le \alpha,$$} and the resulting test is $\eps$-DP.
\end{proposition}

The idea behind the result is that, in practice, for sufficiently large $T$ and $K$, $(n/2^{T - 1}) \log (\vec \nu_{i, T}^\GM(1) / \vec \nu_{i, T}^\GM (0)) \sim 2 \Lambda(1, 0)$ with high probability (with probability at least $1 - \alpha/2$ due to Chebyshev's inequality). The formal proof of \Cref{prop:hypothesis_testing} is in \Cref{sec:app:proof:theorem:hypothesis_testing}.

However, in general, when the distribution of the centralized statistic under the null is not available, we cannot choose $\varrho_c$ based on the UMP centralized LRT. In that case, the centers can perform a generalized likelihood ratio test locally and propagate the statistic. This sacrifices the optimality of the UMP test; however, it is more practical in real-world scenarios where the distribution of the sum of the statistics may be unknown under the null hypothesis. 

Specifically, the generalized likelihood ratio test can be used to test a composite null hypothesis ($\theta \in \tilde \Theta_0$) against a composite alternative ($\theta \in \tilde \Theta_1$, $\tilde \Theta_0 \cap \tilde \Theta_1 = \emptyset$), and requires a log-likelihood function that is twice differentiable and defined in a continuous parameter space $\tilde \Theta = \tilde \Theta_0 \cup \tilde \Theta_1$. Specifically, in that case, each center initializes their belief using $\vec \gamma_i(\tilde \Theta_0) = \sup_{\tilde \theta_0 \in \tilde \Theta_0} \ell_i(\vec s_i | \tilde \theta_0)$ and $\vec \gamma_i(\tilde \Theta) = \sup_{\tilde \theta \in \tilde \Theta} \ell_i(\vec s_i | \tilde \theta)$, and runs the GM algorithm with the binary (meta) state space $\Theta = \{ \tilde \Theta_0, \tilde \Theta \}$. By Wilks' theorem \citep[Theorem 10.3.1]{casella2002statistical}, for a large number of patients $n_i$, each local log-likelihood ratio statistic is asymptotically distributed according to the $\chi^2_1$ distribution, and thus the sum of independent local statistics of the $n$ centers follows a $\chi^2_n$ distribution. We can use the asymptotic distribution of the sum of the local log-likelihood ratio statistics to set the rejection threshold for a distributed level $\alpha$ test to be $\varrho_d = F^{-1}_{\chi^2_n}(1 - \alpha/2) - 1$. In contrast, the centralized test has a threshold of $\varrho_c = F^{-1}_{\chi_1^2}(1 - \alpha)$. We provide additional details for the extension to composite hypothesis tests in \Cref{sec:app:composite-extension}.

\noindent \textbf{Extensions to Online Learning from Intermittent Streams.} For prospective multicenter trials and epidemiological surveillance where centers recruit patients continuously over time, we extend the framework to online learning from intermittent data streams. Unlike the MLE setting, the online setting does not require repeating the algorithm across multiple rounds because the effect of DP noise vanishes asymptotically as a consequence of the Ces\'aro mean and the weak law of large numbers. We provide finite-sample convergence guarantees characterizing the communication cost of privacy and show that the Laplace mechanism remains optimal for minimizing convergence time. Due to space considerations, the algorithm and its analysis are presented in \cref{sec:app:dp-online-learning}.

\section{Real-world Healthcare Data Sharing Applications}\label{sec:sim}

We validate the proposed framework on three healthcare applications of 
increasing scale and institutional complexity. The first two applications 
involve multicenter survival analysis on real-world clinical trial data: 
ACTG Study~175, evaluating the comparative efficacy 
of antiretroviral treatments on patient survival under the Cox 
proportional hazards model \citep{cox1972regression}, and a study of 
advanced cancer patients from \citet{samstein2019tumor}, assessing 
tumor mutational burden as a prognostic factor under the immune checkpoint 
inhibitor therapy. The third application is a large-scale differentially 
private genetic region study instantiated on a federated network 
derived from the New York City hospital system, comprising $84$ hospitals 
across $16$ parent organizations with $N = 25{,}000$ simulated individuals, 
designed to address the practical question of whether federation should 
occur at the hospital or organizational level. 
Across all settings, we benchmark the proposed algorithms against 
centralized baselines, non-private distributed baselines, 
encryption-based privacy-enhancing technologies 
\citep{froelicher2021truly, geva2023collaborative}, and first-order 
differentially private optimization methods \citep{rizk2023enforcing}, 
reporting statistical performance, runtime, and privacy-utility tradeoffs 
as functions of the privacy budget $\varepsilon$ and the network size $n$.

\subsection{Differentially Private Distributed Survival Analysis for Clinical Trials}

\begin{figure}[t]
    \centering
    \includegraphics[width=\linewidth]{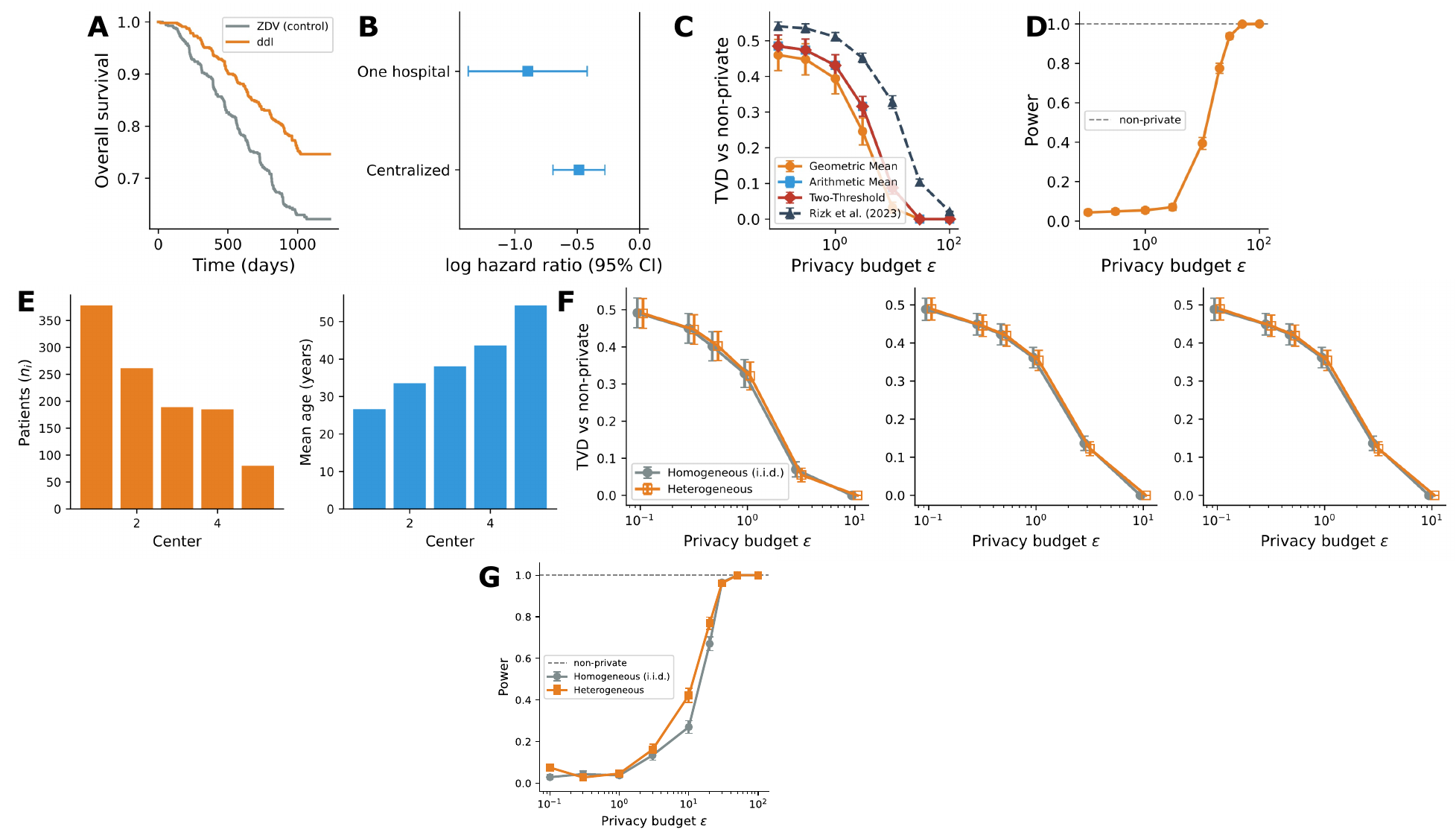}
    \caption{(A-D) Differentially private distributed survival analysis on ACTG~175, testing the ddI treatment effect for $n = 5$ fully connected centers. (A) Kaplan-Meier survival curves for the control (ZDV) and treatment (ddI) arms. (B) ddI log hazard ratio with $95\%$ confidence intervals, fitted centrally and at a single hospital. (C) Total variation distance between the private and non-private beliefs as a function of the privacy budget $\eps$ for AM, GM, and the two-threshold algorithm, together with the differentially private first-order method of \citet{rizk2023enforcing}. (D) Statistical Power of the GM as a function of the privacy budget $\eps$ at significance level $\alpha = 0.05$. (E-G) Robustness of differentially private distributed survival analysis to federation heterogeneity. (E) A non-i.i.d.\ partition incorporating empirically-grounded center-size variations and age-shifted case mixes across sites. (F) Total variation distance between the private algorithms (AM, GM, two-threshold) and the non-private algorithm as a function of the privacy budget $\eps$ under homogeneous vs.\ heterogeneous partitions. (G) Statistical power of the GM estimator under the homogeneous vs.\ heterogeneous partitions at a significance level $\alpha = 0.05$. We report $95\%$ confidence intervals taken over 500 repetitions for $T = 15, K = 2$.}
    \label{fig:clinical_trials}
    \vspace{-10pt}
\end{figure}

For the first simulation study, we focus on the AIDS Clinical Trials Group (ACTG) Study~175 dataset, with zidovudine (ZDV) as the standard care/control and didanosine (ddI) as the treatment. Throughout this section, we test the treatment effect while {adjusting} for baseline prognostic covariates (age, CD4 count, Karnofsky performance score, weight, prior antiretroviral exposure, race, and sex).

\noindent \textbf{Hypothesis testing.} We test whether ddI improves patient survival, that is, $\tilde\Theta_0: \theta = 0$ (no effect) against $\tilde\Theta_1: \theta \neq 0$ (treatment is effective), where $\theta$ is the treatment coefficient of the multivariate proportional hazards model. Because the covariates enter both models identically and cancel in the ratio, the released quantity is a scalar and is exchanged by the belief-propagation algorithms of \cref{sec:dp-algs} unchanged (see \cref{sec:app:clinical_trials} for the global sensitivity analysis and other implementation details). We use $n = 5$ fully connected hospitals, a significance level $\alpha = 0.05$, fitting the model with the open-source package \texttt{lifelines} \citep{davidson2019lifelines} and an $L_2$ penalty $\lambda_{\mathrm{reg}} |\theta|^2$ with $\lambda_{\mathrm{reg}} = 0.05$. The centralized analysis estimates a ddI log hazard ratio of $-0.49$ ($95\%$ CI $[-0.69, -0.28]$, $P < 10^{-5}$; \cref{fig:clinical_trials}B). In \cref{fig:clinical_trials}C and \cref{fig:clinical_trials}D, we observe that the three proposed algorithms (AM, GM, two-threshold) converge to the non-private benchmark as $\eps$ grows for $\eps \in [0.1, 10]$. %Moreover, the statistical power of the GM algorithm reaches 1 as $\eps$ approaches 100 and is above 0.8 for $\eps \approx 30$. 

\noindent \textbf{Runtime study.} In \cref{sec:app:timespacecomplexity}, we characterize the computational complexity of our algorithms, running in polynomial time and space. In \cref{sec:app:runtime_comparison}, we perform a runtime study on our proposed algorithms and show that our algorithm can run between $\sim 10^{-2}$ and $10$ seconds for values of $n$ ranging from $n = 3$ to $n = 96$, which is significantly faster (up to $1000\times$) than existing methods that rely on homomorphic encryption \citep{froelicher2021truly}. Fully homomorphic encryption and multiparty computation are cryptographic techniques that allow computation on encrypted data (by single or multiple servers, respectively) without the need to access decrypted data. The strong cryptographic guarantees of these methods come at a high cost to computational efficiency, and their implementation requires information technology (IT) infrastructure and large-scale, high-performance computational resources across distributed sites, which are beyond the technical capabilities of many organizations. 

\noindent \textbf{Comparison with first-order methods.} First-order distributed optimization methods with differential privacy guarantees offer an alternative approach \citep{rizk2023enforcing}, which, while more general, can be inefficient for distributed maximum likelihood estimation. \cref{fig:clinical_trials}C shows that our algorithms yield answers closer to ground truth compared to the first-order method of \citet{rizk2023enforcing} with significantly smaller errors (details in \cref{sec:app:firstordercomparison}). 

\noindent \textbf{Additional datasets and experiments.} In \cref{sec:app:additionalexperimentsCANCER}, we provide experiments with data from clinical trials in patients with advanced cancer, where the task is to determine whether certain biometric indices affect patient survival.

\noindent \textbf{Robustness to federation heterogeneity.} The adjusted analysis above splits the trial evenly across identical centers over a complete network. Real federated networks are heterogeneous, so we evaluate the method under a highly heterogeneous regime. We assign patients to $n = 5$ centers with unequal sizes drawn from real hospital bed counts \citep{ushospitals} and an age-shifted case mix where we control the mean age of the individuals at each center, maintaining both trial arms at every center (\cref{fig:clinical_trials}E). In \cref{fig:clinical_trials}F and \cref{fig:clinical_trials}G, we observe that the performance of the algorithm remains almost the same under the heterogeneous data split.

\subsection{Differentially Private Distributed Genetic Association Analysis}
\label{sec:gwas}

\begin{figure}[t]
    \centering
    \includegraphics[width=\linewidth]{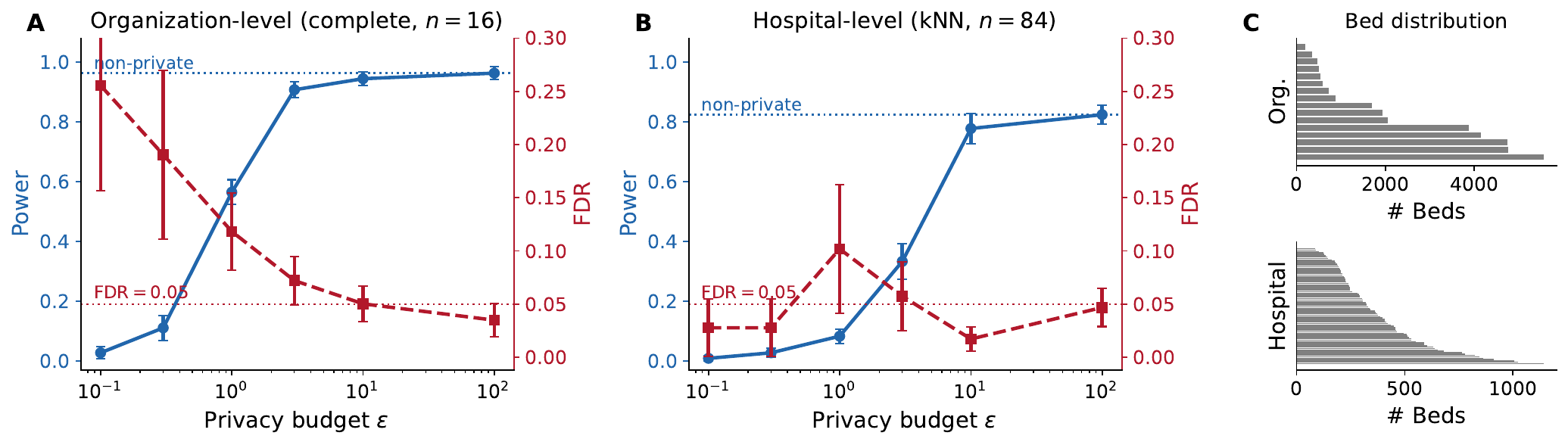}
    \caption{Privacy-utility curves measuring power, FDR (Type I error), for differentially private distributed genetic association analysis under two network architectures derived from the New York City hospital system with $N = 25,000$ individuals assigned according to hospital sizes shown in Panel (C). (A) Privacy-utility curve of the complete organizational network containing $n = 16$ organizations pertaining to the organizations that the hospitals in the New York City system correspond to. (B) Privacy-utility curve for the geographic hospital network containing $n = 84$ hospitals, where we have assumed that each hospital can exchange information with its 3-nearest hospitals. We observe that in the first case (panel A), the critical value of $\eps$, where the DP algorithm has sufficient power (approaching the centralized and the no-DP baselines), and low FDR (less than $5\%$) is $\eps$ being between 1 and 10, which is reasonable by widely-documented standards. In the latter case, the hospital network achieves much lower power showing that the organization level federation is better. We have used $T = 10$ and $K = 2$ and report 95\% confidence intervals over 3 runs.}
    \label{fig:gwas_hospitals}
\end{figure}

Genome-wide association studies exemplify both the necessity and the peril of data sharing. Detecting small per-variant effects demands cohorts larger than any single institution holds, motivating federated consortia. \citet{homer2008resolving} showed that an individual's membership in a study can be inferred from published allele frequencies alone, a result that prompted the withdrawal of aggregate GWAS statistics from public access \citep{zerhouni2008protecting} and still constrains data-sharing policy \citep{erlich2014routes}. Nor can leakage be contained by redacting sensitive loci: linkage disequilibrium allows masked genotypes to be reconstructed from neighboring markers, as when James Watson's withheld APOE status proved recoverable from the adjacent TOMM40 region \citep{nyholt2009jim}, ultimately forcing the removal of a 2 Megabase (Mb) window around the locus. In general, syntactic anonymization offers no guarantees against an adversary with side information or knowledge of the genome's correlation structure. Differential privacy addresses this gap by providing a formal and robust guarantee against arbitrary auxiliary information, which we adopt for federated genetic association analysis in this subsection. 

With DP in place, the operative managerial question is no longer whether to protect aggregate statistics, but at what level of institutional aggregation collaboration should occur. Should individual hospitals participate as independent nodes, each contributing their own privatized statistics, or should parent organizations first pool their member hospitals' data internally and participate as organizational-level nodes? This choice is typically settled through informal negotiation among legal, administrative, and scientific stakeholders, with little quantitative guidance. Our framework brings out the underlying quantitative tradeoffs.

As a case study, we conduct a large-scale regional study to evaluate our method in the context of distributed, differentially private genetic association analysis. 
Specifically, each simulation replicate consists of $N = 25{,}000$ individuals genotyped at a panel of 30 {independent} genomic regions (unlinked loci) of 200kb each, each region spanning roughly fifty variants simulated with the \texttt{msprime} package \citep{baumdicker2022efficient} under a coalescent model that mimics the linkage disequilibrium (LD) structure of European-ancestry samples, the reference population used in large-scale consortia such as the UK Biobank \citep{sudlow2015uk}. Six regions are associated and the remaining 24 are null. Each center summarizes every region by its SKAT statistic \citep{wu2011rare}, and the task is to detect the associated regions. Following real-world studies on Alzheimer's disease \citep{he2021genome}, we consider both rare (Minor Allele Frequency, $\text{MAF} < 0.01$) and common ($\text{MAF} \geq 0.01$) variants, with a realistic composition of 70\% rare and 30\% common variants.  The six associated regions are assigned a uniform range of per-region heritabilities ranging from 3\% to 22\%. This mirrors
the effect-size architecture of complex traits, whose heritability is carried disproportionately by a few larger-effect loci and spread across many weaker loci, which allows us to show the operating regime of our method. 

% Following large-scale AD association studies \citep{he2021genome}, we partition variants at a $1\%$ minor allele frequency threshold into rare ($\text{MAF} < 0.01$) and common ($\text{MAF} \geq 0.01$), with a $70\%/30\%$ rare/common composition which reflects a realistic allele-frequency spectrum of human genomes. Each associated region is calibrated to represent a strong-effect locus with several causal variants per region modeling allelic heterogeneity. We control FDR at the significance level $\alpha = 0.05$ via the Benjamini-Hochenberg proceedure. 
% \footnote{We note that this is a controlled evaluation of the DP mechanism. Detecting realistic, weak per-variant effects requires  $N$ in the hundreds of thousands.}.

Next, we construct a realistic network of federating centers from the publicly available US Hospital Locations dataset from \citet{ushospitals}. We retain all open, acute-care hospitals in New York State lying within the eight counties that comprise the New York City metropolitan area: Manhattan (New York County), Brooklyn (Kings), the Bronx, Queens, Staten Island (Richmond), Nassau, Westchester, and Suffolk. This yields 84 hospitals across 8 counties, totaling 32,989 beds. These hospitals belong to 16 unique organizations (independent hospitals grouped together). We consider two topologies that reflect our main question: \textit{(i)} an organizational-level complete network among the $n = 16$ organizations, assuming that the hospitals within each organization can pool their data together without the need for an agreement, and \textit{(ii)} a hospital-level geographical network among the $n = 84$ hospitals, in which each hospital is connected to the 3-nearest hospitals based on their geographical distance.  We operationalize heterogeneity in hospital size by assigning each hospital $i$ a simulated cohort of $n_i$ individuals proportional to its total bed count (see \cref{fig:gwas_hospitals}C). Here, the sensitivity $\Delta_{i, \Theta}$ grows polynomially with the number $M$ of SNPs, so we clip each individual's contribution to bound the sensitivity and add the appropriately calibrated Laplace noise (see \cref{subsec:main_result,subsec:pure_eps_skat}).

\cref{sec:app:gwas} presents the behavior of the method for different parameters such as the cohort size ($N$), the number of participating centers ($n$), and the network structure, and shows that our method is able to perform inference with a runtime on the order of seconds for a large number of hospitals, which is orders of magnitude smaller than homomorphic encryption methods and with substantially smaller error compared to first order methods. 

\noindent \textbf{Privacy-utility trade-offs.} \cref{fig:gwas_hospitals} compares the statistical power and the false discovery rate of the federated DP-GWAS algorithm under two network architectures derived
from the New York City hospital system. The organizational-level network achieves statistical power that approaches the centralized baseline, with the false discovery rate controlled below $5\%$ at a critical privacy budget of $\eps^*$ between 1 and 10. The hospital-level network, by contrast, requires a larger privacy budget and attains a lower power: it federates over many more centers, each of which injects Laplace noise and mixes more slowly over the sparse geographic network. To understand why, it is instructive to examine the privacy-utility curves for both networks as a function of $\eps$ (\cref{fig:gwas_hospitals}A and B).

On the one hand, for the organizational-level network, there exists a critical value of $\eps$ between 1 and 10 at which the algorithm transitions from the strong privacy regime ($\eps < \eps^*$), where the region statistics are dominated by noise and statistical power is lower, to a regime where power converges to the non-private baseline (\cref{fig:gwas_hospitals}A). On the other hand, for the hospital-level network, this critical $\eps^*$ is much larger, making it less practically deployable (\cref{fig:gwas_hospitals}B).

When $\eps$ is below $\eps^*$, the privatized statistics exchanged between centers are dominated by noise rather than by the true genetic signal, so few associated regions are above the significance threshold and statistical power is low. As $\eps$ increases above $\eps^*$, the signal-to-noise ratio improves and power increases toward the non-private centralized baseline at an FDR below the target level.

\section{Discussion}
\label{sec:discussion}

\noindent \textbf{Managerial Insights.} Our results carry concrete implications for decision-makers at institutions considering participation in distributed collaborative clinical research. Four main themes follow: 

\medskip

\noindent \emph{The critical privacy budget informs decision-making.} The privacy budget $\varepsilon$ can inform decision-making: organizations can offer stronger privacy guarantees (smaller $\eps$) to hesitant partners at a quantifiable and predictable cost to statistical power and communication rounds, making the privacy-utility trade-off tangible up to a critical level $\eps^*$. As shown in \cref{fig:gwas_hospitals}, operating below this critical $\eps^*$, the algorithm operates at a low signal-to-noise-ratio, yielding low power and high FDR. On the other hand, increasing $\eps$ above the critical threshold $\eps^*$ yields high power and low FDR at the expense of privacy. From our experiments, choosing  $\eps^*$ to be approximately between 1 and 10 for the organizational-level network represents a notion of a \emph{minimum deployable privacy budget}.

\medskip

\noindent \emph{Aggregation strategy and error tolerance.} The choice between the AM and GM estimators has a direct interpretation: Organizations in the early-stage screening contexts, where missing a true signal is the costlier error, should prefer the AM estimator, which controls Type-II error and maximizes recall. Those operating in regulatory approval or treatment selection contexts, where false positives carry
legal or clinical liability, should prefer the GM estimator, which controls Type-I error and ensures that only well-supported alternatives are selected. Finally, our two-threshold algorithm offers finer simultaneous control over both error types at a modest increase in communication complexity, and is preferable when neither error type is clearly dominant.

\medskip

\noindent \emph{Implications for network design.} The structure of the collaboration network (communication graph) has quantifiable consequences for statistical performance. Denser, more connected networks converge faster, require fewer communication rounds, and achieve a given accuracy at a smaller privacy budget, as reflected in the dependence of our complexity bounds on the second-largest eigenvalue modulus of the adjacency matrix. This reinforces the argument for
investing in broader inter-institutional partnerships compared to bilateral agreements. Our runtime comparisons further demonstrate that DP-based collaboration can be deployed with existing computational infrastructure and runs significantly faster than homomorphic encryption methods.

Further, the most actionable insight from our empirical evaluation concerns the unit at
which institutions should participate in a federated network. Comparing the
organizational-level and hospital-level networks derived
from the New York City hospital system reveals significant differences: the
organizational network achieves power that approaches the centralized baseline with an FDR below 5\% at the critical $\eps^*$, which is much lower than the hospital-level network. 

This divergence reflects three compounding mechanisms: First, individual hospitals hold smaller cohorts, weakening the per-node log-likelihood signal relative to the noise. Second, the larger number of nodes increases communication complexity and introduces more noise accumulation per round. Third, the sparser
adjacency matrix of the hospital network slows convergence, requiring more iterations to converge. Thus, the {fragmentation of the network into many small nodes degrades privacy-utility performance}, and the organizational-level architecture delivers both stronger privacy and better inference by improving the signal-to-noise ratio before any privatized information is shared externally. 

Therefore, we recommend a two-tier governance architecture in which hospitals first pool data within their parent organization under existing intra-organizational protocols, typically less burdensome than inter-organizational DUAs and not requiring patient-level re-consent in most jurisdictions, and only organizational nodes participate in the external federated network under formal DP guarantees. This architecture reduces communication complexity, strengthens the formal privacy guarantee offered to patients and regulators, and improves statistical power. In practice, large hospital systems already centrally govern data across their member hospitals and hold IRB approvals covering data use across member institutions, making the organizational node a natural and already existing unit of participation. When deciding how to structure participation, the decision rule is straightforward: if the pooled cohort size of the organizational node is large enough for the signal to dominate the DP noise at $\eps < \eps^*$, the organizational-level architecture should be adopted; otherwise, the institution should consider pooling with a peer organization under a bilateral DUA to form larger cohorts, or else admit a larger $\varepsilon$ (incurring privacy costs) or a reduced statistical power to maintain their current granular node-level participation.

\medskip

\noindent \textbf{Limitations and Future Work.} Several limitations of the present work merit discussion. First, our theoretical analysis assumes a fixed, known communication network with an irreducible doubly stochastic adjacency matrix; extending the framework to time-varying or randomly failing networks, which are common in practical distributed systems, remains an open problem. 

Second, communication complexity scales with the number of rounds $K$ and the state space size $|\Theta|$; for very small $\varepsilon$ or large $|\Theta|$, this can become prohibitive. Developing more communication-efficient aggregation strategies, for example, using the Dirichlet mechanism~\citep{gohari2021differential}, is an important direction for future work. 

Third, our finite-sample guarantees are developed for simple hypotheses, where each state in $\Theta$ indexes a single, fully specified likelihood. A natural and important next step is to extend this framework to more complex settings, where each hypothesis is a family of distributions, and the test statistics (e.g., generalized likelihood ratios) must be optimized over that family with appropriate sensitivity bounds for privacy calibration. 

Finally, our empirical evaluation focuses on survival and genetic association analyses; validating the framework across a broader range of distributed inference tasks (e.g., differential expression (DE) for biomarker discovery, best-treatment selection on binary or continuous endpoints, and diagnostic classification and disease subtyping)  would be promising next steps.

\noindent \textbf{Concluding Remarks.} Motivated by the need to share sensitive clinical data across institutions, we have studied distributed estimation and learning in networks of healthcare centers that collaborate without pooling raw patient records, yet still face privacy risks with respect to both their own data and the beliefs they exchange with their neighbors. %We have studied distributed estimation and learning in network environments where centers face privacy risks with respect to both their private data and the beliefs expressed by their neighbors. 
We proposed algorithms for distributed MLE and online learning that incorporate rigorous $\varepsilon$-DP guarantees through calibrated Laplace noise injection and derived finite-sample bounds on Type-I and Type-II error probabilities that characterize the three-way trade-off among privacy, communication complexity, and inference accuracy. Empirical validation on real-world clinical trial and synthetic genomic data  demonstrates that our framework achieves statistical performance that approaches the centralized baseline under practically deployable privacy budgets, at runtimes orders of magnitude faster than cryptographic baselines, and with lower error than existing first-order DP optimization methods. Beyond the
technical contributions, our analysis yields actionable guidance for the design of federated research networks, offering a principled quantitative basis for decisions about privacy budgets, aggregation strategies, network topology, and the level at which institutions should participate.
\clearpage
\ACKNOWLEDGMENT{
    M.P. was partially supported by a LinkedIn Ph.D. Fellowship, an Onassis Fellowship (ID: F ZT 056-1/2023-2024), and grants from the A.G. Leventis Foundation and the Gerondelis Foundation. The authors would like to thank the seminar participants at Rutgers Business School, Jalaj Upadhyay, Saeed Sharifi-Malvajerdi, Jon Kleinberg, Kate Donahue, Richa Rastogi, the attendants of the Information Systems Workshop at Arizona State University, Sue Brown, Maytal Saar-Tschechansky, and Vasilis Charisopoulos for their valuable discussions and feedback.

% Removed for anonymity. 
}

% References here
\bibliographystyle{informs2014}
\bibliography{references}

% Appendix here. The IJDS version ships this material as an e-companion
% (\ECSwitch); in the preprint it is a regular appendix so that everything
% stays in one document with continuous cross-references.
\begin{APPENDICES}
\crefalias{section}{appendix}
\crefalias{subsection}{appendix}
\crefalias{subsubsection}{appendix}

% ===================== Preprint appendix ToC =====================
% APPENDICES restarts the section counter at 1 (A, B, C, ...). That makes the
% appendix section anchors (section.0.1, ...) identical to the body's
% (Introduction = section.0.1, ...), so ToC/\cref links jump to the body and
% pdfTeX prints "destination ... already used". Give appendix sectioning its
% own hyperref anchors, exactly as the e-companion version does.
\renewcommand{\theHsection}{app.\arabic{section}}
\renewcommand{\theHsubsection}{app.\arabic{section}.\arabic{subsection}}
\renewcommand{\theHsubsubsection}{app.\arabic{section}.\arabic{subsection}.\arabic{subsubsection}}

% Route ONLY appendix \section (and \subsection) entries to a private .apptoc
% file, so the appendix ToC lists appendix headings only -- not the body.
\makeatletter
\let\PP@origaddcontentsline\addcontentsline
\renewcommand{\addcontentsline}[3]{%
  \def\PP@type{#2}%
  \def\PP@section{section}%
  \def\PP@subsection{subsection}%
  \ifx\PP@type\PP@section
    \PP@origaddcontentsline{apptoc}{#2}{#3}%
  \else\ifx\PP@type\PP@subsection
    \PP@origaddcontentsline{apptoc}{#2}{#3}%
  \fi\fi
}
% Print the appendix ToC (sections only, matching the journal EC ToC), with
% only the number hyperlinked.
\newcommand{\appendixtableofcontents}{%
  \section*{Appendix Contents}%
  \begingroup
    \hypersetup{linktoc=all}% make the whole entry (letter + title + page) clickable
    \setcounter{tocdepth}{1}%
    \renewcommand*\l@section{\@dottedtocline{1}{0em}{3em}}%
    \renewcommand*\l@subsection{\@dottedtocline{2}{3em}{3.6em}}%
    \@starttoc{apptoc}%
  \endgroup
}
\makeatother
% =================================================================

% --- Make \cref to subfigures show A.1(a) instead of 1(a) ---
\makeatletter
\renewcommand{\p@subfigure}{\thefigure}      % prefix for subfigure references
\renewcommand{\thesubfigure}{(\alph{subfigure})}
\makeatother

% Number appendix floats by appendix letter and restart them in each appendix.
\makeatletter
\@addtoreset{figure}{section}
\@addtoreset{table}{section}
\makeatother
\renewcommand{\thefigure}{\thesection.\arabic{figure}}
\renewcommand{\thetable}{\thesection.\arabic{table}}
\setcounter{figure}{0}
\setcounter{table}{0}

\newpage
\phantomsection
\setcounter{page}{1}
\renewcommand{\thepage}{A-\arabic{page}}
{\noindent \Large \bf Appendix}

\medskip

This appendix is organized as follows. First, \cref{sec:app:notation-table} presents a table of notations used throughout the paper, and \cref{sec:app:summary} summarizes the communication complexity of our proposed algorithms.  \cref{sec:app:normative-foudations} presents the normative foundations behind the log-linear belief exchange, and \cref{sec:app:learning-non-priv-mle} presents the non-private benchmark algorithm for MLE. Two of the private belief propagation algorithms are deferred to this appendix. We present our two-threshold algorithm for private distributed MLE in \cref{app:private_mle_two_threshold} and our algorithm for differentially private distributed online learning from intermittent streams in \cref{sec:app:dp-online-learning}. The convergence proofs for the private belief exchange algorithms are provided in \cref{sec:app:dist-MLE-convergence,sec:app:hypothesis_testing,sec:app:online-learning}. Finally, \cref{sec:app:clinical_trials} presents the sensitivity analysis and additional experiments supporting our distributed survival analysis for clinical trials, and \cref{sec:app:gwas} presents those supporting our distributed genetic association analysis.

\appendixtableofcontents

\clearpage

\section{Table of Notations}\label{sec:app:notation-table}

\begin{table}[htp]
    \centering
    \footnotesize
    \begin{tabular}{lp{11cm}}
    \toprule
       $x$ & scalar \\
       $\ovec x$ & vector \\
       $\vec x$ & element of vector $\ovec x$ \\
       $\| \ovec x \|_p$ & $\ell_p$-norm of $\ovec x$ \\ 
       $X$ & matrix \\
       $\ev{}{\cdot}$, $\var{}{\cdot}$ & expectation and variance operators \\
       \midrule
       $n$ & number of agents  \\
       $\eps$  & privacy budget \\
       $\alpha$, $1 - \beta$ & Type I, Type II error probabilities \\
       $\eta$ & $\eta= \max \{ \alpha, 1-\beta \}$ \\
       $A$ & communication matrix \\
       $\Theta$ & set of states \\
       $\Theta^\star$, $\compTheta$ & set of MLE, non-MLE states \\
       $\theta^\star$, $\bar \theta$, $\theta^{\circ}$ & MLE, non-MLE/false state, true state (in the online learning case) \\ 
       $f^\star$ & density of MLE states ($f^\star = |\Theta^\star| / |\Theta|$) \\
       $\ell_i(\vec s | \hat \theta)$ & likelihood of agent $i$ given dataset $\vec s$ at state $\hat \theta$ \\
       \midrule
       $T$, $K$ & number of iterations, number of rounds \\
       $\vec \gamma_{i, t}(\hat \theta)$  & joint likelihood of agent $i$ given data $\vec s_{i, t}$ of size $n_{i, t}$ at state $\hat \theta$ in the non-private regime \\  
       $\vec \lambda_{i, t}(\hat \theta, \check \theta)$ & joint log likelihood ratio of agent $i$ at iteration $t$ between states $\hat \theta$ and $\check \theta$ in the non-private regime \\ 
       $\Lambda(\hat \theta, \check \theta)$ & joint log likelihood ratio between states $\hat \theta$ and $\check \theta$ over all agents and data in the non-private regime \\
       $\vec \sigma_{i, t}(\hat \theta)$  & joint likelihood of agent $i$ given data $\vec s_{i, t}$ of $n_{i, t}$ patients $\hat \theta$ in the private regime \\  
      $\vec \zeta_{i, k, t}(\hat \theta, \check \theta)$ & joint log likelihood ratio of agent $i$ at iteration $t$ at round $k$ between states $\hat \theta$ and $\check \theta$ in the private regime \\ 
       $Z_k(\hat \theta, \check \theta)$ & joint log likelihood ratio between states $\hat \theta$ and $\check \theta$ at round $k$ over all agents and data in the private regime \\
       $\vec \mu_{i, t}(\hat \theta)$ & belief of agent $i$ at iteration $t$ on state $\hat \theta$ in the non-private regime \\
       $\vec \nu_{i, k, t}(\hat \theta)$ & belief of agent $i$ at iteration $t$ at round $k$ on state $\hat \theta$ in the private regime \\
       $\vec \phi_{i, t}(\hat \theta, \check \theta)$ & log belief ratio of agent $i$ at iteration $t$ on pair of states $\hat \theta$ and $\check \theta$ in the non-private regime \\
       $\vec \psi_{i, k. t}(\hat \theta, \check \theta)$ &  log belief ratio of agent $i$ at iteration $t$ at round $k$ on pair of states $\hat \theta$ and $\check \theta$ in the private regime \\
       $\vec d_{i, t}(\hat \theta)$, $\vec d_{i, k}(\hat \theta)$ &noise variable for agent $i$, state $\hat \theta$, and iteration $t$, or round $k$ \\ 
       $\cD_{i, t}(\hat \theta; \eps)$, $\cD_{i, k}(\hat \theta; \eps)$ & noise distribution for agent $i$, state $\hat \theta$, and iteration $t$, or round $k$ \\
        \midrule
        $\vec \delta_{ij}$ & censoring indicator for patient $j$ in center $i$ \\
       $\vec x_{ij}$ & treatment indicator for patient $j$ in center $i$ \\
       $\vec t_{ij}$ & survival or censoring time in days for patient $j$ in center $i$ \\
       \midrule
       $b_n^\star = \left | \lambda_2(A) \right |$ & second-largest eigenvalue modulus of $A$ \\
       $a_n^\star$ & second largest eigenvalue modulus of $(A + I)/2$, also denoted by $|\lambda_2(A + I)|/2$ \\
       \midrule
       $\Gamma_{n, \Theta}$ & largest likelihood magnitude \\
       $\Delta_{i, \Theta}$ & global sensitivity of centrer $i$ \\
       $\Delta_\Theta^n$ & maximum of the global sensitivities; $\max_{i \in [n]} \Delta_{i, \Theta}$ \\
       $l_{n, \Theta}$ & minimum KL-divergence between non-MLE and MLE states (false and true states in the OL case) \\
       $V_{n, \eps, \Theta}$ & sum of standard deviations of the noises over all agents \\ 
       $Q_{n, \Theta}$ & maximum standard deviation of KL-divergence for any agent and any pair of states \\

    \bottomrule
    \end{tabular}
    % \caption{Notation table.}
    % \label{tab:notation}
\end{table}

\clearpage

\section{Summary Table for the Communication Complexity of MLE Algorithms} \label{sec:app:summary}

\begin{table*}[!h]
\centering
    \scriptsize
    \begin{tabular}{lcc}
    \toprule
    Setting & With DP (ours) & Without DP~\citep{rahimian2016distributed} \\ 
    \midrule
       & \multicolumn{2}{c}{Communication Complexity with respect to $\eps, \eta$} \\
    \midrule 
    AM/GM & $\Opriv {{\color{belize}{\log (1 / \eta) (\log (1 / \eta) + \log (1 / \eps) + \log (1 / \varrho)}}}$ & $\Opriv { {1} }$ \\
    Two-Threshold & $\Opriv {{\color{belize}{ \frac {\log (1 / \eta)} {\pi^2} (\log (1 / \eta) + \log (1 / \eps) + \log (1 / q))}}}$ & $\Opriv { {1} }$ \\
    % Online Learning &  $\Opriv {{\color{belize}{\frac 1 {\eta^{3/2}} {\frac 1 {\eps}}}}}$ & $\Opriv { {1} }$ \\
    \toprule
       & \multicolumn{2}{c}{Communication Complexity with respect to $n$} \\
    \midrule 
    AM/GM & $\Onn {\max \left \{ \log \left ( \frac {n} {l_{n, \Theta}} \right ),  \frac {\log \left ( \frac {n (\Gamma_{n, \Theta} + {\color{belize}{\Delta_\Theta^n}})} {{\color{belize}{\varrho}}}  \right )} {\log (1 / a_n^\star)} \right \}}$ & $\cO_{|\Theta|} \left ( \max \left \{ \log \left ( \frac {n} {l_{n, \Theta}} \right ),  \frac {\log \left ( 
    {n} {\Gamma_{n, \Theta}} \right )} {\log (1 / a_n^\star)} \right \} \right )$ \\
    Two-Threshold & $\Onn {{\color{belize}{\frac {1} {\pi^2}}}  \max \left \{ \log \left ( \frac {n} {l_{n, \Theta}} \right ),  \frac {\log \left ( \frac {n (\Gamma_{n, \Theta} + {\color{belize}{\Delta_\Theta^n}})} {{\color{belize}{q}}}  \right )} {\log (1 / a_n^\star)} \right \}}$ & $\cO_{|\Theta|} \left ( \max \left \{ \log \left ( \frac {n} {l_{n, \Theta}} \right ),  \frac {\log \left ( 
    {n} {\Gamma_{n, \Theta}} \right )} {\log (1 /  a_n^\star)} \right \} \right )$ \\

    % Online Learning & $\Onn {\frac {n(Q_{n, \Theta} + {\color{belize}{\Delta_\Theta^n}}) } {l_{n, \Theta} (1 - b_n^\star)}}$ & $\cO_{|\Theta|} \left ( \frac {nQ_{n, \Theta} } {l_{n, \Theta} (1 - b_n^\star)} \right )$ \\
    \midrule
        & \multicolumn{2}{c}{Communication Complexity with respect to $|\Theta|$} \\
    \midrule
    AM/GM & $\Otheta {{\color{belize}{|\Theta| \log |\Theta|}} \left ( {\color{belize}{\log |\Theta|}} + \log \left ( \frac {\Gamma_{n, \Theta} + {\color{belize}{|\Theta| \log |\Theta| \Delta_\Theta^n}}} {{\color{belize}{\varrho}}} \right )\right )}$ & $\cO_n \left ( {\log \Gamma_{n, \Theta}} \right )$ \\
    Two-Threshold & $\Otheta {{\color{belize}{\frac {\log |\Theta|} {\pi^2}}} \left ( {\color{belize}{\log |\Theta|}} + \log \left ( \frac {\Gamma_{n, \Theta} + {\color{belize}{|\Theta| \log |\Theta| \Delta_\Theta^n}}} {{\color{belize}{q}}} \right ) \right )}$ & $\cO_n \left ( {\log \Gamma_{n, \Theta}} \right )$ \\

    % Online Learning & $\Otheta {\frac {|\Theta| (Q_{n,\Theta} + {\color{belize}{|\Theta| \Delta_\Theta^n}})} {l_{n, \Theta}}}$ &  $\cO_n \left ( \frac {|\Theta| Q_{n,\Theta}} {l_{n, \Theta}} \right )$ \\
    \bottomrule
 
    \end{tabular}
    
    \caption{\footnotesize{Communication complexity of distributed inference algorithms for a fixed privacy budget $\eps > 0$ and target maximum Type I/Type II error $\eta \in (0, 1)$. We use $\cO_{a_1, \dots, a_N} \left ( \cdot \right )$  to denote the big-$\cO$ notation where the constants are allowed to depend on $a_1, \dots, a_N$. The overhead of introducing privacy is denoted in {\color{belize}{blue}}. For compactness in notation, $\eta$ refers to $\max \{ \alpha,  1 - \beta \}$ and $\varrho$ refers to $\min \{ \varrho^\AM, \varrho^\GM \}$ (see \cref{theorem:non_asymptotic_private_distributed_mle} for closed-form lower bounds with explicit dependence on the AM and GM thresholds $\varrho^\AM$ and $\varrho^\GM$ and error probabilities $\alpha$ and $1 - \beta $). In the case of the two-threshold algorithm, $\pi$ refers to $\min \{ \pi_1, \pi_2 \}$, and $q$ refers to $\min \{q_1, 1 - q_2  \}$ (see \cref{theorem:non_asymptotic_threshold} and \cref{{alg:threshold}} for details of how the two thresholds are determined in terms of parameters $\pi_1, \pi_2$ and $q_1, q_2$). The AM/GM algorithms have matching communication complexities with the two-threshold algorithm whenever $q = \cO(1)$ and $\pi = \cO \left (1 / \sqrt {|\Theta|} \right )$. The non-DP results are due to \citet{rahimian2016distributed}.}} \label{tab:summary}
\end{table*}

\clearpage

\section{Differentially Private Log-Linear Belief Updates and Opinion Pools} \label{sec:app:normative-foudations}

The fastest way to (asymptotically) compute the joint log-likelihood would be for the agents to run a linear consensus algorithm \citep{DeGrootModel,olshevsky2017linear} in the (privatized) log-likelihood space for each $\theta \in \Theta$. This rule is equivalent to performing log-linear updates (i.e., geometric means) in a (non-Bayesian) belief space \citep{rahimian2016distributed}. On the other hand, simple DeGroot (linear) averaging of beliefs converges to the average of initial likelihoods (rather than log-likelihoods) and can give biased estimates if used for maximum likelihood estimation. \citet{kayaalp2023arithmetic} and \citet[Corollary 2]{lalitha2018social} point out that log-linear learning converges faster than linear learning in the belief space. \citet{marden2012revisiting} provide guarantees on the percentage of time that the action profile will be a potential maximizer in potential games using log-linear best-response dynamics. 

The $\log$-linear updates that we use to exchange belief statistics among neighboring agents also have parallels as opinion pools to combine probability distributions from multiple experts in the decision and risk analysis literature \citep{clemen1999combining}. \citet{abbas2009kullback}, in particular, proposes scoring rules based on the KL divergence between the expert distributions and aggregate beliefs that yield linear and $\log$-linear aggregates as the solution to the expert aggregation problem: If KL divergence is computed from the aggregate belief to the expert beliefs then we obtain log-linear updates, otherwise (i.e., if the KL divergence is computed from the expert beliefs to the aggregate belief) we obtain linear updates. Denoting expert beliefs by $\vec \nu_j$ with weights $a_{ij}, j \neq i$, the $\log$-linear belief aggregate $\nu^\star$ sets the belief in every state $\vec \nu(\theta)$ proportionally to $\prod_{j\in[n]}\vec \nu_j^{a_{ij}}(\theta)$ and is the solution to $\arg\min\sum_{j\neq i} a_{ij}\kldiv {\vec \nu_{i}} {\vec \nu_{j}}$. Such rules are especially sensitive to experts who put zero or small beliefs on some states and can be useful for rejecting non-MLE states or null hypotheses. In this section, we use the framework of \citet{abbas2009kullback} with explicitly modeled privacy costs to justify our update rules beyond the algorithmic performance framework of \Cref{sec:performance-metrics} and our theoretical guarantees in \Cref{sec:dp-algs}. 

Following \citet{abbas2009kullback}, consider a decision maker ($i$) who is interested in choosing the best option among a set of alternatives ($\Theta$), given access to their own private information ($\vec s_i$) and other opinions from $n$ experts. The decision maker forms a time-varying public belief $\vec \nu_{i, t} \in \mathrm {Simplex}(\Theta)$ over $\Theta$ for treatments that capture all expert distributions and its own data. Initially (at $t = 0$),  the agent forms an intrinsic belief $\vec \gamma_i \in \mathrm {Simplex}(\Theta)$ based on its own data, i.e., $\vec \gamma_i(\hat \theta) = \frac {\ell_i(\vec s_i | \hat \theta)} {\sum_{\bar \theta \in \Theta} \ell_i(\vec s_i | \bar \theta)}$ for all $\hat \theta \in \Theta$. In subsequent steps ($t \ge 1$), the decision maker chooses an expert at random with probability $a_{ij}$ corresponding to the strength of the connection in the communication matrix $A$ and pays $C_i$ to consult with them and $C_i$ to consult its own data and incurs a revenue $R_i$ regardless of its decision. Since the decision maker consults its own data and communicates its beliefs with others, it must protect its own datasets due to privacy risks.

The $\eps$-DP mechanism $\cM_i$, satisfying \Cref{def:DP}, applied to $\vec \gamma_i$ produces $\cM_i(\vec \gamma_i)$ which is the privatized belief over the state space $|\Theta|$.  Putting everything together, the expected payoff of agent $i$ at time $t$ is given as (see Eq. (9) in \citet{abbas2009kullback}): 
{
\begin{equation} \label{eq:utility_dataset_dp}
    U_{i, t}(\vec \nu_{i, t}, \ovec \nu_{t-1}) = \begin{cases}
         {R_i - C_i \sum_{j = 1}^n a_{ij} \kldiv {\vec \nu_{i, t}} {\vec \nu_{j, t-1}} - C_i \kldiv {\vec \nu_{i, t}} {\vec \nu_{i, t - 1}}}, & t \ge 1 \\
         {R_i - C_i \kldiv {\vec \nu_{i, 0}} {\cM_i(\vec \gamma_i)}}, & t = 0.
    \end{cases}
\end{equation}}

To implement the mechanism $\cM_i$, the decision maker draws appropriately chosen, without loss of generality, zero-mean noise variables $\vec d_i(\hat \theta)$ for each state $\hat \theta \in \Theta$ and adds them to the corresponding log-likelihoods $\log \vec \gamma_i(\hat \theta)$. Thus, $\cM_i(\vec \gamma_i)$ can be written as the product of the two distributions $\vec \gamma_i$ and $\mathfrak D_i$, where $\mathfrak D_i(\hat \theta) = \frac {e^{\vec d_i(\hat \theta)}} {\sum_{\bar \theta \in \Theta} e^{\vec d_i(\bar \theta)}}$. If we write $\vec \nu_{i, 0}$ as the product of $\vec \nu_{i, 0}$ and the uniform distribution $\cU_\Theta$ on $\Theta$, then we can use the properties of the KL divergence to write $\kldiv {\vec \nu_{i, 0}} {\cM_i(\vec \gamma_i)} = \kldiv {\vec \nu_{i, 0}} {\vec \gamma_i} + \kldiv  {\cU_\Theta} {\mathfrak D_i}$. The term $\kldiv  {\cU_\Theta} {\mathfrak D_i}$ corresponds to the privacy loss for the utility in \cref{eq:utility_dataset_dp} and equals
{
\begin{equation*}
    \kldiv  {\cU_\Theta} {\mathfrak D_i} = \sum_{\hat \theta \in \Theta} \frac {1} {|\Theta|} \log \left ( \frac {\sum_{\bar \theta \in \Theta} e^{\vec d_i(\bar \theta)}} {|\Theta | e^{\vec d_i(\hat \theta)}}  \right ) = \log \left ( {\sum_{\bar \theta \in \Theta} e^{\vec d_i(\bar \theta)}} \right ) - \log |\Theta| - \sum_{\hat \theta \in \Theta} \vec d_i(\hat \theta).
\end{equation*}
}

 Taking the expectation over $\cM_i$ we get $\ev {\cM_i} {\kldiv  {\cU_\Theta} {\mathfrak D_i}} = \ev {\cM_i} {\log \left ( {\sum_{\bar \theta \in \Theta} e^{\vec d_i(\bar \theta)}} \right )} - \log |\Theta|$. Since the KL divergence is nonnegative, the expected privacy loss with respect to $\cM_i$ is minimized whenever the noise variables $\vec d_i(\hat \theta)$ are independent of each other and their distribution does not depend on $\hat \theta$, i.e., $\vec d_i(\hat \theta) = \vec d_i$. 

To satisfy the $\eps$-DP requirement in \cref{def:DP}, the noise $\vec d_i$ is set according to the Laplace mechanism with scale ${|\Theta|\Delta_\Theta^n}/{\eps}$ and the per-state budget allocation detailed in \cref{sec:performance-metrics}.

In \cref{sec:app:sensitivity}, we bound the sensitivity \eqref{eq:sensitivity} for the proportional hazards model. More generally, the dataset structure may be such that the sensitivity is unbounded when computed globally over the dataset space. In such cases, an additive relaxation of the privacy constraints in \cref{def:DP} can be used with a smoothed notion that looks at sensitivity locally at the realized dataset values \citep{nissim2007smooth}. The Laplace mechanism possesses further benefits since it minimizes the convergence time of the belief exchange algorithms (cf. \cref{sec:dp-algs} and prior results by \citet{papachristou2023differentially} and \citet{koufogiannis2015optimality}). The best update rule that optimizes the time-varying utility of \cref{eq:utility_dataset_dp} yields the log-linear update rules \citep[Proposition 4]{abbas2009kullback}.

The above can be extended to the online setting where data $\vec s_{i, t}$ with $n_{i, t} \ge 0$ data points for each agent $i$ are intermittently arriving at each iteration $t$, by introducing a time-varying utility and measuring the likelihood at each iteration $t$. In that regime, first, the agents form the probability measure $\vec \gamma_{i, t}$ such that $\vec \gamma_{i, t}(\hat \theta) = \frac {\ell_i(\vec s_{i, t} | \hat \theta)} {\sum_{\bar \theta \in \Theta} \ell_i(\vec s_{i, t} | \bar \theta)}$  for all $\hat \theta \in \Theta$ (when $n_{i, t} = 0$ we set $\ell_{i}(\vec s_{i, t} | \hat \theta) = 1$ for all $\hat \theta \in \Theta$). Then the agents have the time-varying utility: 
{
\begin{equation*} U_{i, t}(\vec \nu_{i, t}, \ovec \nu_{t - 1})  = {R_i - C_i \sum_{j = 1}^n a_{ij} \kldiv {\vec \nu_{i, t}} {\vec \nu_{j, t - 1}} - C_i \kldiv {\vec \nu_{i, t}} {\cM_i(\vec \gamma_{i, t})}}.
\end{equation*}}
Optimizing this utility yields log-linear learning, which we leverage in our OL \cref{alg:private_online_learning}. 

\clearpage

% \section{Benchmark Algorithms for Non-Private Belief Propagation}\label{sec:app:learning-non-priv}

\section{Non-private Benchmark for Distributed MLE}\label{sec:app:learning-non-priv-mle}

In the non-private regime, agents form beliefs $\vec \mu_{i, t}(\hat \theta)$ over the states $\hat{\theta}\in\Theta$ at every iteration $t$ and exchange their beliefs with their neighbors until they can detect the MLE. \citet{rahimian2016distributed} give \cref{alg:non_private_mle} for belief exchange and prove that beliefs converge to a uniform distribution over the MLE set (see Theorem 1 in their paper).

\begin{algorithm}[h]
\captionsetup{font=footnotesize}
\caption{Non-Private Distributed MLE} \label{alg:non_private_mle}
\footnotesize
\begin{flushleft}
The agents begin by forming: $ \vec\gamma_{i}(\hat{\theta}) = \ell_i(\vec{s}|\hat{\theta})$, and initializing their beliefs to ${\vec\mu}_{i,0}(\hat{\theta})$ $ =$ $  \vec\gamma_{i}(\hat{\theta}) /\sum_{\tilde{\theta}\in\Theta} \vec\gamma_{i}(\tilde{\theta})$.   In any future time period, the agents update their belief after communication with their neighboring agents, and according to the following update rule:  
\begin{equation*}
\hspace{-10pt}{\vec\mu}_{i,t}(\hat{\theta})
=  \frac{{\vec\mu}^{1+a_{ii}}_{i,t-1}(\hat{\theta}) \prod\limits_{j\in\mathcal{N}_i }{{\vec\mu}^{a_{ij}}_{j,t-1}(\hat{\theta})} }{\sum\limits_{\tilde{\theta} \in \Theta}{\vec\mu}_{i,t-1}^{1+a_{ii}}(\tilde{\theta}) \prod\limits_{j\in\mathcal{N}_i} {{\vec\mu}^{a_{ij}}_{j,t-1}(\tilde{\theta})} }, \text{for all $\hat \theta \in \Theta$ and $t \in [T]$.}  \end{equation*} 
\end{flushleft}
\end{algorithm}

The convergence of this algorithm relies on studying the behavior of the log-belief ratio between two states $\hat \theta$ and $\check \theta$, whose dynamics are governed by the powers of the primitive matrix $A + I$, see \citep[Theorem 1]{rahimian2016distributed}, with modulus-ordered eigenvalues $0 < |\lambda_{n} (A+ I)| \le \dots \le |\lambda_{2} (A + I)| < \lambda_1(A + I) = 2$. Specifically, \citet[Theorem 1]{rahimian2016distributed} state that there is a dominant term due to the maximum eigenvalue $\lambda_1(A + I) = 2$ that dominates the log-belief ratios for large $t$, and $n - 1$ terms that decay exponentially with rate $(|\lambda_{2} (A + I)|/2)^t$. The dominant term depends on the difference of the log-likelihoods between the states $\hat \theta$ and $\check \theta$, namely $\Lambda(\hat \theta, \check \theta) = \Lambda(\hat \theta) - \Lambda(\check \theta)$. Subsequently, the log-belief ratio approaches $- \infty$ as $t \to \infty$ whenever $\check \theta \in \Theta^\star$, and we can recover the MLEs.

%\section{Private Belief Propagation Algorithms}\label{sec:app:learning-priv}

\section{The Two-Threshold Algorithm for Private Distributed MLE} \label{app:private_mle_two_threshold}

In \cref{alg:threshold}, we provide a two threshold algorithm to simultaneously control both Type I and Type II error probabilities, with more design flexibility that comes at an increased cost of communication complexity.   The analysis of \cref{theorem:asymptotic_private_distributed_mle}  shows that in a given round $k$, the probability that an MLE state $\theta^\star \in \Theta$ ends up positive as $T \to \infty$ is at least $1 / |\compTheta|$, so in expectation at least $K / |\compTheta|$ rounds will yield a positive belief. On the other hand, we know that for a non-MLE state $\bar \theta$, in expectation, at most $(1 - 1 / |\Theta^\star|)K$ trials will come up heads. For brevity, we define $p_1 = 1 - 1 / |\Theta^\star|$ and $p_2 = 1 / |\compTheta|$. Moreover, we define $\vec N_{i, T}^{\thres, 1}(\hat \theta)$ (resp. $\vec N_{i, T}^{\thres, 2}(\hat \theta)$) as the (average) number of times the belief $\vec \nu_{i, k, t}(\hat \theta)$ exceeds a threshold $\hattauthres{1}$ (resp. $\hattauthres{2}$) for $k \in [K]$.  

Because $\vec N_{i, T}^{\thres, 1}(\hat \theta)$ (resp. $\vec N_{i, T}^{\thres, 2}(\hat \theta)$) is an average of independent indicator variables for all $\hat \theta \in \Theta$, the Chernoff bound indicates that it concentrates around its mean $\ev {} {\vec N_{i, T}^{\thres, 1} (\hat \theta)}$ (resp. $\ev {} {\vec N_{i, T}^{\thres, 2} (\hat \theta)}$). This prompts the development of the following simple algorithm, which resembles boosting algorithms such as AdaBoost \citep{freund1997decision}. The algorithm uses two thresholds $\tauthres{1}$ and $\tauthres{2}$ to estimate the MLE states as those whose beliefs exceed $\hattauthres{1}$ and $\hattauthres{2}$ at least $\tauthres{1}$ and $\tauthres{2}$ times, leading to output sets $\hat \Theta^{\thres, 1}_{i, T}$ and $\hat \Theta^{\thres, 2}_{i, T}$, respectively. 

We first present an asymptotic result (proved in \cref{sec:app:proof:theorem:asymptotic_threshold}): 

\begin{theorem} \label{theorem:asymptotic_threshold}
    Consider \cref{alg:threshold} run with state-independent noise distributions $\cD_{i} (\hat \theta; \eps) = \cD_i (\eps)$ that satisfy $\eps$-DP, and set $p_1 = 1 - 1 / |\Theta^\star|$ and $p_2 = 1 / |\compTheta|$. Then, for every center $i \in [n]$, as $\varrho^{\thres, 1} \to \infty$ and $\varrho^{\thres, 2} \to \infty$:
    
    \begin{itemize}
        \item for any $\pi_1 > 0$, with $\tauthres{1} = (1 + \pi_1) p_1$ and $K \ge \frac {\log (|\compTheta| / \alpha)} {2 \pi_1^2}$, we have $\lim_{T \to \infty} \Pr \left [ \hat \Theta^{\thres, 1}_{i, T} \subseteq \Theta^\star \right ] \ge 1 - \alpha$;
        \item  for any $\pi_2 > 0$, with $\tauthres{2} = (1 - \pi_2) p_2$ and $K \ge \frac {\log (|\Theta^\star| / (1 - \beta))} {2 \pi_2^2}$, we have $\lim_{T \to \infty} \Pr \left [ \Theta^\star \subseteq \hat \Theta^{\thres, 2}_{i, T} \right ] \ge \beta$.
    \end{itemize}
\end{theorem}

\pfsketch{The proof is very similar to \cref{theorem:asymptotic_private_distributed_mle}, with the only difference that instead of relying on independent events, we use the Chernoff bound on $\vec N_{i, T}^{\thres, 1}(\hat \theta)$ (resp. $\vec N_{i, T}^{\thres, 2}(\hat \theta)$).}

Similarly to \cref{theorem:asymptotic_private_distributed_mle}, the estimator $\hat \Theta^{1, \thres}_{i, T}$ has a low Type I error, and the estimator $\hat \Theta^{2, \thres}_{i, T}$ has a low Type II error. If agents do not know $|\Theta^\star|$, as long as they choose thresholds ${\tauthres{1}}' \geq (1 + \pi_1) p_1 = \tauthres{1}$ and ${\tauthres{2}}'  \le  (1 - \pi_2) p_2 = \tauthres{2}$, they can obtain the same guarantees. For example, one can choose the following upper and lower bounds to set the thresholds: ${\tauthres{1}}'  =  (1 + \pi_1) (1-1/|\Theta|) \geq (1 + \pi_1) p_1$ and ${\tauthres{2}}' = (1 - \pi_2) ( 1/|\Theta|)  \le  (1 - \pi_2) p_2$. Also, running the algorithm for $K \ge \max \left \{ \frac {\log (|\Theta| / \alpha)} {2 \pi_1^2} \mathpunct{\raisebox{0.5ex}{,}} \frac {\log (|\Theta| / (1 - \beta))} {2 \pi_2^2} \right \}$ we can guarantee that as $T \to \infty$, we have $\Theta^{\thres, 2}_{i, T} \subseteq \Theta^\star \subseteq \Theta^{\thres, 1}_{i, T}$ with probability at least $\beta - \alpha$.  

Since the counts for MLE and non-MLE states concentrate around means separated by $p_2 - p_1$, perfect detection of $\Theta^\star$ is achievable when the two modes are sufficiently separated, i.e., $p_2 > p_1$, as the following corollary shows:

\begin{corollary}[Exact Recovery with a Single Threshold] \label{corollary:separation_threshold} 
    If the density of MLE states $f^\star$ satisfies: 
    {\begin{equation*} \begin{cases}
                0 \le f^\star \le 1, & \text{if } 1 \le |\Theta| \le 4 \\
                0 \le f^\star < \frac 1 2 - \frac 1 2 \sqrt {\frac {|\Theta| - 4} {|\Theta|}} \quad \vee \quad \frac 1 2 + \frac  1 2 \sqrt {\frac {|\Theta| - 4} {|\Theta|}} < f^\star \le 1, & \text{otherwise}
            \end{cases},
    \end{equation*}} then setting the thresholds equal, $\tauthres{2} = \tauthres{1}$ and $\hattauthres{1} = \hattauthres{2}$ (so that $\hat \Theta^{\thres, 2}_{i, T} = \hat \Theta^{\thres, 1}_{i, T}$), with $\pi_1 = (1 - \pi_2) \frac {p_2} {p_1} - 1$ for some $\pi_2 > 0$ and $K \ge \max \left \{ \frac {\log (|\Theta| / \alpha)} {2 \pi_1^2} , \frac {\log (|\Theta| / (1 - \beta))} {2 \pi_2^2} \right \}$, yields $\lim_{T \to \infty} \Pr \left [ \hat \Theta^{\thres, 2}_{i, T} = \Theta^\star \right ] \ge \beta - \alpha$ for every center $i \in [n]$.
\end{corollary}

\pfsketch{The range of $f^\star$ is derived by solving the inequality $\frac {1} {(1 - f^\star) |\Theta|} > 1 - \frac {1} {f^\star |\Theta|}$, and the rest follows by applying \cref{theorem:asymptotic_threshold}.}

By performing an analysis similar to \cref{theorem:non_asymptotic_private_distributed_mle}, we derive a non-asymptotic result (proved in \cref{sec:app:proof:theorem:non_asymptotic_threshold}): 

\begin{theorem} \label{theorem:non_asymptotic_threshold}

    Let $q_1 , q_2 \in (0, 1)$ and $\varrho^{\thres, 1}, \varrho^{\thres, 2} > 0$. Then, for \cref{alg:threshold}, the following hold for every center $i \in [n]$: \begin{itemize}
        \item For any $\pi_1 > 0$, if $\tauthres{1} = (1 + \pi_1) q_1$, {\footnotesize$T \ge \max \left \{ \frac {\log \left ( \frac {2 \varrho^{\thres, 1} n} {l_{n, \Theta}} \right )} {\log 2}, \frac {\log \left ( \frac {|\Theta|^2 (n-1) (n\Gamma_{n, \Theta} + V_{n, \eps, \Theta})} {2 q_1 \varrho^{\thres, 1}} \right )} {\log (1 / a_n^\star)}\right \}$}, and $K \ge \frac {\log (|\compTheta| / \alpha)} {2 \pi_1^2}$, then $\Pr \left [ \hat \Theta^{\thres, 1}_{i, T} \subseteq \Theta^\star \right ] \ge 1 - \alpha$.
        \item For any $\pi_2 > 0$, if $\tauthres{2} = (1 - \pi_2) q_2$, {\footnotesize $T \ge\max \left \{ \frac {\log \left ( \frac {2 \varrho^{\thres, 2} n} {l_{n, \Theta}} \right )} {\log 2}, \frac {\log \left ( \frac {|\Theta|^2 (n-1) (n\Gamma_{n, \Theta} + V_{n, \eps, \Theta})} {2 (1 - q_2) \varrho^{\thres, 2}}  \right )} {\log (1 /  a_n^\star)} \right \}$}, and $K \ge \frac {\log (|\Theta^\star| / (1 - \beta))} {2 \pi_2^2}$, then $\Pr \left [ \Theta^\star \subseteq \hat \Theta^{\thres, 2}_{i, T} \right ] \ge \beta$.
        \item    The noise distributions that minimize the convergence time $T$ for both estimators are the Laplace distributions $\cD_i^\star(\eps) = \mathrm{Lap} \left ( {\Delta_{i, \Theta} K |\Theta|}/{\eps} \right )$.
    \end{itemize}
    
\end{theorem}

\pfsketch{The proof utilizes the convergence results proved in \cref{theorem:non_asymptotic_private_distributed_mle}, and the concentration of$\vec N_{i, T}^{\thres, 1}(\hat \theta)$ (resp. $\vec N_{i, T}^{\thres, 2}(\hat \theta)$) similarly to \cref{theorem:asymptotic_threshold}.}

\begin{algorithm}[t]
\footnotesize
\captionsetup{font=footnotesize}
\caption{Private Distributed MLE (Two Threshold Algorithm)} \label{alg:threshold}

\begin{flushleft}
\noindent \textbf{Inputs:} Privacy budget $\eps$, Error probabilities $\alpha, 1 - \beta$, Log-belief Thresholds $\varrho^{\thres, 1}, \varrho^{\thres, 2} > 0$.

\noindent \textbf{Initialization:} Set the number of iterations $T$ and the number of rounds $K$, and the thresholds $\tauthres{2}, \tauthres{1}$ as indicated by \cref{theorem:asymptotic_threshold} (for the asymptotic case) or \cref{theorem:non_asymptotic_threshold} (for the non-asymptotic case), and $\hattauthres{1} = \frac {1} {1 + e^{\varrho^{\thres, 1}}}$ (resp. $\hattauthres{2}$). The DP noise distribution for protecting beliefs is $\cD_{i} (\hat \theta; \eps)$ which can be set optimally according to \cref{theorem:non_asymptotic_threshold}.

\noindent \textbf{Procedure:} The following is repeated in $K$ rounds indexed by $k\in[K]$. In each round $k$, the agents begin by forming the noisy likelihoods $\vec \sigma_{i, k}(\hat \theta) = e^{\vec d_{i, k} (\hat \theta)} \vec \gamma_i(\hat \theta)$, where $ \vec\gamma_{i}(\hat{\theta}) =\ell_i(\vec{s}_{i}|\hat{\theta})$ and $\vec d_{i, k} (\hat \theta)  \sim \cD_{i} (\hat \theta; \eps)$ independently across agents $i \in [n]$ and states $\hat \theta \in \Theta$. The agents initialize their beliefs to ${\vec\nu}_{i,k,0}(\hat{\theta})$ $ =$ $  \vec\sigma_{i, k}(\hat{\theta}) /\sum_{\tilde{\theta}\in\Theta} \vec\sigma_{i, k}(\tilde{\theta})$. Over the next $T$ time steps, the agents update their belief after communicating with their neighbors, and according to the following update rule:  
\begin{equation*}
\hspace{-10pt}{\vec\nu}_{i,k,t}(\hat{\theta})
=  \frac{{\vec\nu}^{1+a_{ii}}_{i, k, t-1}(\hat{\theta}) \prod\limits_{j\in\mathcal{N}_i }{{\vec\nu}^{a_{ij}}_{j, k,t-1}(\hat{\theta})} }{\sum\limits_{\tilde{\theta} \in \Theta}{\vec\nu}_{i,k, t-1}^{1+a_{ii}}(\tilde{\theta}) \prod\limits_{j\in\mathcal{N}_i} {{\vec\nu}^{a_{ij}}_{j, k,t-1}(\tilde{\theta})} } \quad \text{for every $\hat \theta \in \Theta$, $k \in [K]$ and $t \in [T]$.} \end{equation*} 

After $T$ iterations, each agent aggregates the results of the $K$ rounds as: 

\begin{equation*}
    \vec N_{i, T}^{\thres, 1} (\hat \theta) = \frac 1 K \sum_{k \in [K]} \one \left \{ \vec \nu_{i, k, T}(\hat \theta) >  \hattauthres{1} \right \}, \quad \vec N_{i, T}^{\thres, 2} (\hat \theta) = \frac 1 K \sum_{k \in [K]} \one \left \{ \vec \nu_{i, k, T}(\hat \theta) >  \hattauthres{2} \right \} \quad \text{for every $\hat \theta \in \Theta$.}
\end{equation*}

\noindent \textbf{Outputs:} Return
\begin{equation*}
    \hat \Theta^{\thres, 1}_{i, T} = \left \{ \hat \theta \in \Theta : \vec N_{i, T}^{\thres, 1}(\hat \theta) \ge \tauthres{1} \right \} \mathpunct{\raisebox{0.5ex}{,}} \quad \hat \Theta^{\thres, 2}_{i, T}  = \left \{ \hat \theta \in \Theta : \vec N_{i, T}^{\thres, 2} (\hat \theta) \ge \tauthres{2} \right \} \cdot    
\end{equation*}

\end{flushleft}
\end{algorithm}

To minimize the number of iterations, we can pick the thresholds as 
{\small
\begin{align*}
    \varrho^{\thres, 1}_\star & = \left ( \frac {|\Theta|^2 (n - 1) (n \Gamma_{n, \Theta} + V_{n, \eps, \Theta})} {2 q_1 \sqrt K} \right )^{\frac {\log 2} {\log (2 / a_n^\star)}} \left ( \frac {l_{n, \Theta}} {2 n} \right )^{\frac {\log (1 /  a_n^\star)} {\log (2 / a_n^\star)}},\\
    \varrho^{\thres, 2}_\star & = \left ( \frac {|\Theta|^2 (n - 1) (n \Gamma_{n, \Theta} + V_{n, \eps, \Theta})} {2 (1 - q_2) \sqrt K} \right )^{\frac {\log 2} {\log (2 / a_n^\star)}} \left ( \frac {l_{n, \Theta}} {2 n} \right )^{\frac {\log (1 / a_n^\star)} {\log (2 / a_n^\star)}}.
\end{align*}}

As in \cref{corollary:separation_threshold}, perfect recovery is also achievable with a single threshold when $q_2 > q_1$ (setting $\pi_1 = (1 - \pi_2) q_2 / q_1 - 1$); communication complexity is then minimized by taking $1 - q_2$ as close as possible to $q_1$, but exact recovery costs a polylogarithmic factor in $1 / |1 - q_2 - q_1|$.

% In what follows, we show that the GM estimator can be applied naturally to conduct distributed hypothesis tests at significance level $\alpha$.

\section{Extension to Private Distributed Online Learning from Intermittent Streams}\label{sec:app:dp-online-learning}

%\noindent \textbf{Online Learning from Intermittent Streams.} 
In the online setting, we consider a network of centers making streams of observations intermittently over time. At each time step $t$, center $i$ observes $\vec{s}_{i,t}$ that are distributed according to a model $\ell_i(\mathord{\cdot}|\hat \theta), \hat \theta \in \Theta$. In the online setting, we assume that the data are generated according to a true parameter $\theta^\circ \in \Theta$, which is uniquely identifiable from the collective observations of the centers. The collective identifiability condition can be expressed as $\sum_{i = 1}^n \kldiv {\ell_i(\cdot | \bar \theta)} {\ell_i(\cdot | \theta^\circ)} > 0, \forall \bar \theta \neq \theta^\circ$. Note that, unlike the MLE setting, where centers can increase the power of their inferences by exchanging belief statistics, in the online setting, each center alone has access to an infinite signal stream. 

An informational advantage of collective inference in the online setting is apparent where centers individually face identification problems; for example, for a center $i\in [n]$, a pair of states ($\hat\theta_1$ and $ \hat\theta_2$) may induce the same distribution of observations, making them indistinguishable from the perspective of center $i$. In such cases, by exchanging beliefs, centers can benefit from each other's observational abilities to collectively identify the truth even though they may face an identification problem individually.     

%\citet{rahimian2016distributed} give a non-private algorithm for belief exchange that is able to asymptotically recover the true state $\theta^\circ$ from a stream of observations (cf.  \Cref{sec:app:learning-non-priv-online}).% % and communicate their beliefs at every time period

\noindent\textbf{The Non-private Online Learning from Benchmark.}
%\label{sec:app:learning-non-priv-online} 
\citet{rahimian2016distributed} give \cref{alg:non_private_intermittent_streams} for the agents to determine the true state $\theta^{\circ}$ from their streams of observations in the online learning regime, and prove that this algorithm converges to learning the true state asymptotically. Their argument relies on the fact that the time average of the log-belief ratio between any state $\hat{\theta} \in \Theta$ 
 and the true state $\theta^{\circ}$, $(1/t)\log(\vec \mu_{i, t}(\hat \theta)/\vec \mu_{i, t}(\theta^{\circ}))$, converges to the weighted sum of the KL divergences of all agents, which is less than zero if the models are statistically identifiable.

\begin{algorithm}[h]
\footnotesize
\captionsetup{font=footnotesize}
\caption{Non-Private Distributed Online Learning} \label{alg:non_private_intermittent_streams}

\begin{flushleft}
Every time $t \in\mathbb{N}_0$, each agent forms the likelihood product of the datasets that it has received at that iteration: $
\vec\gamma_{i,t}(\hat{\theta}) = \ell_i(\vec{s}_{i,t}|\hat{\theta})$, and $\gamma_{i,t}(\hat{\theta}) = 1$ if no data is present. The agent then updates its belief according to:
\begin{equation*}
\hspace{-15pt}{\vec\mu}_{i,t}(\hat{\theta}) = \frac{\vec\gamma_{i,t}(\hat{\theta}) \vec\mu^{a_{ii}}_{i,t-1}(\hat{\theta}) \prod_{j\in\mathcal{N}_i}{{\vec\mu}^{a_{ij}}_{j,t-1}(\hat{\theta})}} {\sum\limits_{\tilde{\theta} \in \Theta} \vec\gamma_{i,t}(\tilde{\theta})\vec\mu^{a_{ii}}_{i,t-1}(\tilde{\theta})  \prod_{j\in\mathcal{N}_i}{{\vec\mu}^{a_{ij}}_{j,t-1}(\tilde{\theta})}},  \text{for all $\hat \theta \in \Theta$ and $t \in [T]$}, 
\end{equation*} initialized by:  ${\vec\mu}_{i,0}(\hat{\theta}) =  \vec\gamma_{i,0}(\hat{\theta}) /\sum_{\tilde{\theta}\in\Theta} \vec\gamma_{i,0}(\tilde{\theta})$.
\end{flushleft}
\end{algorithm}

\cref{alg:private_online_learning} facilitates differentially private distributed learning in an online setting, where centers receive  datasets at each time step $t$, calculate the likelihood $\vec\gamma_{i,t}(\hat{\theta}) =\ell_i(\vec{s}_{i,t}|\hat{\theta})$ and its noisy version $\vec \sigma_{i, t} (\hat \theta) = e^{\vec d_{i, t} (\hat \theta)} \vec \gamma_{i, t}(\hat \theta)$ to preserve DP, and then aggregate it with their self- and neighboring beliefs from iteration $t - 1$. Unlike the MLE setup, for online learning, we do not need to repeat the algorithm in multiple rounds because DP noise cancels out as $T \to \infty$.

\begin{algorithm}[!ht]
\footnotesize
\captionsetup{font=footnotesize}
\caption{Private Distributed Online Learning} \label{alg:private_online_learning}

\begin{flushleft}
\noindent \textbf{Inputs:}  Privacy budget $\eps$, Error probabilities $\eta$.

\noindent \textbf{Initialization:} Set the number of iterations $T$ as in \cref{theorem:non_asymptotic_private_online_learning}. The DP noise distribution for protecting beliefs is $\cD_{i} (\hat \theta; \eps)$ which can be set optimally according to \cref{theorem:non_asymptotic_private_online_learning}.

\noindent \textbf{Procedure:} Every time $t \in\mathbb{N}_0$, each center forms the likelihood product of the datasets that it has received at that time period: $\vec\gamma_{i,t}(\hat{\theta}) = \ell_i(\vec{s}_{i,t}|\hat{\theta})$,  and $\gamma_{i,t}(\hat \theta) = 1$ if no data. The center then draws a noise variable $\vec d_{i, t} (\hat \theta) \sim \cD_{i, t} (\hat \theta; \eps)$ for all $i \in [n]$, $\hat \theta \in \Theta$ independently and forms $\vec \sigma_{i, t} (\hat \theta) = e^{\vec d_{i, t} (\hat \theta)} \vec \gamma_{i, t}(\hat \theta)$ for all $\hat \theta \in \Theta$. The center then updates its belief as:
\begin{equation*}
{\vec\nu}_{i,t}(\hat{\theta}) = \frac{\vec\sigma_{i,t}(\hat{\theta}) \vec\nu^{a_{ii}}_{i,t-1}(\hat{\theta}) \prod_{j\in\mathcal{N}_i}{{\vec\nu}^{a_{ij}}_{j,t-1}(\hat{\theta})}} {\sum\limits_{\tilde{\theta} \in \Theta} \vec\sigma_{i,t}(\tilde{\theta})\vec\nu^{a_{ii}}_{i,t-1}(\tilde{\theta})  \prod_{j\in\mathcal{N}_i}{{\vec\nu}^{a_{ij}}_{j,t-1}(\tilde{\theta})}} \quad \text{for every $\hat \theta \in \Theta$ and $t \in [T]$},
\end{equation*} initialized by:  ${\vec\nu}_{i,0}(\hat{\theta}) =  \vec\sigma_{i,0}(\hat{\theta}) /\sum_{\tilde{\theta}\in\Theta} \vec\sigma_{i,0}(\tilde{\theta})$. 

\noindent \textbf{Outputs:} After $T$ iterations, return $\hat \theta^\intt_{i, T} = \argmax_{\hat \theta \in \Theta} \vec \nu_{i, T}(\hat \theta)$. 
    
\end{flushleft}

\end{algorithm}

In addition to the global sensitivity parameters $\Delta_{i, \Theta}$ and $\Delta_{\Theta}^n = \max_{i \in [n]} \Delta_{i, \Theta}$ (which determine the privacy noise level), the finite-time convergence and implied communication complexity bounds for \cref{alg:private_online_learning} depend on the following statistical properties of the information environment (similarly to \cref{eq:l,eq:Q}, which we encountered in the main text for MLE):

\begin{itemize}
%\emph{\textbf{Private signal structures:}} 
    \item The minimum divergence between the true state $\theta^\circ$ and any other state $\bar \theta \neq \theta^\circ$:
    { \begin{equation} l_{n, \Theta} = \min_{\bar \theta \neq\theta^{\circ}} \left | \sum_{i = 1}^n  \kldiv {\ell_i(\cdot | \bar \theta)} {\ell_i(\cdot | \theta^\circ)}\right |, \mbox{ and } \label{eq:l-online}
    \end{equation}}
    \item The maximum standard deviation of the $\log$-likelihood ratio statistics: 
    { \begin{equation}
    Q_{n, \Theta} = \max_{i \in [n], \bar \theta \neq \theta^\circ} \sqrt {\var {} {\log \left ( \frac {\ell_i(\vec s_i | \bar \theta)} {\ell_i(\vec s_i | \theta^\circ)} \right )}}. \label{eq:Q-online}
    \end{equation}}%similarly to \cref{eq:l,eq:Q} that we encountered in the main text for MLE.
\end{itemize}

The asymptotic dynamics of beliefs under \cref{alg:private_online_learning} become clear by noticing  that the log-belief ratio between $\theta^\circ$ and any other state $\bar \theta\neq \theta^{\circ}$ converges to $(-1/n) \sum_{i = 1}^n  \kldiv {\ell_i(\cdot | \bar \theta)} {\ell_i(\cdot | \theta^\circ)}$ as $t \to \infty$, and the effect of DP noise disappears as a consequence of the C\'esaro mean and the weak law of large numbers. Therefore, centers learn the true state $\theta^{\circ}$ exponentially fast, asymptotically (one can show that the asymptotic exponential rate is $-l_{n, \Theta} / n$, where $l_{n, \Theta}$ is given by \cref{eq:l}).

The next theorem characterizes the non-asymptotic behavior of \cref{alg:private_online_learning} (proved in \cref{sec:app:online-learning}):

\begin{theorem}\label{theorem:non_asymptotic_private_online_learning} 
Suppose the collective identifiability condition $\sum_{i = 1}^n  \kldiv {\ell_i(\cdot | \bar \theta)} {\ell_i(\cdot | \theta^\circ)} > 0$ holds for all $\bar \theta \neq \theta^\circ$. Let $l_{n, \Theta}$ and $Q_{n, \Theta}$ be the statistical properties of the datasets defined in \cref{eq:l-online,eq:Q-online}, and denote the second-largest eigenvalue modulus (SLEM) of $A$ by $b_n^\star = |\lambda_2(A)|$. Then running \cref{alg:private_online_learning} for $T = \frac {\log (|\Theta|) + \frac {|\Theta| \frac {\sqrt 2} {n} (n Q_{n, \Theta} + V_{n, \eps, \Theta})} {2 \eta (1 - {b_n^\star})}} {l_{n, \Theta} / n} = \cO \left ( \frac {n |\Theta| (Q_{n, \Theta} +  {\Delta_\Theta^n |\Theta|}/{\eps})} {l_{n, \Theta} \eta (1 - b_n^\star)}  \right )$ iterations yields $\Pr \left [\theta^\circ = \hat \theta^\intt_{i, T} \right ] \ge 1 - \eta$ for every center $i \in [n]$. Moreover, the noise distributions that minimize the runtime are the Laplace distributions $\cD_{i, t}^\star = \mathrm{Lap} \left (  {\Delta_{i, \Theta} |\Theta|}/{\eps} \right )$, and the resulting estimates are $\eps$-DP with respect to the private datasets.
\end{theorem}

Note that similarly to distributed MLE algorithms, here we also need to scale the privacy budget by $|\Theta|$ to account for the fact that the algorithm accesses the private datasets and adds noise for each of the $|\Theta|$ states, providing an attacker with as many observations of the private datasets. 

The following table compares the communication complexity of distributed online learning for a fixed privacy budget $\eps > 0$ in \cref{theorem:non_asymptotic_private_online_learning} against the non-private benchmark due to \citet{rahimian2016distributed}:

\begin{table*}[!h]
\centering
    \scriptsize
    \begin{tabular}{lcc}
    \toprule
    Comparison w.r.t. & With DP (ours) & Without DP~\citep{rahimian2016distributed} \\ 
    \midrule
    %    & \multicolumn{2}{c}{Communication Complexity with respect to $\eps, \eta$} \\
    % \midrule 
    % MLE (AM/GM) & $\Opriv {{\color{belize}{\log (1 / \eta) (\log (1 / \eta) + \log (1 / \eps) + \log (1 / \varrho)}}}$ & $\Opriv { {1} }$ \\
    % MLE (2-Threshold) & $\Opriv {{\color{belize}{ \frac {\log (1 / \eta)} {\pi^2} (\log (1 / \eta) + \log (1 / \eps) + \log (1 / q))}}}$ & $\Opriv { {1} }$ \\
    $\eps, \eta$ &  $\Opriv {{\color{belize}{\frac 1 {\eta} {\frac 1 {\eps}}}}}$ & $\Opriv { {1} }$ \\
    % \toprule
    %    & \multicolumn{2}{c}{Communication Complexity with respect to $n$} \\
    % \midrule 
    % MLE (AM/GM) & $\Onn {\max \left \{ \log \left ( \frac {n} {l_{n, \Theta}} \right ),  \frac {\log \left ( \frac {n (\Gamma_{n, \Theta} + {\color{belize}{\Delta_\Theta^n}})} {{\color{belize}{\varrho}}}  \right )} {\log (1 / a_n^\star)} \right \}}$ & $\cO_{|\Theta|} \left ( \max \left \{ \log \left ( \frac {n} {l_{n, \Theta}} \right ),  \frac {\log \left ( 
    % {n} {\Gamma_{n, \Theta}} \right )} {\log (1 / a_n^\star)} \right \} \right )$ \\
    % MLE (2-Threshold) & $\Onn {{\color{belize}{\frac {1} {\pi^2}}}  \max \left \{ \log \left ( \frac {n} {l_{n, \Theta}} \right ),  \frac {\log \left ( \frac {n (\Gamma_{n, \Theta} + {\color{belize}{\Delta_\Theta^n}})} {{\color{belize}{q}}}  \right )} {\log (1 / a_n^\star)} \right \}}$ & $\cO_{|\Theta|} \left ( \max \left \{ \log \left ( \frac {n} {l_{n, \Theta}} \right ),  \frac {\log \left ( 
    % {n} {\Gamma_{n, \Theta}} \right )} {\log (1 /  a_n^\star)} \right \} \right )$ \\

    $n$ & $\Onn {\frac {n(Q_{n, \Theta} + {\color{belize}{\Delta_\Theta^n}}) } {l_{n, \Theta} (1 - b_n^\star)}}$ & $\cO_{|\Theta|} \left ( \frac {nQ_{n, \Theta} } {l_{n, \Theta} (1 - b_n^\star)} \right )$ \\
    % \midrule
    %     & \multicolumn{2}{c}{Communication Complexity with respect to $|\Theta|$} \\
    % \midrule
    % MLE (AM/GM) & $\Otheta {{\color{belize}{|\Theta| \log |\Theta|}} \left ( {\color{belize}{\log |\Theta|}} + \log \left ( \frac {\Gamma_{n, \Theta} + {\color{belize}{|\Theta| \log |\Theta| \Delta_\Theta^n}}} {{\color{belize}{\varrho}}} \right )\right )}$ & $\cO_n \left ( {\log \Gamma_{n, \Theta}} \right )$ \\
    % MLE (2-Threshold) & $\Otheta {{\color{belize}{\frac {\log |\Theta|} {\pi^2}}} \left ( {\color{belize}{\log |\Theta|}} + \log \left ( \frac {\Gamma_{n, \Theta} + {\color{belize}{|\Theta| \log |\Theta| \Delta_\Theta^n}}} {{\color{belize}{q}}} \right ) \right )}$ & $\cO_n \left ( {\log \Gamma_{n, \Theta}} \right )$ \\

    $|\Theta|$ & $\Otheta {\frac {|\Theta| (Q_{n,\Theta} + {\color{belize}{|\Theta| \Delta_\Theta^n}})} {l_{n, \Theta}}}$ &  $\cO_n \left ( \frac {|\Theta| Q_{n,\Theta}} {l_{n, \Theta}} \right )$ \\
    \bottomrule
 
    \end{tabular}
    
   \caption{Communication complexity of private and non-private distributed online learning. The overhead of introducing privacy is outlined in {\color{belize}{blue}}.%\footnotesize{Comparing communication complexity of distributed online learning for a fixed privacy budget $\eps > 0$ against the non-private benchmark due to \citet{rahimian2016distributed}.}
   } \label{tab:summary-online-learning}
\end{table*}

\clearpage

\section{Proofs and Convergence Analysis for Private Distributed MLE}\label{sec:app:dist-MLE-convergence}

\subsection{Log-Belief Ratio Notations}\label{sec:app:log-belief-ratio-notation}

Let $\vec \mu_{i, t}(\hat \theta)$ be the beliefs for the nonprivate system, and let $\vec \nu_{i, k, t}(\hat \theta)$ be the private estimates for agent $i \in [n]$, at the time step $t \ge 0$ of round $k \in [K]$. For the non-private algorithm (\cref{alg:non_private_mle}) all pairs $\hat \theta, \check \theta \in \Theta$ we let $\vec \phi_{i, t}(\hat \theta, \check \theta) = \log \left ( \frac {\vec \mu_{i, t}(\hat \theta)} {\vec \mu_{i, t}(\check \theta)} \right )$ and $\vec \lambda_i (\hat \theta, \check \theta) = \log \left ( \frac {\vec \gamma_i(\hat \theta)} {\vec \gamma_i(\check \theta)} \right )$ be the log belief ratios and the log-likelihood ratios of the agent $i \in [n]$ in the time step $t \ge 0$. For the private algorithm (\cref{alg:private_distributed_mle}) we let $\vec \psi_{i, k, t}(\hat \theta, \check \theta)  = \log \left ( \frac {\vec \nu_{i, k, t}(\hat \theta)} {\vec \nu_{i, k, t}(\check \theta)} \right ), \vec \zeta_{i,k}(\hat \theta, \check \theta)  = \log \left ( \frac {\vec \sigma_{i, k}(\hat \theta)} {\vec \sigma_{i, k}(\check \theta)} \right ),$ and $\vec \kappa_{i, k}(\hat \theta, \check \theta) = \vec d_{i, k}(\hat \theta) - \vec d_{i, k}(\check \theta)$ be the private log-belief ratio, log-likelihood ratio and log of the noise ratio between states $\hat \theta$ and $\check \theta$ for agent $i \in [n]$  at time step $t \ge 0$ of round $k \in [K]$, and denote their vectorized versions by $\ovec \phi_{t}(\hat \theta, \check \theta)$, $\ovec \psi_{k, t}(\hat \theta, \check \theta)$, $\ovec \lambda(\hat \theta, \check \theta)$, and $\ovec \zeta_k(\hat \theta, \check \theta)$, respectively, where the vectorization is over the agents $i \in [n]$. We define $\Lambda(\hat \theta, \check \theta) = \one^T \ovec \lambda (\hat \theta, \check \theta)$, $Z_k(\hat \theta, \check \theta) = \one^T \ovec \zeta_k (\hat \theta, \check \theta) = \one^T \left ( \ovec \lambda (\hat \theta, \check \theta) + \ovec \kappa_k(\hat \theta, \check \theta) \right )$.

We say that a stochastic process $\{ X_t \}_{t \in \Nbb}$ converges in $L_2$ to $X$, and write $X_t \overset {L_2} \to X$, if and only if $\lim_{t \to \infty} \ev {} {\| X_t - X \|_2} = 0$. Convergence in $L_2$ implies convergence in probability (i.e., $X_t \overset p \to X$) due to Markov's inequality.

% \subsection{Auxiliary Lemmas}

To prove the results, we need the following auxiliary lemma:

\begin{lemma} \label{lemma:record}
    Let $X_1, \dots, X_n$ be i.i.d. draws from a distribution $\cD$. Then, for every $i \in [n]$, we have $\Pr [X_i \ge \max_{j \neq i} X_j] = 1 / n$.
\end{lemma}

    Let $E_i = \{ X_i \ge \max_{j \neq i} X_j \}$. Note that because $X_1, \dots, X_n$ are i.i.d. $\Pr [E_1] = \Pr[E_2] = \dots = \Pr [E_n]$. Moreover, note that since the maximum is unique, we have that $E_1, \dots, E_n$ are a partition of the sample space. Therefore, we get $\sum_{i \in [n]} \Pr [E_i] = 1$, and subsequently $\Pr [E_i] = 1 / n$. \qed

\subsection{Proof of \texorpdfstring{\cref{theorem:asymptotic_private_distributed_mle}}{theorem-asymptotic-private-distributed-mle}: Asymptotic Behavior of the AM/GM Algorithm}\label{sec:app:proof:theorem:asymptotic_private_distributed_mle}

\noindent Similarly to \cite[Theorem 1]{rahimian2016distributed}, we have that for any $k \in [K]$
\begin{align*}
    \left \| \ovec \psi_{k, t}(\hat \theta, \check \theta) - \frac {2^t} {n} Z_k(\hat \theta, \check \theta) \one \right \|_2 & = \left \| \sum_{i = 2}^n \left ( \frac {\lambda_i(A+I)} 2 \right )^t \ovec l_i \ovec r_i^T \ovec \zeta_k(\hat \theta, \check \theta)  \right \|_2 \\ 
    & \le \sum_{i = 2}^n \left | \frac {\lambda_i(A+I)} 2 \right |^t \| \ovec l_i \|_2 \left | \ovec r_i^T \ovec \zeta_k(\hat \theta, \check \theta) \right | \\
    & \le \sum_{i = 2}^n \left | \frac {\lambda_i(A+I)} 2 \right |^t \left \| \ovec \zeta_k (\hat \theta, \check \theta) \right \|_2 
    \\ & \le \sum_{i = 2}^n \left | \frac {\lambda_i(A+I)} 2 \right |^t \left ( \left \| \ovec \lambda (\hat \theta, \check \theta) \right \|_2 + \left \| \ovec \kappa_k (\hat \theta, \check \theta) \right \|_2\right ). 
\end{align*}

Taking expectations and applying Jensen's inequality, we can bound the above as
\begin{equation} \label{eq:exp_residual-asymptotic}
    \ev {} { \left \| \ovec \psi_{k, t}(\hat \theta, \check \theta) - \frac {2^t} {n} Z_k(\hat \theta, \check \theta) \one \right \|_2} \le 2 (n - 1) \left | \frac {\lambda_2(A+I)} 2 \right |^t \left [ n \Gamma_{n, \Theta} + \sum_{i = 1}^n \sqrt {\var {} {\vec d_{i, k}(\hat \theta)}} \right ],
\end{equation}
where $\Gamma_{n, \Theta} = \max_{i \in [n], \hat \theta \in \Theta} | \log \vec \gamma_i(\hat \theta)|$ and $|\lambda_2(A+I)|/2 <1$ is the SLEM of $(A + I) / 2$, henceforth also denoted by $a^{\star}_n$; $A$ is irreducible doubly stochastic, so $A + I$ is primitive and $0 < |\lambda_{n} (A+ I)| \le |\lambda_{n - 1} (A + I)|  \le  \dots \le |\lambda_{2} (A + I)| < \lambda_1(A + I) = 2$. Note that the right-hand side in \cref{eq:exp_residual-asymptotic} goes to $0$ as $t \to \infty$, which implies that (by Markov's inequality) $\ovec \psi_{k, t}(\hat \theta, \check \theta) \overset {L_2} {\to} \frac {2^t} n Z_k(\hat \theta, \check \theta) \one$. 

Let $\theta^\star \in \Theta^\star$ and $\bar \theta \in \compTheta$. Based on our definition of $\Lambda(\cdot, \cdot)$ we should have $\Lambda(\bar \theta, \theta^\star) < 0$ and the corresponding non-private algorithm would have $\phi_{i, t} (\bar \theta, \theta^\star) \to - \infty$ for every $\bar \theta \in \compTheta$ and $\theta^\star \in \Theta^\star$. However, when noise is introduced, there could be mistakes introduced in the log-belief ratios, even if $\Lambda(\bar \theta, \theta^\star) < 0$ we can have $Z_k(\bar \theta, \theta^\star) \ge 0$ which implies $\vec \psi_{i, k, t} (\bar \theta, \theta^\star) \not \to - \infty$. 

\noindent \textbf{AM estimator.} For all $\hat \theta \in \Theta$ we let $\vec Y_j(\hat \theta) = \sum_{i = 1}^n \vec d_{i, k} (\hat \theta)$. We focus on a single $\theta^\star \in \Theta^\star$. It is easy to show that because the noise is independent across the agents, and i.i.d. between the states for a given agent $i$, then $\vec Y_j(\bar \theta)$ and $\vec Y_j(\theta^\star)$ are also i.i.d. for all $\bar \theta \in \compTheta$, and therefore we have that (see \cref{lemma:record})
\begin{equation} \label{eq:symmetry_1}
    \lim_{T \to \infty} \Pr \left[ \vec \nu_{i, k, T}(\theta^\star) > 0 \right ]  \ge \Pr \left [ \bigcap_{\bar \theta \in \compTheta} \left \{ \vec Y_k (\bar \theta) \le \vec Y_k (\theta^\star) \right \} \right ] = \frac {1} {|\compTheta|}.
\end{equation}

% Similarly we can get that for every $\hat \theta \in \compTheta$ we have that $\Pr [\lim_{t \to \infty} \vec \nu_{i, j, t}(\theta^\star) = 0] \ge \frac $

Therefore, for the AM estimator, we have the following: 
\begin{equation*}
    \lim_{T \to \infty} \Pr \left [ \vec \nu^{\AM}_{i, T} (\theta^\star) = 0 \right ] = \lim_{T \to \infty} \Pr \left [ \bigcap_{k \in [K]} \left \{ \vec \nu_{i, k, T} (\theta^\star) = 0 \right \}  \right ] \le \left ( 1 - \frac {1} {|\compTheta|} \right )^K \le e^{- \frac {K} {|\compTheta|}},
\end{equation*}
and the failure probability of the AM estimator can be calculated by applying the union bound as
\begin{align*}
    \lim_{T \to \infty} \Pr \left [ \Theta^\star \not \subseteq \hat \Theta^\AM_{i, T} \right ] & = \lim_{T \to \infty} \Pr \left [ \exists \theta^\star \in \Theta^\star : \vec \nu^\AM_{i, T}(\theta^\star) = 0 \right ] \\
    & \le \lim_{T \to \infty} \sum_{\theta^\star \in \Theta^\star} \Pr \left [ \vec \nu^\AM_{i, T}(\theta^\star) = 0 \right ] \\
    & \le |\Theta^\star| e^{- \frac K {|\compTheta|}}.
\end{align*}

Setting $K = |\compTheta| \log (|\Theta^\star| / (1 - \beta))$, we can ensure that the Type II error rate is at most $1 - \beta$.

\noindent \textbf{GM estimator.} Similarly, for $\vec \nu_{i, T}^{\GM}$ we have the following:
\begin{equation} \label{eq:symmetry_2}
\lim_{T \to \infty} \Pr \left [\vec \nu_{i, k, T}(\bar \theta) = 0 \right ] \ge \Pr \left [ \bigcap_{\theta^\star \in \Theta^\star} \left \{ \vec Y_k(\bar \theta) \le \vec Y_k(\theta^\star) \right \}\right ] = \frac {1} {|\Theta^\star|},    
\end{equation}

for all $\bar \theta \in \compTheta$ and subsequently $$\lim_{T \to \infty} \Pr \left [ \vec \nu_{i, T}^{\GM} (\bar \theta) > 0 \right ] = \lim_{T \to \infty} \Pr \left [ \bigcap_{k \in [K]} \left \{  \vec \nu_{i, k, T}(\bar \theta) > 0 \right \} \right ] \le \left ( 1 - \frac {1} {|\Theta^\star|} \right )^K \le e^{-\frac {K} {|\Theta^\star|}},$$ and 
\begin{align*}
\lim_{T \to \infty} \Pr \left [ \hat \Theta^\GM_{i, T} \not \subseteq \Theta^\star  \right ] & =  \lim_{T \to \infty} \Pr \left [ \hat \Theta^\GM_{i, T} \cap \compTheta  \neq \emptyset \right ] \\ & =  \lim_{T \to \infty} \Pr \left [ \exists \bar \theta \in \compTheta :  \vec \nu^\GM_{i, T}(\bar \theta) > 0 \right ] \\ & \le \lim_{T \to \infty} \sum_{\bar \theta \in \compTheta} \Pr \left [  \vec \nu^\GM_{i, T}(\bar \theta) > 0 \right ] \\ & \le |\compTheta| e^{-\frac {K} {|\Theta^\star|}}.
\end{align*}

Therefore, selecting $K = |\Theta^\star| \log (|\compTheta| / \alpha)$ produces a Type I error rate of at most $\alpha$.

\noindent \textbf{Negative Impact of $K$ on AM Estimator.} From the above we know that setting $K \ge |\compTheta| \log (|\Theta^\star| / (1 - \beta))$ we have that $\Theta^\star \subseteq \hat \Theta^\AM_{i, T}$ with probability at least $\beta$ as $T \to \infty$. We call this ``good event`` $E^{\AM}$. Conditioned on $E^{\AM}$, we have that: 

\begin{align*}
    \lim_{T \to \infty} \Pr \left [|\Theta \setminus \hat \Theta^\AM_{i, T}| = 1 | E^\AM \right ] & \overset {(i)} {\ge} \lim_{T \to \infty}  \Pr \left [ \exists \bar \theta \in \compTheta : \vec \nu_{i, T}(\bar \theta) = 0 | E^\AM \right ] \\
    & \overset {(ii)} {\ge} \lim_{T \to \infty}  \Pr \left [ \vec \nu_{i, T}(\bar \theta_0) = 0 | E^\AM \right ] \\
    & \overset {(iii)} {=} \lim_{T \to \infty}  \left ( 1 -  \Pr \left [ \vec \nu_{i, T}(\bar \theta_0) > 0 | E^\AM \right ] \right ) \\
    & \overset {(iv)} {\ge} 1 - e^{-K/|\compTheta|}, 
\end{align*}

where we have used the following facts: (i) conditioned on $E$ a rejected state can only belong to $\compTheta$, (ii) the probability of the union of events over $\compTheta$ is a superset of the event $\{ \vec \nu_{i, T}(\bar \theta_0) = 0 \}$ for any fixed $\bar \theta_0 \in \compTheta$, (iii) complement of the event, and (iv) using \cref{eq:symmetry_1}. Therefore 

\begin{align*}
\lim_{T \to \infty} \Pr \left [|\Theta \setminus \hat \Theta^\AM_{i, T}| \ge 1 \right ]  & \ge \lim_{T \to \infty}  \Pr [E^\AM]  \Pr \left [|\Theta \setminus \hat \Theta^\AM_{i, T}| \ge 1 | E^\AM \right ] \\ & \ge (1 - (1 - \beta)) (1 - e^{-K/|\compTheta|}) \\ & \ge 1 - (1 - \beta) - e^{-K/|\compTheta|}.
\end{align*}

\noindent \textbf{Negative Impact of $K$ on GM Estimator.} From the above we know that setting $K \ge |\Theta^\star| \log (|\compTheta| / \alpha)$ we have that $\Theta^\star \supseteq \hat \Theta^\GM_{i, T}$ with probability at least $1 - \alpha$ as $T \to \infty$. We call this ``good event`` $E^{\GM}$. Fix some $\theta_0^\star \in \Theta^\star$. Conditioned on $E^{\GM}$, we have that: 

\begin{align*}
\lim_{T \to \infty}  \Pr \left [ |\Theta^\GM_{i, T} | = 1 | E^\GM \right ] & \overset {(i)} {=} \lim_{T \to \infty}  \Pr \left [ \bigcup_{\theta^\star \in \Theta} \{ \Theta^\GM_{i, T} = \{ \theta^\star \} \} | E^\GM \right ] \\ 
    & \overset {(ii)} \ge \lim_{T \to \infty}  \Pr \left [ \Theta^\GM_{i, T} = \{ \theta_0^\star \} | E^\GM \right ] \\ 
    & \overset {(iii)} {=} \lim_{T \to \infty}  \Pr \left [ \bigcap_{\theta' \in \Theta^\star : \theta' \neq \theta_0^\star} \{ \vec \nu_{i, T} (\theta') = 0 \} | E^\GM \right ] \\
    & \overset {(iv)} = \lim_{T \to \infty}   \left ( 1 -  \Pr \left [ \bigcup_{\theta' \in \Theta^\star : \theta' \neq \theta^\star} \{ \vec \nu_{i, T} (\theta') > 0 \} | E^\GM \right ] \right ) \\
    & \overset {(v)} \ge \lim_{T \to \infty}   \left ( 1 - \sum_{\theta' \in \Theta^\star : \theta' \neq \theta_0^\star}  \Pr \left [  \{ \vec \nu_{i, T} (\theta') > 0 \} | E^\GM \right ] \right ) \\
    & \overset {(vi)} \ge 1 - |\Theta^\star|e^{-K/|\Theta^\star|},
\end{align*}

where we have used the following facts: (i) definitiona of the $\hat \Theta^\GM_{i, T}$ being a signleton conditioned on $E^\GM$, (ii) the fact that the union of events is a superset of a single event,  (iii) the fact that the resulting state is a singleton if and only if all other states from $\Theta^\star$ are rejected, (iv) complement of the event, (v) union bound over $\Theta^\star \setminus \{ \theta^\star \}$, and (vi) \cref{eq:symmetry_2}. Finally, combining everything we get that: 

\begin{align*}
    \lim_{T \to \infty} \Pr \left [ |\hat \Theta^\GM_{i, T}| = 1 \right ] &  \ge \lim_{T \to \infty} \Pr [E^\GM] \Pr \left [ |\hat \Theta^\GM_{i, T}| = 1 | E^\GM \right ] \\
    & \ge (1 - \alpha) (1 - |\Theta^\star|e^{-K/|\Theta^\star|}) \\
    & \ge 1 - \alpha - |\Theta^\star|e^{-K/|\Theta^\star|}.
\end{align*}

% \noindent \textbf{Negative impact of $K$.} Moreover, to show the negative impact of $K$, we can also show that the probability of rejecting states from $\Theta^\star$ also goes to 0 as $K$ increases. Notice that 

% where the steps follow from complementarity, the union bound, and the intersection of independent events with equal probability $1 - 1 / |\compTheta|$ (cf. \cref{eq:symmetry_1}). Finally, we can deduce that 

% \begin{equation*}
%     \lim_{T \to \infty} \Pr \left [ \hat \Theta_{i, T}^\GM = \emptyset \right ] \ge 1 - |\Theta^\star| e^{-K/|\compTheta|} - |\compTheta| e^{-\frac {K} {|\Theta^\star|}} 
% \end{equation*}

% With an exactly symmetrical argument, we can show that for AM, all states from $\compTheta$ will be accepted with probability.

% \begin{equation*}
%     \lim_{T \to \infty} \Pr \left [ \compTheta \subseteq \hat \Theta^\AM_{i, T} \right ] \ge 1 - |\compTheta| e^{-K/|\Theta^\star|}
% \end{equation*}

% and therefore

% \begin{equation*}
%     \lim_{T \to \infty} \Pr \left [ \Theta^\AM_{i, T} = \Theta \right ] \ge 1 - |\compTheta| e^{-K/|\Theta^\star|} - |\Theta^\star| e^{-K/|\compTheta|}
% \end{equation*}

% And, finally,

% \begin{equation*}
%     \lim_{T \to \infty} \Pr \left [ \left \{ \Theta^\AM_{i, T} = \Theta \right \} \cap \left \{ \hat \Theta_{i, T}^\GM = \emptyset \right \} \right ] \ge 1 - 2 \left ( |\compTheta| e^{-K/|\Theta^\star|} - |\Theta^\star| e^{-K/|\compTheta|} \right )
% \end{equation*}

\noindent \textbf{Differentially Private Outputs. } Due to the postprocessing property of DP (see Proposition 2.1 of \citet{dwork2014algorithmic}), the resulting estimates are $\eps$-DP with respect to the private datasets.

\qed

\subsection{Proof of \texorpdfstring{\cref{theorem:non_asymptotic_private_distributed_mle}}{theorem-non-asymptotic-private-distributed-mle}: Finite-Time Guarantees for AM/GM Algorithm}\label{sec:app:proof:theorem:non_asymptotic_private_distributed_mle}

For simplicity of exposition, we have set $\varrho^\AM = \varrho^\GM = \varrho$ which corresponds to $\tau^\AM = \tau^\GM = \tau$. 

\noindent \textbf{GM estimator.} We let $\vec Y_k(\hat \theta) = \sum_{i \in [n]} \vec d_{i, k} (\hat \theta)$. We define the event $F = \left\{\exists k \in [K], \bar \theta \in \compTheta, \theta^\star \in \Theta^\star : Z^\GM(\bar \theta, \theta^\star) > \Lambda(\bar \theta, \theta^\star) \right \}$. We have that 
\begin{align*}
    \Pr [F] & \ensuremath{\stackrel{\text{(i)}}{\le}} \sum_{\bar \theta \in \compTheta} \Pr \left [\exists k \in [K], \theta^\star \in \Theta^\star : Z_k(\bar \theta, \theta^\star) > \Lambda(\bar \theta, \theta^\star) \right ] \\ 
    & \ensuremath{\stackrel{\text{(ii)}}{=}}   
    \sum_{\bar \theta \in \compTheta} \left ( 1 - \Pr \left [\forall \theta^\star \in \Theta^\star : Z_1(\bar \theta, \theta^\star) \le \Lambda(\bar \theta, \theta^\star) \right ] \right )^K \\
    & \ensuremath{\stackrel{\text{(iii)}}{\le}}  |\compTheta| \left ( 1 - \Pr \left [ \bigcap_{\bar \theta \in \compTheta} \left \{ \vec Y_1 (\bar \theta) \le \vec Y_1 (\theta^\star) \right \} \right ]  \right )^K \\
    & \ensuremath{\stackrel{\text{(iv)}}{\le}} |\compTheta| \left ( 1 - \frac {1} {|\Theta^\star|} \right )^K \\
    & \ensuremath{\stackrel{\text{(v)}}{\le}} |\compTheta| e^{-K / |\Theta^\star|},
\end{align*} where the above follows (i) from the application of the union bound, (ii) using the fact that the $K$ rounds are independent, (iii) the fact that $\{ Z_1 (\bar \theta, \theta^\star) \le \Lambda(\bar \theta, \theta^\star) \} \supseteq \bigcap_{\bar \theta \in \compTheta} \left \{ \vec Y_k (\bar \theta) \le \vec Y_k (\theta^\star) \right \}$ and that $\{ \vec Y_k(\bar \theta) \}_{\bar \theta \in \compTheta}$ and $\vec Y_k(\theta^\star)$ are i.i.d., (iv) \cref{lemma:record}, and (v) $1 + x \le e^x$ for all $x \in \Rbb$.  

For each round $k$ of the algorithm and each pair of states $\hat \theta, \check \theta \in \Theta$, the following holds for the log-belief ratio of the geometrically averaged estimates:
\begin{equation*}
    \vec \psi_{i, t}^{\GM}(\hat \theta, \check \theta) = \frac 1 K \sum_{k \in [K]} \vec \psi_{i, k, t}(\hat \theta, \check \theta).
\end{equation*}
We let $Z^\GM(\hat \theta, \check \theta) = \frac 1 K \sum_{k = 1}^K Z_k(\hat \theta, \check \theta)$. By the triangle inequality and \cref{theorem:asymptotic_private_distributed_mle} 
\begin{equation}
    \ev {} {\left \| \ovec \psi_t(\hat \theta, \check \theta) - \frac {2^t}{n} Z^\GM(\hat \theta, \check \theta) \one \right \|_2}  \le \frac {1} {\sqrt K} \sum_{k = 1}^K \ev {} {\left \| \ovec \psi_{k, t}(\hat \theta, \check \theta) - \frac {2^t} {n} Z_k(\hat \theta, \check \theta) \one \right \|_2}. \label{eq:gm_error_2} 
\end{equation}

For every round $k \in [K]$,
\begin{align*}
    \left \| \ovec \psi_{k, t}(\hat \theta, \check \theta) - \frac {2^t} {n} Z_k(\hat \theta, \check \theta) \one \right \|_2 & = \left \| \sum_{i = 2}^n \left ( \frac {\lambda_i(I+A)} 2 \right )^t \ovec l_i \ovec r_i^T \ovec \zeta_k(\hat \theta, \check \theta)  \right \|_2 \\ 
    & \le \sum_{i = 2}^n \left | \frac {\lambda_i(I+A)} 2 \right |^t \| \ovec l_i \|_2 \left | \ovec r_i^T \ovec \zeta_k(\hat \theta, \check \theta) \right | \\
    & \le \sum_{i = 2}^n \left | \frac {\lambda_i(I+A)} 2 \right |^t \left \| \ovec \zeta_k (\hat \theta, \check \theta) \right \|_2 
    \\ & \le \sum_{i = 2}^n \left | \frac {\lambda_i(I+A)} 2 \right |^t \left ( \left \| \ovec \lambda (\hat \theta, \check \theta) \right \|_2 + \left \| \ovec \kappa_k (\hat \theta, \check \theta) \right \|_2\right ). 
\end{align*}

Taking expectations and applying Jensen's inequality, we can bound the above as
\begin{equation} \label{eq:exp_residual}
    \ev {} { \left \| \ovec \psi_{k, t}(\hat \theta, \check \theta) - \frac {2^t} {n} Z_k(\hat \theta, \check \theta) \one \right \|_2} \le 2 (n - 1) \left | \frac {\lambda_2(I+A)} 2 \right |^t \left [ n \Gamma_{n, \Theta} + \sum_{i = 1}^n \sqrt {\var {} {\vec d_{i, k}(\hat \theta)}} \right ].
\end{equation}

Using $a_n^\star = \left | \frac {\lambda_2(I+A)} 2 \right |$, and combining  \cref{eq:gm_error_2} with \cref{eq:exp_residual}, we get:
\begin{equation*}
    \ev {} {\left \| \ovec \psi_t(\hat \theta, \check \theta) - \frac {2^t}{n} Z^\GM(\hat \theta, \check \theta) \one \right \|_2}  \le  \frac {2(n - 1) (a_n^\star)^t [n \Gamma_{n, \Theta} + V_{n, \eps, \Theta}]} {\sqrt K}.
\end{equation*}

By Markov's inequality and \cref{eq:gm_error_2}, we get that for every $z > 0$
\begin{equation*}
    \Pr \left [ \left \| \ovec \psi_t^\GM(\hat \theta, \check \theta) - \frac {2^t} {n} Z^\GM(\hat \theta, \check \theta) \one \right \|_2 > z  \right ] \le \frac {2(n - 1) (a_n^\star)^t [n \Gamma_{n, \Theta} + V_{n, \eps, \Theta}]} {z \sqrt K}. \label{eq:tail_bound}
\end{equation*}

We let the RHS be equal to $\alpha / (|\Theta^\star||\compTheta|)$, which corresponds to letting $z = |\compTheta| |\Theta^\star | \frac {2(n - 1) (a_n^\star)^t [n \Gamma_{n, \Theta} + V_{n, \eps, \Theta}]} {\alpha  \sqrt K}$. By applying a union bound over $\compTheta$ and $\Theta^\star$ we have that for every $\bar \theta \in \compTheta$ and $\theta^\star \in \Theta^\star$, with probability at least $1 - \alpha$, 
\begin{equation*}
    \vec \psi_{i, t}^\GM(\bar \theta, \theta^\star)  \le \frac {2^t} {n} Z^\GM(\bar \theta, \theta^\star) + |\compTheta||\Theta^\star| \frac {2(n - 1) (a_n^\star)^t [n \Gamma_{n, \Theta} + V_{n, \eps, \Theta}]} {\alpha  \sqrt K}. 
\end{equation*}

Conditioned on $F^c$, the above becomes
\begin{align}
    \vec \psi_{i, t}^\GM(\bar \theta, \theta^\star) & \le \frac {2^t} {n} \Lambda(\bar \theta, \theta^\star) + |\compTheta| |\Theta^\star| \frac {2(n - 1) (a_n^\star)^t [n \Gamma_{n, \Theta} + V_{n, \eps, \Theta}]} {\alpha  \sqrt K} \nonumber \\
    & \le - \frac {2^t} {n} l_{n, \Theta} + |\Theta|^2 \frac {(n - 1) (a_n^\star)^t [n \Gamma_{n, \Theta} + V_{n, \eps, \Theta}]} {2 \alpha  \sqrt K} \label{eq:gm_upper_bound_final}.
\end{align}

Note that we have used $|\compTheta| |\Theta^\star|  = (|\Theta| -  |\Theta^\star|) |\Theta^\star|  \le  |\Theta|^2/4$. To make \cref{eq:gm_upper_bound_final} at most $-\varrho$ for some $\varrho > 0$, we need to set
\begin{equation*}
    t \ge \max \left \{ \frac {\log \left ( \frac {2 \varrho n} {l_{n, \Theta}} \right )} {\log 2}, \frac {\log \left ( \frac {|\Theta|^2 (n-1) (n\Gamma_{n, \Theta} + V_{n, \eps, \Theta})} {2 \alpha \varrho \sqrt K} \right )} {\log (1 / a_n^\star)}\right \} = T.
\end{equation*}

The log-belief ratio threshold implies that for all $\theta^\star \in \Theta^\star, \bar \theta \in \compTheta$ we have that $\vec \nu_{i, k, T}(\theta^\star) \ge e^{\varrho} \vec \nu_{i, k, T} (\bar \theta)$. To determine the belief threshold $\tau$, note that 
\begin{align*}
    1 & = \sum_{\hat \theta \in \hat \Theta^\GM_{i, T}} \vec \nu_{i, k, T} (\hat \theta) + \sum_{\hat \theta \notin \hat \Theta^\GM_{i, T}} \vec \nu_{i, k, T} (\hat \theta) \ge (1 + e^\varrho) \max_{\hat \theta \notin \Theta^\GM_{i, T}} \vec \nu_{i, k, T} (\hat \theta) \\
    & \implies \max_{\hat \theta \notin \Theta^\GM_{i, T}} \vec \nu_{i, k, T} (\hat \theta) \le \frac {1} {1 + e^\varrho}.
\end{align*}

Moreover, we can prove that $\min_{\hat \theta \in \hat \Theta^\GM_{i, T}} \vec \nu_{i, k, T}(\hat \theta) \ge \frac {1} {1 + e^{-\varrho}} \ge \frac {1} {1 + e^\varrho}$ since $\varrho > 0$ which shows that any value in $[1 / (1 + e^\varrho), 1 / (1 + e^{-\varrho})]$ is a valid threshold. This yields that $\Pr [\hat \Theta^\GM_{i, T} \subseteq \Theta^\star | F^c] = \Pr [\hat \Theta^\GM_{i, T} \cap \compTheta = \emptyset | F^c] = \Pr [\exists \bar \theta \in \compTheta : \vec \nu_{i, T}^\GM(\bar \theta) > \tau | F^c] \ge 1 - \alpha$. Subsequently, by letting $K = |\Theta^\star| \log (|\compTheta| / \alpha)$, we get that $$\Pr [\hat \Theta^\GM_{i, T} \subseteq \Theta^\star] \ge \Pr [\hat \Theta^\GM_{i, T} \subseteq \Theta^\star | F^c] \Pr [F^c] \ge (1 - \alpha)^2 \ge 1 - 2 \alpha.$$

Finally, we show that $\hat{\Theta}^{\mathrm{GM}}_{i,T}$ is non-empty with high probability.
Conditioned on $F^c$, we have that $\hat{\Theta}^{\mathrm{GM}}_{i,T} \neq \emptyset$ since at least
one $\theta^\star \in \Theta^\star$ satisfies $\boldsymbol{\nu}^{\mathrm{GM}}_{i,T}(\theta^\star) \geq \tau$.
Using the law of total probability and the bound $\mathrm{P}[F] \leq |\bar{\Theta}|e^{-K/|\Theta^\star|}$,
we get
\begin{equation*}
\mathrm{P}[\hat{\Theta}^{\mathrm{GM}}_{i,T} \neq \emptyset]
\geq \mathrm{P}[\hat{\Theta}^{\mathrm{GM}}_{i,T} \neq \emptyset \mid F^c] \mathrm{P}[F^c]
\geq (1-\alpha)(1 - |\bar{\Theta}|e^{-K/|\Theta^\star|})
\geq 1 - 2\alpha - |\Theta^\star|e^{-K/|\Theta^\star|},
\end{equation*}
where the last step uses $K = |\Theta^\star|\log(|\bar{\Theta}|/\alpha)$, so
$|\bar{\Theta}|e^{-K/|\Theta^\star|} = \alpha$.

\medskip

\noindent \textbf{AM estimator.} We let $E = \left \{ \exists k \in [K], \bar \theta \in \compTheta, \theta^\star \in \Theta^\star : Z_k(\theta^\star, \bar \theta) < \Lambda(\theta^\star, \bar \theta) \right \}$. Using similar arguments to \cref{theorem:non_asymptotic_private_distributed_mle}, we can deduce that $\Pr [E] \le |\Theta^\star| e^{-K/|\compTheta|}$. By setting $K = |\compTheta| \log (|\Theta^\star| / (1 - \beta))$, we make $\Pr [E] \le 1 - \beta$. 

Conditioned on $E^c$ and by applying Markov's inequality, similarly to \cref{eq:gm_upper_bound_final} for the GM estimator, we have that for all $\theta^\star \in \Theta^\star, \bar \theta \in \compTheta$ and run $k \in [K]$
\begin{align} \label{eq:am_lower_bound}
    \vec \psi_{i, k, t}(\theta^\star, \bar \theta) & \ge \frac {2^t} {n} Z_k(\theta^\star, \bar \theta) - |\Theta^\star | |\compTheta | \frac {2 (n - 1) (a_n^\star)^t (n \Gamma_{n, \Theta} + V_{n, \eps, \Theta})} {1 - \beta'} \\
    & \ge \frac {2^t} {n} \Lambda(\theta^\star, \bar \theta) - |\Theta|^2 \frac {(n - 1) (a_n^\star)^t (n \Gamma_{n, \Theta} + V_{n, \eps, \Theta})} {2 (1 - \beta')} \nonumber \\
    & \ge \frac {2^t} {n} l_{n, \Theta} - |\Theta|^2 \frac {2(n - 1) ( a_n^\star)^t (n \Gamma_{n, \Theta} + V_{n, \eps, \Theta})} {2 (1 - \beta')}, \nonumber
\end{align} with probability $\beta'$ for some $\beta' \in (0, 1)$. To make the above at least $\varrho$ for some $\varrho > 0$ it suffices to pick 
\begin{equation*}
    t \ge \max \left \{ \frac {\log \left ( \frac {2 \varrho n} {l_{n, \Theta}} \right )} {\log 2}, \frac {\log \left ( \frac {|\Theta|^2 (n-1) (n\Gamma_{n, \Theta} + V_{n, \eps, \Theta})} {2 (1 - \beta') \varrho}  \right )} {\log (1 / a_n^\star)} \right \} = T.
\end{equation*}

The log-belief ratio threshold implies that for all $\theta^\star \in \Theta^\star, \bar \theta \in \compTheta$ we have that $\vec \nu_{i, k, T}(\theta^\star) \ge e^{\varrho} \vec \nu_{i, k, T} (\bar \theta)$. Similarly, any value in $[1/(1 + e^\varrho), 1/(1 + e^{-\varrho})]$ is a valid threshold, we can show that 

\begin{align*}
    \Pr [\Theta^\star \subseteq \hat \Theta^\AM_{i, T}  | E^c ] & = \Pr \left [ \forall \theta^\star \in \Theta^\star : \vec \nu^\AM_{i, T}(\theta^\star) > {\tau} \bigg | E^c \right ] \\
    & \ge \Pr \left [ \bigcup_{k \in [K]} \left \{ \forall \theta^\star \in \Theta^\star : \vec \nu_{i, k, T} (\theta^\star) > \tau \right \} \bigg | E^c \right ] \\
    & = 1 - \Pr \left [ \bigcap_{k \in [K]} \left \{ \exists \theta^\star \in \Theta^\star : \vec \nu_{i, k, T}(\theta^\star) < \tau \right \} \bigg | E^c \right ] \\
    & = 1 - \left ( \Pr \left [ \exists \theta^\star \in \Theta^\star : \vec \nu_{i, k, T} (\theta^\star) < \tau \bigg | E^c \right ] \right )^K \\
    & = 1 - \left (1 -  \Pr \left [ \forall \theta^\star \in \Theta^\star : \vec \nu_{i, k, T} (\theta^\star) > \tau \bigg | E^c \right ] \right )^K \\
    & = 1 - (\beta')^K \\
    & \ge 1 - e^{-(1 - \beta') K}.
\end{align*}

We set $1 - \beta' = \log (1 / (1 - \beta)) / K$ and have that $$\Pr [\Theta^\star \subseteq \hat \Theta^\AM_{i, T}] \ge \Pr [E^c]  \Pr [\Theta^\star \subseteq \hat \Theta^\AM_{i, T}  | E^c ] \ge (\beta)^2 \ge 1 - 2 (1 - \beta).$$ These finally yield
\begin{equation*}
    T = \max \left \{ \frac {\log \left ( \frac {2 \varrho n} {l_{n, \Theta}} \right )} {\log 2}, \frac {\log \left ( \frac {|\Theta|^2 (n-1) K (n\Gamma_{n, \Theta} + V_{n, \eps, \Theta})} {2 \log (1 / (1 - \beta)) \varrho}  \right )} {\log (1 / a_n^\star)} \right \}. 
\end{equation*}

Similarly, we show that $\hat{\Theta}^{\mathrm{AM}}_{i,T} \neq \Theta$ with high probability.
Conditioned on $E^c$, at least one $\bar{\theta} \in \bar{\Theta}$ satisfies
$\boldsymbol{\nu}^{\mathrm{AM}}_{i,T}(\bar{\theta}) < \tau$, so $\hat{\Theta}^{\mathrm{AM}}_{i,T} \neq \Theta$.
Using the law of total probability and the bound $\mathrm{P}[E] \leq |\Theta^\star|e^{-K/|\bar{\Theta}|}$,
we get
\begin{equation*}
\mathrm{P}[\hat{\Theta}^{\mathrm{AM}}_{i,T} \neq \Theta]
\geq \mathrm{P}[\hat{\Theta}^{\mathrm{AM}}_{i,T} \neq \Theta \mid E^c] \mathrm{P}[E^c]
\geq (1-(1-\beta))(1 - |\Theta^\star|e^{-K/|\bar{\Theta}|})
\geq 2\beta - 1 - e^{-K/|\bar{\Theta}|},
\end{equation*}
where the last step uses $K = |\bar{\Theta}|\log(|\Theta^\star|/(1-\beta))$, so
$|\Theta^\star|e^{-K/|\bar{\Theta}|} = 1-\beta$.

\noindent \textbf{Privacy (for both estimators).} Minimizing the convergence time $T$ corresponds to minimizing $V_{n, \eps, \Theta}$. Since $V_{n, \eps, \Theta}$ is separable over the agents, it suffices to solve the problem of minimizing the variance of each noise variable independently. 

If an adversary can eavesdrop only once and has access to any round $k \in [K]$, and we have $|\Theta|$ states, by the composition theorem, the budget $\eps$ should be divided by $K |\Theta|$. Thus, the problem of finding the optimal distribution $\cD_i (\eps)$ on $\Rbb$, corresponds to the following optimization problem studied in \cite{koufogiannis2015optimality,papachristou2023differentially}, for all $k \in [K]$: 
\begin{align*}
    \min_{{\cD_i(\eps)} \in \mathrm {Simplex}(\Rbb)} & \quad  \var {\vec d_{i, k} \sim \cD_i(\eps)} {\vec d_{i, k}} \\
    \text{s.t.} & \quad  \cD_i(\eps) \text { is $\frac {\eps} {K |\Theta|}$-DP and $\ev {\vec d_{i, k} \sim \cD_i(\eps)} {\vec d_{i, k}} = 0$}. \nonumber
\end{align*} 

The optimal solution to this problem is given by \citet{koufogiannis2015optimality,papachristou2023differentially}. For privatizing beliefs in our problem, this corresponds to selecting $\cD_i^\star(\eps) = \mathrm{Lap} \left ( \frac {\Delta_{i, \Theta} K |\Theta|} {\eps} \right )$ where $\Delta_{i, \Theta}$ is the sensitivity of the log-likelihood of the $i$-th agent. 

\noindent \textbf{Differentially Private Outputs. } Due to the postprocessing property of DP (see Proposition 2.1 of \citet{dwork2014algorithmic}), the resulting estimates are $\eps$-DP with respect to the private datasets.

\qed

\subsection{Proof of \texorpdfstring{\cref{theorem:asymptotic_threshold}}{asymptotic-threshold}: Asymptotic Behavior of the Two-Threshold Algorithm}\label{sec:app:proof:theorem:asymptotic_threshold}

    For simplicity, we define $p_2 = 1 / |\compTheta|$ and $p_1 = 1 - 1 / |\Theta^\star|$, and assume that $\varrho^{\thres, 1} = \varrho^{\thres, 2} = \varrho$ which implies that $\hattauthres{1} = \hattauthres{2} =  \tau$ and $\vec N_{i, t}^{\thres, 1}(\hat \theta) = \vec N_{i, t}^{\thres, 2}(\hat \theta) = \vec N_{i, T}(\hat \theta)$.

    \noindent \textbf{Type I estimator ($\hat \Theta^{\thres, 1}_{i, T}$).} To determine the asymptotic Type I error probability $\alpha$ of $\hat \Theta^{\thres, 1}_{i, T}$, we assume that the threshold takes the form $\tauthres{1} = (1 + \pi_1) p_1$. Then
    \begin{align*} 
        \lim_{T \to \infty} \Pr \left [ \hat \Theta^{\thres, 1}_{i, T} \not \subseteq \Theta^\star \right ] 
        & \ensuremath{\stackrel{\text{(i)}}{=}} \lim_{T \to \infty} \Pr \left [ \exists \bar \theta \in \compTheta : \bar \theta \in \hat \Theta^{\thres, 1}_{i, T} \right ] \\
        & \ensuremath{\stackrel{\text{(ii)}}{\le}}  \lim_{T \to \infty} \sum_{\bar \theta \in \compTheta} \Pr \left [ \bar \theta \in \hat \Theta^{\thres, 1}_{i, T} \right ] \nonumber \\
        & \ensuremath{\stackrel{\text{(iii)}}{\le}} \lim_{T \to \infty} \sum_{\bar \theta \in \compTheta} \Pr \left [ \vec N_{i, T}(\bar \theta) \ge \tauthres{1} \right ] \nonumber \\
        & \le  \lim_{T \to \infty} \sum_{\bar \theta \in \compTheta} \Pr \left [ \vec N_{i, T}(\bar \theta) \ge (1 + \pi_1) p_1 \right ] \nonumber \\
        & \ensuremath{\stackrel{\text{(iv)}}{\le}}  \lim_{T \to \infty} \sum_{\bar \theta \in \compTheta} \Pr \left [ \vec N_{i, T}(\bar \theta) \ge (1 + \pi_1) \ev {} {N_{i, T} (\bar \theta)} \right ] \nonumber \\
        & \ensuremath{\stackrel{\text{(v)}}{\le}} |\compTheta| e^{-2K\pi_1^2}, \nonumber
    \end{align*} where the result is derived by applying (i) the definition of $\hat \Theta^{\thres, 1}_{i, T}$, (ii) union bound, (iii) the definition of the threshold $\tauthres{1} = (1 + \pi_1) p_1$, (iv) the fact that $\ev {} {\vec N_{i, T}(\bar \theta)} \le p_1$ for all $\bar \theta \in \compTheta$ as $T \to \infty$, and the Chernoff bound on $\vec N_{i, T}(\bar \theta)$. Therefore, to make the above $\alpha$, it suffices to choose $K = \frac {\log (|\compTheta| / \alpha)} {2 \pi_1^2}$.

    % \noindent \textbf{Combining the guarantees.} In the sequel, we combine the guarantees found above. Specifically, we consider two cases: the former case considers the perfect distinguishability of the MLE states from the non-MLE states, and the latter case considers the case when MLE states:

    % \begin{itemize}
    %     \item \emph{Distinguishability of states.} In order for the MLE states to be separable from the non-MLE states, it suffices to require $1 - p_2 \le p_1$. We solve the inequality and get:  

    %     In this case we can use a single threshold which corresponds to setting $\tauthres{2} = \tauthres{1}$, which implies that $\hat \Theta^{\thres, 2}_{i, T} = \hat \Theta^{\thres, 1}_{i, T} = \hat \Theta^{\thres}_{i, T}$, namely: 

    %     \begin{equation*}
    %         \pi_1 = (1 - \pi_2) \frac {p_1} {1 - p_2} - (1 - p_2).
    %     \end{equation*}

    %     In this case, setting $\alpha = 1 - \beta = \eta$ and $K = \max \left \{  \frac {\log (|\Theta^\star| / \eta)} {2 \pi_2^2}, \frac {\log (|\compTheta| / \eta)} {2 \pi_1^2} \right \}$, and taking the union bound over the type I and type II events (i.e., \cref{eq:type_1,eq:type_2}) we get an accuracy guarantee

    %     \begin{equation*}
    %         \lim_{T \to \infty} \Pr \left [ \hat \Theta^{\thres}_{i, T}  \neq \Theta^\star \right ] \le 2 \eta.
    %     \end{equation*}

    %     \item In all the other cases 

    % \end{itemize}

    \noindent \textbf{Type II estimator ($\hat \Theta^{\thres, 2}_{i, T}$).} To determine the asymptotic Type II error probability $1 - \beta$ of $\hat \Theta^{\thres, 2}_{i, T}$, we assume that the threshold takes the form $\tauthres{2} = (1 - \pi_2) p_2$. Then
    \begin{align*} 
        \lim_{T \to \infty} \Pr \left [ \Theta^\star \not \subseteq \hat \Theta^{\thres, 2}_{i, T} \right ] 
        & \ensuremath{\stackrel{\text{(i)}}{=}}  \lim_{T \to \infty} \Pr \left [ \exists \theta^\star \in \Theta^\star : \theta^\star \notin  \Theta^{\thres, 2}_{i, T} \right ] \\
        & \ensuremath{\stackrel{\text{(ii)}}{\le}} \lim_{T \to \infty} \sum_{\theta^\star \in \Theta^\star} \Pr \left [ \theta^\star \notin  \Theta^{\thres, 2}_{i, T} \right ] \nonumber \\
        & \ensuremath{\stackrel{\text{(iii)}}{\le}} \lim_{T \to \infty} \sum_{\theta^\star \in \Theta^\star} \Pr \left [ \vec N_{i, T}(\theta^\star) \le \tauthres{2} \right ] \nonumber \\
        & \ensuremath{\stackrel{\text{(iii)}}{\le}}  \lim_{T \to \infty} \sum_{\theta^\star \in \Theta^\star} \Pr \left [ \vec N_{i, T}(\theta^\star) \le (1 - \pi_2) p_2 \right ] \nonumber \\
        & \ensuremath{\stackrel{\text{(iv)}}{\le}} \lim_{T \to \infty} \sum_{\theta^\star \in \Theta^\star} \Pr \left [ \vec N_{i, T}(\theta^\star) \le (1 - \pi_2) \ev {} {\vec N_{i, T}(\theta^\star)} \right ] \nonumber \\
        & \ensuremath{\stackrel{\text{(v)}}{\le}} |\Theta^\star| e^{-2K \pi_2^2},  \nonumber
    \end{align*} where the result is derived by applying (i) the definition of $\hat \Theta^{\thres, 2}_{i, T}$, (ii) union bound, (iii) the definition of the threshold $\tauthres{2} = (1 - \pi_2) p_2$, (iv) the fact that $\ev {} {\vec N_{i, T}(\theta^\star)} \ge p_2$ for all $\theta^\star \in \Theta^\star$ as $T \to \infty$, and (v) the Chernoff bound on $\vec N_{i, T}(\hat \theta)$. Therefore, to make the above less than $1 - \beta$, it suffices to choose $K = \frac {\log (|\compTheta| / (1 - \beta))} {2 \pi_2^2}$.

\noindent \textbf{Differentially Private Outputs. } Due to the immunity to post-processing of DP \citep[Proposition 2.1]{dwork2014algorithmic}, the resulting estimates are $\eps$-DP with respect to private datasets.

\qed

\subsection{Proof of \texorpdfstring{\cref{theorem:non_asymptotic_threshold}}{non-asymptotic-threshold}: Finite-Time Guarantees for the Two-Threshold Algorithm}\label{sec:app:proof:theorem:non_asymptotic_threshold}

For simplicity, we assume that $\varrho^{\thres, 1} = \varrho^{\thres, 2} = \varrho$ which implies that $\hattauthres{1} = \hattauthres{2} =  \tau$ and $\vec N_{i, t}^{\thres, 1}(\hat \theta) = \vec N_{i, t}^{\thres, 2}(\hat \theta) = \vec N_{i, T}(\hat \theta)$.

\noindent \textbf{Type I estimator.} From the analysis of \cref{theorem:non_asymptotic_private_distributed_mle} (see \cref{eq:gm_upper_bound_final}), setting 
$$t \ge \max \left \{ \frac {\log \left ( \frac {2 \varrho n} {l_{n, \Theta}} \right )} {\log 2}, \frac {\log \left ( \frac {(n - 1) |\Theta|^2 (n\Gamma_{n, \Theta} + V_{n, \eps, \Theta})} {q_1 \varrho} \right )} {\log (1 / a_n^\star)}\right \} = T,$$ 
guarantees that $\ev {} {\vec N_{i, T} (\bar \theta)} = \Pr \left [\vec \nu_{i, k, T}(\bar \theta) \le \tau \right ] \le q_1$ for all $\bar \theta \in \compTheta$. Then following the analysis similar to \cref{theorem:non_asymptotic_threshold}, using the union bound and the Chernoff bound, we can prove that $\Pr [\Theta^{\thres, 1}_{i, T} \subseteq \hat \Theta^\star] \le |\compTheta| e^{-2K \pi_1^2}$. Subsequently, setting $K \ge \frac {\log (|\Theta^\star| / \alpha)} {2 \pi_1^2}$ makes the error to be at most $\alpha$. 

\noindent \textbf{Type II estimator.} From the analysis of \cref{theorem:non_asymptotic_private_distributed_mle} (see \cref{eq:am_lower_bound}), setting 
$$t \ge\max \left \{ \frac {\log \left ( \frac {2 \varrho n} {l_{n, \Theta}} \right )} {\log 2}, \frac {\log \left ( \frac {(n-1) |\Theta|^2 (n\Gamma_{n, \Theta} + V_{n, \eps, \Theta})} {2 (1 - q_2) \varrho}  \right )} {\log (1 / a_n^\star)} \right \} = T,$$ 
guarantees that $\ev {} {\vec N_{i, T} (\theta^\star)} = \Pr \left [\vec \nu_{i, k, T}(\theta^\star) > \tau \right ] \ge q_2$ for all $\theta^\star \in \Theta^\star$. Then following a similar analysis to \cref{theorem:non_asymptotic_threshold} using the union bound and the Chernoff bound, we can prove that $\Pr [\Theta^\star \not \subseteq \hat \Theta^{\thres, 2}_{i, T}] \le |\Theta^\star| e^{-2K \pi_2^2}$. Subsequently, setting $K \ge \frac {\log (|\Theta^\star| / (1 - \beta))} {2 \pi_2^2}$ makes the error at most $1 - \beta$. 

\noindent \textbf{Optimal Distributions.} The proof is exactly the same as in \cref{theorem:non_asymptotic_private_distributed_mle}.

\noindent \textbf{Differentially Private Outputs.} Due to the post-processing property of DP, see \citet[Proposition 2.1]{dwork2014algorithmic}, the resulting estimates are $\eps$-DP with respect to private datasets.

\qed

\clearpage

\section{Proofs and Convergence Analysis for Private Distributed Hypothesis Testing} \label{sec:app:hypothesis_testing}

\subsection{Proof of \texorpdfstring{\cref{prop:hypothesis_testing}}{theorem:hypothesistesting}: Simple Hypothesis Testing at Significance Level \texorpdfstring{$\alpha$}{alpha}}\label{sec:app:proof:theorem:hypothesis_testing}

Let $0< \alpha', \alpha'' < 1, \alpha'+ \alpha'' = \alpha$ to be determined later. The centralized UMP test at level $\alpha''$ defines a threshold $\varrho_c$ such that $\theta = 0$ is rejected if $2\Lambda(1, 0) \ge \varrho_c$. The threshold $\varrho_c$ is selected such that $\Pr [2 \Lambda(1, 0) \ge \varrho_c| \theta = 0] = \alpha''$. For the decentralized test, by the GM algorithm convergence analysis, we know that after $K$ runs and $T$ iterations given by the GM algorithm with Type I guarantee $\alpha'$ and threshold $\varrho^\GM = 1$ we have the following event
\begin{equation*}
    E = \left \{ \left | \frac {n} {2^{T - 1}} \vec \psi_{i, T}(1, 0) -  2\Lambda(1, 0) \right | \le \frac {n} {2^{T - 1}}  \right \},
\end{equation*}

Thus, for these values of $T, K$ we set $\varrho_d = \varrho_c - 1$. From the above relations, we get that under $\theta = 0$ and $E$: $\left \{ 2\Lambda(1, 0) \ge \varrho_c \right \} \implies \left \{ \vec \psi_{i, T}(1, 0) \ge \varrho_d\right \}$. This means that if the centralized test rejects, then the decentralized test also rejects $\theta = 0$ with probability at least $1 - \alpha'$. The Type I error is: 
\begin{align*}
    \Pr [\vec \psi_{i, t}^\GM(1, 0) \ge \varrho_d | \theta = 0] & = \Pr [E]  \Pr [\vec \psi_{i, t}^\GM (1, 0) \ge \varrho_d | \theta = 0, E] \\ &  + \Pr [E^c]  \Pr [\vec \psi_{i, t}^\GM (1, 0) \ge \varrho_d | \theta = 0, E^c] \\ & \le \alpha'' + \alpha'.
\end{align*}

We choose $\alpha' = \alpha/2, \alpha'' = \alpha/2$ such that $\alpha' + \alpha'' = \alpha$. The privacy guarantee is a direct consequence of DP's post-processing property. 

\qed

\subsection{Extension to Composite Hypotheses}\label{sec:app:composite-extension}

The extension to composite hypotheses is straightforward since, if we consider the generalized likelihood ratio test and run the GM algorithm with the parameters of \cref{prop:hypothesis_testing} we would obtain that for $T$ and $K$ set as in above, with probability at least $1 - \alpha/2$ the log-belief ratio statistic satisfies

$$\left | \frac {n} {2^{T - 1}} \vec \psi_{i, T}^\GM(\tilde \Theta, \tilde \Theta_0) - \sum_{i \in [n]} 2  \log \left ( \frac {\sup_{\tilde \theta_i \in \tilde \Theta} \ell_i(\vec s_i | \tilde \theta_1)} {\sup_{\tilde \theta_0 \in \tilde \Theta_0} \ell_i(\vec s_i | \tilde \theta_0)} \right ) \right | \le  \frac {n} {2^{T - 1}},$$
where the centralized generalized log-likelihood ratio statistic $\sum_{i \in [n]} 2  \log \left ( \sup_{\tilde \theta_i \in \tilde \Theta} \ell_i(\vec s_i | \tilde \theta_1) / \sup_{\tilde \theta_0 \in \tilde \Theta_0} \ell_i(\vec s_i | \tilde \theta_0) \right )$ converges in distribution to $\chi^2_n$ for sufficiently large $n_i$.

If $\varrho_d = F^{-1}_{\chi^2_n} (1 - \alpha/2) - 1$, then the composite hypothesis test is $\eps$-DP and has Type I error at most $\alpha$. 

\clearpage

\section{Proofs and Convergence Analysis for Private Distributed Online Learning}\label{sec:app:online-learning}

\subsection{Log-Belief Ratio Notations}

Let $\vec \mu_{i, t}(\hat \theta)$ be the beliefs for the non-private system, and let $\vec \nu_{i, t}(\hat \theta)$ be the private estimates for agent $i \in [n]$ and round $t \ge 0$. For the non-private algorithm (\cref{alg:non_private_intermittent_streams}) all pairs $\hat \theta, \check \theta \in \Theta$ we let $\vec \phi_{i, t}(\hat \theta, \check \theta) = \log \left ( \frac {\vec \mu_{i, t}(\hat \theta)} {\vec \mu_{i, t}(\check \theta)} \right )$ and $\vec \lambda_{i, t} (\hat \theta, \check \theta) =  \log \left ( \frac {\vec \gamma_{i, t}(\hat \theta)} {\vec \gamma_{i, t}(\check \theta)} \right )$ be the log belief ratios and the log-likelihood ratios respectively for agent $i \in [n]$ round $t \ge 0$. For the private algorithm (\cref{alg:private_online_learning}) we let $\vec \psi_{i, t}(\hat \theta, \check \theta)  = \log \left ( \frac {\vec \nu_{i, t}(\hat \theta)} {\vec \nu_{i, t}(\check \theta)} \right ), \vec \zeta_{i, t}(\hat \theta, \check \theta)  = \log \left ( \frac {\vec \sigma_{i, t}(\hat \theta)} {\vec \sigma_{i, t}(\check \theta)} \right ),$ and $\vec \kappa_{i, t}(\hat \theta, \check \theta) = \vec d_{i, t} (\hat \theta) - \vec d_{i, t} (\check \theta)$ be the private log-belief ratio, log-likelihood ratio, and noise difference ratios respectively for agent $i \in [n]$, and round $t \ge 0$. We also let the vectorized versions $\ovec \phi_{t}(\hat \theta, \check \theta), \ovec \psi_{t}(\hat \theta, \check \theta), \ovec \lambda_t(\hat \theta, \check \theta), \ovec \zeta_t(\hat \theta, \check \theta)$ respectively, where the vectorization is over the agents $i \in [n]$. Finally, for each agent $i \in [n]$ and pair of states $\hat \theta, \check \theta \in \Theta$ we define the KL divergence between the states $\Lambda_i(\hat \theta, \check \theta) = \ev {\theta} {\log \left ( \frac {\ell_i(\vec s_{i, 0} | \hat \theta)} {\ell_i(\vec s_{i, 0} | \check \theta)} \right )} = \kldiv {\ell_i(\cdot | \hat \theta)} {\ell_i(\cdot | \check \theta)}$. 

\subsection{Auxiliary Lemmas}

Before proving the main result for online learning, we prove the following lemma regarding the rate of convergence of the C\'esaro means. 

\begin{lemma} \label{lemma:intermittent_non_asymptotic_helper}
    Let $\ovec X_0, \dots, \ovec X_{t-1}$ be i.i.d. random variables (vectors), let $A$ be an irreducible and aperiodic (primitive) doubly stochastic matrix with the second largest eigenvalue modulus $|\lambda_2(A)|<1$, and let $\vec x_\tau =  {\frac 1 n \one^T \ovec X_\tau}$ for all $0 \le \tau \le t-1$, with $\max_{1 \le \tau \le t} \sqrt {\var {} {\vec x_\tau}} \le V$. Then
    \begin{equation*}
    \ev {} {\left \| \frac 1 t \sum_{\tau = 0}^{t - 1} A^{t - \tau} \ovec X_\tau - \left ( \frac 1 t \sum_{\tau = 0}^{t - 1} \vec x_\tau \right ) \one \right \|_2} \le  \frac {(n - 1) V} {(1 - |\lambda_2(A)|) t}.
    \end{equation*}
\end{lemma}
 
{\it Proof.} For $0 \le  \tau  \le  t-1$, we have 
    \begin{align*}
        \ev {} {\left \| A^{t - \tau} \ovec X_\tau -  \vec x_\tau \one \right \|_2} & = \ev {} {\left \| \frac 1 n \one \one^T \ovec X_\tau + \sum_{i = 2}^n \lambda_i(A)^{t - \tau} \ovec r_i \ovec l_i^T \ovec X_\tau - \vec x_{\tau} \one \right \|_2} \\
        & = \ev {} {\left \| \vec x_\tau \one + \sum_{i = 2}^n \lambda_i(A)^{t - \tau} \ovec r_i \ovec l_i^T \ovec X_\tau - \vec x_\tau \one \right \|_2} \\
        & \le \sum_{i = 2}^n \left | \lambda_i(A) \right |^{t - \tau} \| \ovec r_i \|_2 \| \ovec l_i \|_2 \ev {} {\| \ovec X_\tau \|_2} \\
        & \le (n - 1) {\left | \lambda_2(A) \right |}^{t - \tau} \ev {} {\| \ovec X_\tau \|_2} \\
        & \le  (n - 1) {\left | \lambda_2(A) \right |}^{t - \tau}  \ev {} {\| \ovec X_0 \|_2} \\
        & \le  (n - 1) {\left | \lambda_2(A) \right |}^{t - \tau} V.
    \end{align*}

Therefore, 
\begin{equation*}
    \ev {} {\left \| \frac 1 t \sum_{\tau = 0}^{t - 1} A^{t - \tau} \ovec X_\tau - \left (\frac 1 t \sum_{\tau = 0}^{t - 1} \vec x_\tau \right ) \one \right \|_2} \le \frac {(n - 1) V} {(1 -\left | \lambda_2(A) \right |) t},
\end{equation*}
where we use $\sum_{\tau = 0}^t {\left | \lambda_2(A) \right |}^{t - \tau} \le \sum_{\tau \ge 0} {\left | \lambda_2(A) \right |}^\tau = \frac {1} {1 - \left | \lambda_2(A) \right |}$. \qed

In the sequel, we show the main result for online learning.

\subsection{Proof of \texorpdfstring{\cref{theorem:non_asymptotic_private_online_learning}}{alg-private-online-learning}: Finite-Time Guarantees for Private, Distributed Online Learning} \label{sec:app:proof:theorem:non_asymptotic_private_online_learning}

 %Similarly to the asymptotic case, it suffices to pick $K = 1$. 

The dynamics of the log-belief ratio obey $\ovec \psi_t(\hat \theta, \check \theta) = A \ovec \psi_{t - 1}(\hat \theta, \check \theta) + \ovec \zeta_{t}(\hat \theta, \check \theta) = \sum_{\tau = 0}^{t - 1} A^{t - \tau} \ovec \zeta_{\tau}(\hat \theta, \check \theta)$, where $\ovec \zeta_{\tau} (\hat \theta, \check \theta) = \ovec \lambda_\tau (\hat \theta, \check \theta) + \ovec \kappa_\tau(\hat \theta, \check \theta)$. We apply \cref{lemma:intermittent_non_asymptotic_helper}, and get that 
\begin{equation*}
    \ev {} {\left \| \frac 1 t \ovec \psi_t(\hat \theta, \check \theta) - \frac 1 {nt} \sum_{\tau = 0}^{t - 1} \sum_{i = 1}^n \vec \zeta_{i, \tau}(\hat \theta, \check \theta) \one \right \|_2}  \le  \frac {(n - 1) V_{n, \eps, \Theta}'} {t (1 - \left | \lambda_2(A) \right |)}.
\end{equation*}

where,
\begin{equation*}
    V_{n, \eps, \Theta}' = \max_{\bar \theta \neq\theta^{\circ}} \sqrt {\var {} {\frac 1 {nt} \sum_{i = 1}^n \sum_{\tau = 0}^{t - 1}  \vec \zeta_{i, \tau} (\bar \theta, \theta^{\circ}) }} \le \frac {\sqrt 2} {n} (n Q_{n, \Theta} + V_{n, \eps, \Theta}).
\end{equation*}

Moreover,
\begin{equation*}
    \ev {} {\left \| \frac 1 {nt} \sum_{\tau = 0}^{t - 1} \sum_{i = 1}^n \vec \zeta_{i, \tau}(\hat \theta, \check \theta) - \frac 1 {n} \sum_{i = 1}^n \Lambda_i(\hat \theta, \check \theta) \right \|_2 } \le \frac {V_{n, \eps, \Theta}'} {t}.
\end{equation*}

Therefore, by Markov's inequality and the triangle inequality, we get that for every $z > 0$
\begin{equation*}
    \Pr \left [ \left \| \frac 1 t \ovec \psi_t(\hat \theta, \check \theta) - \frac 1 n \sum_{i = 1}^n \Lambda_i(\hat \theta, \check \theta) \one \right \|_2 > z  \right ] \le \frac {V_{n, \eps, \Theta}'} {z (1 -  {\left | \lambda_2(A) \right |}) t}.
\end{equation*}

Therefore, we can show that by applying a union bound over $\Theta^\star$, we have that with probability $1 - \eta$, for all $\bar \theta \in \compTheta$
\begin{equation*}
    \log \vec \nu_{i, t}(\bar \theta) \le \vec \psi_{i, t}(\bar \theta, \theta^\star) \le - \frac {t} {n} l_{n, \Theta} + \frac {|\Theta| V_{n, \eps, \Theta}'} {2 \eta (1 - {\left | \lambda_2(A) \right |})}.
\end{equation*}

To make the RHS  the log-belief threshold at most some value $- \varrho$ for some $\varrho > 0$, we require
\begin{equation*}
    t \ge T = \frac {\varrho + \frac {|\Theta| V_{n, \eps, \Theta}'} {2 \eta (1 - {\left | \lambda_2(A) \right |})}} {l_{n, \Theta} / n}.
\end{equation*}

To determine $\varrho$, note that $\vec \nu_{i, t}(\bar \theta) \le e^{-\varrho}$ for all $\bar \theta \neq \theta^\star$, and, thus, 
\begin{equation*}
    \vec \nu_{i, t}(\theta^\star) = 1 - \sum_{\bar \theta \neq \theta^\star} \vec \nu_{i, t} (\bar \theta) \ge 1 - (|\Theta| - 1) e^{-\varrho}.
\end{equation*}

To determine a valid value of $\varrho$, we require $1 - (|\Theta| - 1) e^{-\varrho} \ge e^{-\varrho}$ which yields $\varrho \ge \log (|\Theta|)$. Therefore, setting $\varrho = \log (|\Theta|)$ and subsequently 
\begin{equation*}
    T = \frac {\log (|\Theta|) + \frac {|\Theta| \frac {\sqrt 2} {n} (n Q_{n, \Theta} + V_{n, \eps, \Theta})} {2 \eta (1 - {\left | \lambda_2(A) \right |})}} {l_{n, \Theta} / n} ,
\end{equation*} we get that $\Pr \left [\theta^\star = \hat \theta^\intt_{i, T} \right ] \ge 1 - \eta.$

\medskip

\noindent \textbf{Privacy.} Similarly to \cref{theorem:non_asymptotic_private_distributed_mle}, the optimal distributions are those that minimize variance subject to privacy constraints. Because there are $|\Theta|$ states, the budget $\eps$ should be divided by $|\Theta|$, resulting in $\cD_{i, t}^\star(\eps) = \mathrm{Lap} \left ( \frac {\Delta_{i, \Theta} |\Theta|} {\eps} \right )$. \qed

\noindent \textbf{Differentially Private Outputs. } Due to the post-processing property of DP (see Proposition 2.1 of \citet{dwork2014algorithmic}), the resulting estimates are $\eps$-DP with respect to the private datasets.

\clearpage

\section{Differentially Private Distributed Survival Analysis for Clinical Trials} \label{sec:app:clinical_trials}

\subsection{Sensitivity of the Proportional Hazards Model} \label{sec:app:sensitivity}

The released statistic is,

\begin{equation}
    \Lambda_i(\vec s_i) = \log \ell_i(\vec s_i \mid \theta_1, \vec\theta_\ctrl) - \log \ell_i(\vec s_i \mid 0, \vec\theta_\ctrl),
    \label{eq:cox_lr}
\end{equation}
which tests $H_0 : \theta_\trt = 0$ against $H_1 : \theta_\trt = \theta_1$ with the controls $\vec\theta_\ctrl$ held common to the two models. 

First, we take $\vec x_{ij} \in \{0,1\}$, $|\theta_1| \le B_\theta$, $\|\vec z_{ij}\| \le B_z$, and $\|\vec\theta_\ctrl\| \le B_{\vec\theta_\ctrl}$. Using $\log \ell_i(\theta) = \sum_{j:\delta_{ij}=1}\big[\eta_{ij}(\theta) - \log \sum_{r\in R(j)} e^{\eta_{ir}(\theta)}\big]$ with $\eta_{ir}(\theta) = \theta_\trt \vec x_{ir} + \vec\theta_\ctrl^T \vec z_{ir}$ and $R(j) = \{j' : \vec t_{ij'} \ge \vec t_{ij}\}$. The numerator terms regarding the controls cancel between the two log-likelihoods and
\begin{equation}
    \Lambda_i(\vec s_i) = \sum_{j:\delta_{ij}=1} \Big[\, \theta_1 \vec x_{ij} - \log \tfrac{S_j(\theta_1)}{S_j(0)} \,\Big], \qquad S_j(\theta_\trt) = \sum_{r\in R(j)} e^{\theta_\trt \vec x_{ir} + \vec\theta_\ctrl^T \vec z_{ir}}.
    \label{eq:cox_lr_terms}
\end{equation}

Second, because $\vec x_{ir} \in \{0,1\}$, split the risk mass into untreated and treated parts, $A_j = \sum_{r\in R(j):\,\vec x_{ir}=0} e^{\vec\theta_\ctrl^T \vec z_{ir}}$ and $B_j = \sum_{r\in R(j):\,\vec x_{ir}=1} e^{\vec\theta_\ctrl^T \vec z_{ir}}$, and let $\pi_j = B_j/(A_j+B_j) \in [0,1]$ be the treated fraction of the risk set. 

Then $S_j(\theta_\trt) = A_j + e^{\theta_\trt} B_j = (A_j+B_j)\big(1 - \pi_j + e^{\theta_\trt}\pi_j\big)$, so $\log \{S_j(\theta_1)/S_j(0)\} = \varphi(\pi_j)$ with
\begin{equation}
    \varphi(\pi) = \log\!\big(1 + (e^{\theta_1}-1)\,\pi\big), \qquad
    \Lambda_i(\vec s_i) = \sum_{j:\delta_{ij}=1} \big[\, \theta_1 \vec x_{ij} - \varphi(\pi_j) \,\big].
    \label{eq:cox_lr_pi}
\end{equation}

Each event contributes $\lambda_j = \theta_1 \vec x_{ij} - \varphi(\pi_j)$, which yields $|\lambda_j| \le |\theta_1| \le B_\theta$.

Next, we let $\vec s_i$ and $\vec s_i'$ differ in patient $k$. We have two cases:

\begin{enumerate}
    \item If $k$ experiences an event, its own contribution $\lambda_k$ is added or removed, contributing at most $B_\theta$.
    \item With weight $w_k = e^{\vec\theta_\ctrl^T \vec z_{ik}}$, patient $k$ belongs to the risk set of \emph{every} event with $\vec t_{ij} \le \vec t_{ik}$, adding $w_k$ to $A_j$ (if untreated) or $B_j$ (if treated), thereby changing $\varphi(\pi_j)$. Since $\varphi$ obeys $|\varphi'(\pi)| \le e^{\theta_1}-1 \le e^{B_\theta}-1$, and removing $k$ shifts the treated fraction by $$|\pi_j - \pi_j^{-k}| = \frac {w_k A_j} {(A_j{+}B_j)(A_j{+}B_j-w_k)} \le \frac {w_k} {A_j{+}B_j-w_k} \le 1,$$, where $\pi_j^{-k}$ is the treated fraction of the risk set without $k$. This yields
\begin{equation*}
    \big| \Lambda_i(\vec s_i) - \Lambda_i(\vec s_i \setminus k) \big| \;\le\; B_\theta \;+\; \big(e^{B_\theta}-1\big)\, \Xi_{ik},
    \label{eq:cox_influence}
\end{equation*}

with

$$\Xi_{ik} = \sum_{\substack{j:\,\delta_{ij}=1,  k \in R(j)}} \frac{w_k}{A_j + B_j - w_k} \le n_i.$$

Therefore, the global sensitivity satisfies 

$$\Delta_{i, \Theta} = B_\theta + (e^{B_\theta} - 1) n_i.$$
\end{enumerate}

\noindent\textbf{A Practical Mechanism.} In practice,  for a clipping threshold $C > 0$, we release $\Lambda_i^{\mathrm{priv}} = \Lambda_i^{\mathrm{clip}} + \mathrm{Lap}\big(2C K/\eps\big)$, where $\Lambda_i^{\mathrm{clip}}$ is the magnitude-clipped statistic:

$$\Lambda_i^{\mathrm{clip}}(\vec s_i) = \mathrm{sign}\left ( \Lambda_i(\vec s_i) \right ) \min (|\Lambda_i(\vec s_i)|, C).$$Since $\Lambda_i^{\mathrm{clip}}$ takes values in $[-C, C]$, its $\ell_1$-sensitivity is at most $2C$, so the Laplace mechanism at scale $2C K/\eps$ is $\eps$-DP over the $K$ rounds, and the subsequent belief exchange is also protected by post-processing immunity.

\subsection{Additional Experimental Results with the Cancer Dataset}\label{sec:app:additionalexperimentsCANCER}

Mirroring the adjusted analysis of \cref{sec:sim}, we replicate the pipeline on the advanced-cancer cohort of \citet{samstein2019tumor}: $N = 1{,}631$ of the $1{,}662$ patients treated with immune checkpoint inhibitors (excluding 31 with zero recorded survival time), with $810$ deaths. The exposure is a binary indicator of high tumor mutational burden (TMB in the top quintile, the standard immunotherapy cutoff), and we adjust for age, sex, drug class (PD-1/PD-L1, CTLA-4, or combination), sequencing panel, and cancer type. Because TMB is measured on panel-specific assays and its distribution varies significantly across tumor types, the multicenter oncology study would have to account for them.

\Cref{fig:tmb_adjusted} summarizes the results. High TMB is strongly protective: the centralized adjusted analysis estimates a log hazard ratio of $-0.61$ ($95\%$ CI $[-0.79, -0.43]$, $P < 10^{-10}$), whereas a single hospital holding one fifth of the cohort recovers a consistent but far less certain estimate ($-0.48$, $95\%$ CI $[-0.88, -0.08]$). The distributed differentially private belief exchange reproduces the centralized conclusion, with the privacy-utility gap closing as $\eps$ grows and the geometric-mean rule again converging fastest.

\begin{figure}[t]
    \centering
    \includegraphics[width=0.8\linewidth]{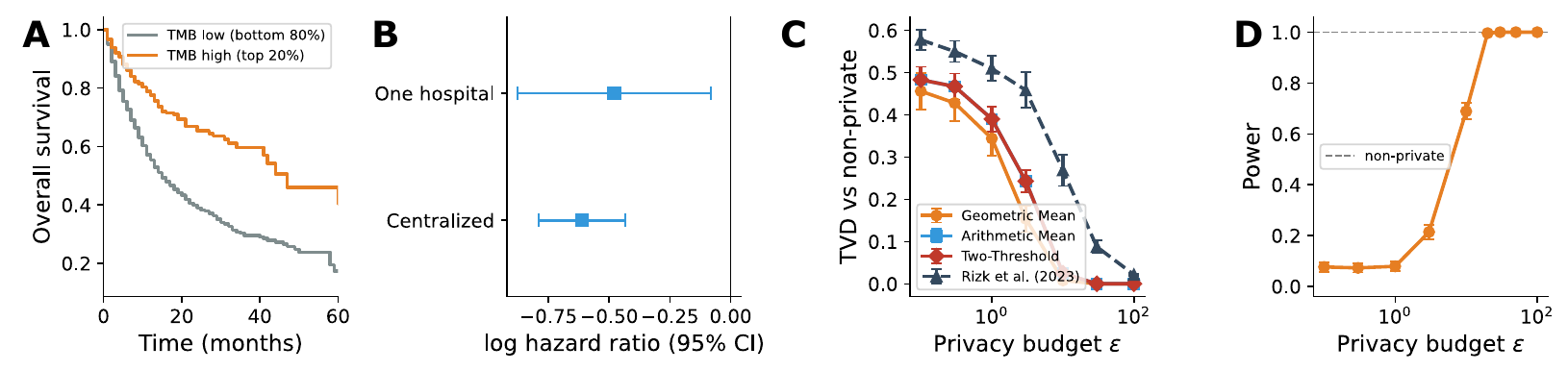}
    \caption{\textbf{Differentially private distributed survival analysis on the advanced-cancer cohort} of \citet{samstein2019tumor}, testing the effect of high tumor mutational burden (top quintile) {adjusted} for age, sex, drug class, sequencing panel, and cancer type, and $n = 5$ fully connected centers. (A) Kaplan-Meier survival curves for the TMB-high and TMB-low groups. (B) TMB-high log hazard ratio with $95\%$ confidence intervals, fitted centrally and at a single hospital. (C) Total variation distance between the private and non-private beliefs versus $\eps$, together with the differentially private first-order method of \citet{rizk2023enforcing} (D) Statistical power of the GM algorithm against $\eps$ at a significance level $\alpha = 0.05$. The dashed line shows the non-private power. Because this cohort is larger and the TMB effect stronger than the ddI effect, the federated test reaches full power by $\eps \approx 5$, well below the budget required on ACTG~175 (\cref{fig:clinical_trials}).}
    \label{fig:tmb_adjusted}
\end{figure}

% \subsection{Toy Example: MLE and OL for a Single Treatment}\label{sec:app:sim:simpleHT} 

% This toy example revisits the single-treatment ACTG analysis of \cref{sec:federation} in detail, with $n = 5$ hospitals, $\Theta = \{ 0, -\log 2 \}$ (the alternative $-\log 2$ posits that patients are half as likely to die under treatment), $\eps = 1$, and $\alpha = 1 - \beta = 0.05$. \cref{subfig:am_gm_algorithm,subfig:two_threshold} show that the AM/GM and two-threshold (single-threshold) estimators both recover the true maximizer $\Theta^\star = \{ -\log 2 \}$ (since $\Lambda(-\log 2) > \Lambda(0)$), matching the centralized baseline in which all centers share their data without privacy. \cref{subfig:online_learning} repeats the analysis online, with centers exchanging beliefs every 10 days; the time-averaged log-belief ratio tracks the non-DP limiting value of \citet{rahimian2016distributed}, and the agents again identify the true state $\theta^\circ = -\log 2$.

\subsection{Time and Space Complexity}\label{sec:app:timespacecomplexity}

We first start by analyzing the time complexity of our method. Specifically, in the distributed MLE setting, if agent $i$ has $n_i$ data points and model $\ell_i$ which can be computed in $\cT_i(n_i)$ for one value of $\theta \in \Theta$, then the distributed MLE is parallelizable between the agent and the total time complexity per agent is $\cO \left ( |\Theta| K \left ( \cT_i(n_i) + T \deg_{\cG} (i) \right ) \right )$ which includes an initialization cost of $|\Theta| K \cT_i(n_i)$ to calculate noisy likelihoods. For example, in the case of the distributed proportional hazards model, the likelihoods can be computed in $\cT_i(n_i) = \cO \left ( n_i \log n_i \right )$. Then, there is a communication cost, which consists of the number of belief exchanges dictated by the communication complexity $KT$, and the cost of propagation for each iteration, which is $\cO(|\Theta| K \deg_{\cG} (i))$ for all states. Similarly, in the OL regime, the total time complexity per agent is $\cO \left ( |\Theta| \sum_{t = 1}^T  \left (\cT_i(n_{i, t}) + \deg_{\cG} (i) \right ) \right )$. Finally, in terms of space complexity, each agent can simply maintain their belief in each state, giving a space complexity $\cO (K|\Theta|)$ during communication. 

\subsection{Runtime Comparison with Homomorphic Encryption Methods} \label{sec:app:runtime_comparison}

To further test the practical applicability of our method, we compare the runtime of our method with existing homomorphic encryption methods (HE) and, in particular, the FAMHE method introduced in \citet{froelicher2021truly} as the number of providers ($n$) grows and the number of data grows. Both our method and FAMHE are capable of performing survival analysis in a distributed regime; however, they apply different privacy protections and cannot be compared prima facie. Generally, HE methods have the strongest possible privacy. However, these protections fall short in scalability considerations, for which DP with a small $\eps$ is a better alternative in terms of efficiency. Furthermore, the most recent HE methods designed for multicenter trials do not have open-source implementations, making direct comparison difficult (see, e.g. \citet{froelicher2021truly,geva2023collaborative}).

\begin{figure}[t]
    \centering
    \subfigure[Cancer tumor mutational load study]{\includegraphics[width=0.75\linewidth]{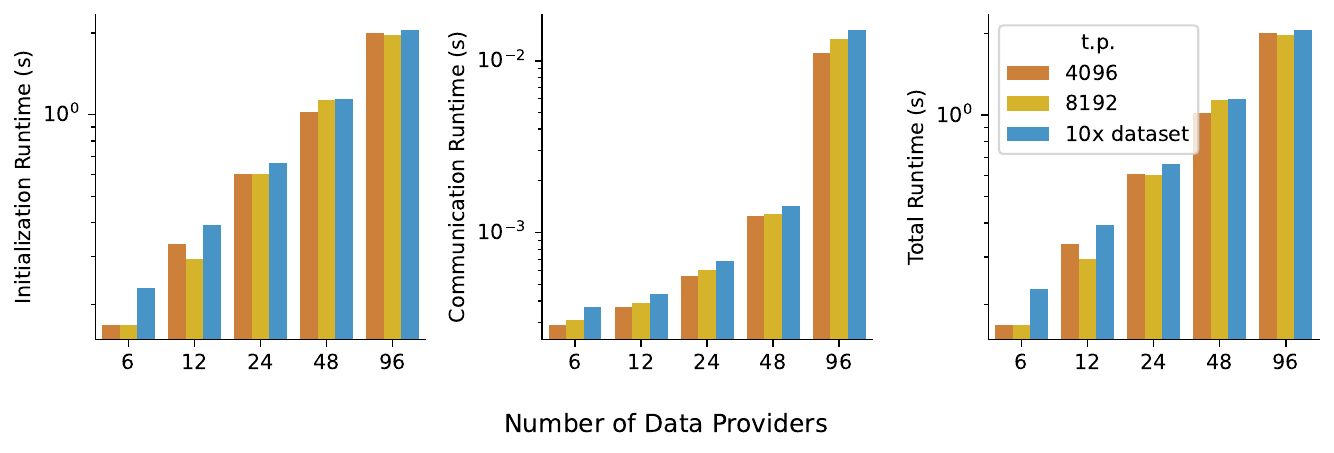}} \subfigure[ACTG study 175]{\includegraphics[width=0.75\linewidth]{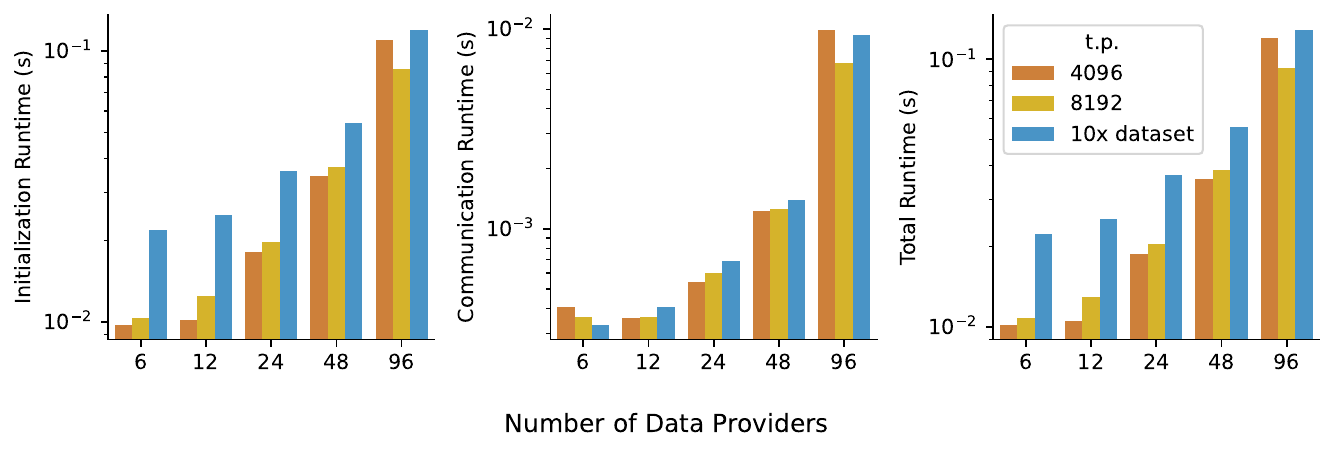}}
    \caption{Runtime of distributed MLE algorithm for (a) the study by \citet{samstein2019tumor} on the top and (b) the ACTG study on the bottom, for varying values of $n$ used in measuring the performance of the FAMHE method \citet{froelicher2021truly}. The privacy budget is set to $\eps = 1$ (tight privacy), and the error rates are set to $\alpha = 1 - \beta = 0.05$, and the thresholds are set to $\varrho^\AM = \varrho^\GM = 0.1$. In accordance with \citet{froelicher2021truly}, we have constructed 3 datasets: a dataset consisting of 4096 timepoints (t.p.) via resampling the original data with replacement, a dataset of 8192 t.p. via resampling the original data with replacement, and a dataset which consists of 10 times the original data.}
    \label{fig:runtime}
\end{figure}

First, our aim is to compare the two methods in the same dataset and experiment, which corresponds to distributed survival analysis with cancer data introduced in \citet{samstein2019tumor}. Specifically, \citet{samstein2019tumor} finds that tumor mutational burden (TMB) is a significant predictor of survival outcomes in patients with metastatic cancer undergoing immune checkpoint inhibitor (ICI) therapy. Analyzing data from 1,662 advanced cancer patients, the authors demonstrated that higher TMB levels are associated with better survival prospects.

\citet{froelicher2021truly} use the data from \citet{samstein2019tumor} and perform survival analysis (see Figure 2 in their paper) by splitting the data equally among $n$ providers. Then, the authors report the runtime as a function of $n$ for three datasets: with 4096 time points (t.p.), 8192 t.p., and a dataset that consists of 10 times the original data, and report the total runtime in Figure 2(c). To construct the fairest possible comparison, we consider the same dataset and the distributed proportional hazards model with TMB as a covariate and two hypotheses, null ($\theta = 0$) and a composite alternative ($\theta \neq 0$), similarly to our initial example for ACTG (see \cref{sec:federation}). The first two datasets (4096 and 8192 t.p.) are constructed by resampling the original data with replacement. We choose $\eps$ to be 0.1, which corresponds to a very strong privacy regime, significantly less than the privacy budgets used in the 2020 US Census \citep{censusDP,garfinkel2022differential} and other healthcare contexts \citep{ficek2021differential,dwork2019differential}.  Finally, we chose the error bounds to be no more than $0.1$. In \cref{fig:runtime}, we report the per-agent cost of inference for $n \in \{ 6, 12, 24, 48, 96 \}$ in a fully connected topology, analogous to Figure 2(c) of \citet{froelicher2021truly}. Our reported runtimes range from $\sim 10^{-2} \mathrm{s}$ to $\sim 1 \mathrm{s}$, which is a $10\times-1000\times$ improvement over the runtimes reported in \citet{froelicher2021truly}.

\subsection{Comparison with First-order Methods}\label{sec:app:firstordercomparison}

We compare with the first-order method of \citet{rizk2023enforcing} with graph-homomorphic noise. The distributed optimization problem we are solving is 
\begin{equation*}
    \max_{\theta \in \Rbb} \frac 1 n \sum_{i \in [n]} 2 \left ( \log \ell_i(\vec s_i | \theta) - \log \ell_i(\vec s_i | 0) \right ).
\end{equation*}

The distributed updates on the parameters with graph-homomorphic noise according to \citet{rizk2023enforcing} are equivalent to
{\begin{equation*}
    \vec \theta_{i, t} = \sum_{j \in [n]} a_{ij} \vec \theta_{j, t - 1} + \eta_{\mathrm{lr}} \mathrm{Clip} \left ( \frac {\mathrm d \log \ell_i(\vec s_i | \vec \theta_{i, t -1})} {\mathrm d \vec \theta_{i, t-1}}, 2 B_{\theta} B_x \right ) + \mathrm {Lap} \left ( \frac {2 B_xB_{\theta} T \eta_{\mathrm{lr}}} {\eps} \right ),
\end{equation*}}where $\eta_{\mathrm{lr}}$ is the learning rate, and the $\mathrm{Clip}$ operation clips the gradient to have $\ell_1$ norm at most $2B_\theta B_x$. To derive the Laplace noise, %note that the sensitivity of the clipped gradient is, at most, $2 \eta_{\mathrm{lr}} B_\theta B_x$.
note that the gradient decomposes over patients, $\frac{\mathrm d \log \ell_i(\vec s_i \mid \vec\theta)}{\mathrm d \vec\theta} = \sum_{j=1}^{n_i} \vec g_{ij}$ with $\vec g_{ij} = \frac{\mathrm d \log \ell(\vec s_{ij} \mid \vec\theta)}{\mathrm d \vec\theta}$, and each per-patient gradient satisfies $\|\vec g_{ij}\|_1 \le B_\theta B_x$ under $\|\vec x_{ij}\| \le B_x$ and $\|\vec\theta\| \le B_\theta$. Hence, two datasets differing by the addition or removal of one patient yield summed gradients that differ by at most $B_\theta B_x$ in $\ell_1$ norm, and because the $\ell_1$ clip is $2$-Lipschitz, their clipped gradients differ by at most $2 B_\theta B_x$. The $\ell_1$-sensitivity of the update term $\eta_{\mathrm{lr}} \mathrm{Clip}(\cdot)$ is therefore at most $2 \eta_{\mathrm{lr}} B_\theta B_x$. 
Additionally, since the per-agent budget is $\eps$, we need to scale the budget by $T$. In the simulations, we use $B_{\theta} = 1$, $\eta_{\mathrm{lr}} = 0.001$, and in order to have a fair comparison to our algorithm, we set $T$ to be equal to the number of iterations we run our algorithms. For large sample sizes the $P$ values are given as $P_i = 1 - F_{\chi_1^2}\left ( 2 \left [ \log \ell_i(\vec s_i | \vec \theta_{i, T}) - \log \ell_i(\vec s_i | 0) \right ] \right )$.

\clearpage

\clearpage

\section{Differentially Private Distributed Genetic Association Analysis}
\label{sec:app:gwas}

This section provides the supporting details for the region study of \cref{sec:gwas}, including the sensitivity analysis underlying the privacy calibration and additional experimental results.

\subsection{Setup and Notation}
\label{subsec:setup}

Let $\vec s_i = \{(\vec{g}_{ij}, y_{ij})\}_{j=1}^{n_i}$ denote the dataset
at center $i$, where $\vec{g}_{ij} \in \{0,1,2\}^M$ is the genotype vector and
$y_{ij} \in \Rbb$ is the phenotype of individual $j$, generated under the
linear model
\begin{equation}
    y_{ij} = \vec{g}^{\star T}_{ij}\vec{\theta} + \xi_{ij},
    \qquad
    \xi_{ij} \overset{\mathrm{iid}}{\sim} \cN(0, \sigma^2).
    \label{eq:linear_model}
\end{equation}
Let $\vec{G}^\star_i \in \Rbb^{n_i \times M}$ denote the standardized
genotype matrix with $(j,m)$ entry
$g^\star_{ij,m} = (g_{ij,m} - 2\hat{p}_m)/\sqrt{2\hat{p}_m(1-\hat{p}_m)}$,
and let $\vec{y}^c_i \in \Rbb^{n_i}$ denote the centered phenotype vector
with entries $y^c_{ij} = y_{ij} - \bar{y}_i$. The  weight matrix is
$\vec{W} = \mathrm{diag}(w_1,\ldots,w_M)$ with weights
$w_m = \mathrm{Beta}(\mathrm{MAF}_m; 1,25)^2$, optionally clipped at a maximum
value $w_{\max} \le 1$. Define the $n_i \times n_i$ kernel matrix:
\begin{equation*}
    \vec{K}_i = \vec{G}^\star_i \vec{W} \vec{G}^{\star T}_i,
    \end{equation*}
whose $(j,j')$ entry $K_{ijj'} = \vec{g}^{\star T}_{ij}\vec{W}\vec{g}^\star_{ij'}$
is the weighted genetic similarity between individuals $j$ and $j'$. We employ the SKAT statistic \citep{wu2011rare}: 

\begin{equation}
    {\Lambda}_i(\vec s_i | \theta_{\max})
    = \frac{\theta_{\max}}{n_i^2 \hat{\sigma}^2_i}
      \vec{y}^{cT}_i \vec{K}_i \vec{y}^c_i
    - \frac{\theta_{\max}^2}{2n_i\hat{\sigma}^2_i}
      \mathrm{tr} \left(\vec{W}\hat{\vec{V}}_i\right),
    \label{eq:llr}
\end{equation}
where $\hat{\vec{V}}_i = \vec{G}^{\star T}_i\vec{G}^\star_i/n_i$ is the
empirical LD matrix and $\hat{\sigma}^2_i = \|\vec{y}^c_i\|_2^2/n_i$. 

The global $\ell_1$-sensitivity is:
\begin{equation*}
    \Delta_{i, \Theta}
    = \max_{\vec s_i, \vec s_i' : \vec s_i \sim \vec s_i'}
      \left|\Lambda_i(\vec s_i | \theta_{\max})
           - \Lambda_i(\vec s_i' | \theta_{\max})\right|.
    \end{equation*}

for datasets $\vec s_i, \vec s_i'$ such that they differ in at most one patient.

\subsection{Bounding the Sensitivity of the SKAT Statistic}
\label{subsec:main_result}

\begin{theorem}
\label{thm:kernel_sensitivity}
Under model \eqref{eq:linear_model}, assume $|\theta_{\max}| \leq B_\theta$ and
let $\delta \in (0,1)$ be a failure probability. Then, with probability at least $1 - \delta$:

$$\Delta_{i, \Theta} \le 8 B_{\theta}M \frac {\sqrt{n_i} + \sqrt {M \log (3M / \delta)}} {\sqrt{n_i} - \sqrt {\log (3 / \delta)}} \frac{\log(6n_i/\delta)}{\sqrt{n_i}}.$$
\end{theorem}

\noindent \textbf{Proof.} Let $\delta \in (0, 1)$. We apply a union bound to guarantee that three separate random events behave well simultaneously, allocating a failure probability of $\delta/3$ to each.

First, because the phenotypes are drawn from a sub-Gaussian distribution, we can bound the maximum absolute value of the centered phenotype vector. With probability at least $1-\delta/3$, the maximum centered phenotype is bounded by $C_{\sigma}$, where:

$$C_{\sigma}^2 = 2\sigma^2\log(6n_i/\delta)$$

We denote this high-probability event as $\mathcal{E}_1 = \{\max_{j\in[n_{i}]}\|y_{ij}^{c}\|\le C_{\sigma}\}$.

Second, the SKAT statistic divides by the empirical variance $\hat{\sigma}_i^2$. We must ensure this denominator does not become arbitrarily small. Because $y_{ij}^c$ are sub-Gaussian, $\hat{\sigma}_i^2 = \frac{1}{n_i}\|\vec y_i^c\|_2^2$, the standard sub-exponential tail bounds yields that, with probability at least $1-\delta/3$:

$$\hat{\sigma}_i^2 \ge \sigma^2 (1 - \epsilon_1)$$

where $\epsilon_1 = \sqrt{\frac{\log(3/\delta)}{n_i}}$. We denote this event as $\mathcal{E}_2$.

Third, we apply the Matrix Bernstein inequality to bound the spectral gap between the empirical covariance matrix $\hat {\vec V}_i$ and the population correlation matrix $\vec R$. With probability at least $1-\delta/3$:

$$\|\hat {\vec V}_i - \vec R\|_2 \le \epsilon_2$$

where $\epsilon_2 = \sqrt{\frac{M \log(3M/\delta)}{n_i}}$. By Weyl's inequality, we can bound the maximum eigenvalue of the weighted matrix:

$$\lambda_{\max}(\vec W^{1/2}\hat {\vec V}_i \vec W^{1/2}) \le \lambda_{\max}(\vec W^{1/2}R \vec W^{1/2}) + \|\vec W\|_2 \epsilon_2$$

Because $\vec R$ is a correlation matrix (diagonals equal to 1), $\lambda_{\max}(\vec W^{1/2} \vec R \vec W^{1/2}) \le \mathrm{tr}(\vec W \vec R) = \mathrm{tr}(\vec W) \le M$. Furthermore, because $\vec W$ is diagonal with positive weights, $\|\vec W\|_2 \le \mathrm{tr}(\vec W) \le M$. Therefore:

$$\lambda_{\max}(\vec W^{1/2}\hat {\vec V}_i \vec W^{1/2}) \le M(1 + \epsilon_2)$$

We denote this event as $\mathcal{E}_3$. By the union bound, the intersection event $\mathcal{E} = \mathcal{E}_1 \cap \mathcal{E}_2 \cap \mathcal{E}_3$ holds with probability at least $1-\delta$.

Next, we evaluate the sensitivity between two adjacent datasets, $\vec s_i$ and $\vec s_i^\prime$, differing only by individual $k$. We decompose the centered phenotype vector as $\vec y_i^c = \vec y_{-k}^c + y_{ik}^ce_k$.

Expanding the quadratic form of the SKAT statistic for $\vec s_i$ and $\vec s_i^\prime$ and taking the difference $Q_i - Q_i^\prime$, the terms not involving $k$ cancel out, leaving a cross term $(A_1)$ and a squared term $(A_2)$:

$$Q_i - Q_i^\prime = 2(\vec K_i \vec y_{-k}^c)_k y_{ik}^c - 2(\vec K_i \vec y_{-k}^c)_k y_{ik}^{\prime c} + K_{ikk}(y_{ik}^c)^2 - K_{ikk}^\prime (y_{ik}^{\prime c})^2$$

Conditioned on $\mathcal{E}$, we apply the Cauchy-Schwarz inequality and spectral identities to bound~$(A_1)$:

$$\|(A_1)\| \le 2\|K_i\|_2 \|\vec y_{-k}^c\|_2 (\|y_{ik}^c\| + \|y_{ik}^{\prime c}\|) \le 4n_i^{3/2}\lambda_{\max}(W^{1/2}\hat {\vec V}_i W^{1/2})C_{\sigma}^2$$

Using the triangle inequality and bounding the diagonal entries by the trace $K_{ikk} \le \mathrm{tr}(\vec W)$, we bound~$(A_2)$:

$$\|(A_2)\| \le 2\mathrm{tr}(\vec W)C_{\sigma}^2$$

The trace term changes by at most $\frac{2 \mathrm{tr}(\vec W)}{n_i}$, contributing $\frac{B_{\theta}^2 \mathrm{tr}(\vec W)}{n_i^2 \hat{\sigma}_i^2}$ to the sensitivity. For $n_i$ large enough such that $C_{\sigma} \ge \sigma\sqrt{B_{\theta}}$, this is bounded by $\frac{B_{\theta}\mathrm{tr}(\vec W)C_{\sigma}^2}{n_i^2 \hat{\sigma}_i^2}$.

Finally, dividing the quadratic bounds by the denominator $n_i^2 \hat{\sigma}_i^2$ and combining with the trace term gives the intermediate bound:

$$\Delta_{i, \Theta} \le \frac{4B_{\theta}C_{\sigma}^2 \lambda_{\max}(\vec W^{1/2}\hat {\vec V}_i \vec W^{1/2})}{\sqrt{n_i}\hat{\sigma}_i^2} + \frac{4B_{\theta}\mathrm{tr}(\vec W)C_{\sigma}^2}{n_i^2 \hat{\sigma}_i^2}$$

We absorb the second term, as it decays at a faster rate ($1/n_i^2$) and is negligible relative to the $1/\sqrt{n_i}$ term. We now substitute our finite-sample bounds from events $\mathcal{E}_1$, $\mathcal{E}_2$, and $\mathcal{E}_3$:

$$\Delta_{i, \Theta} \le \frac{4B_{\theta}(2\sigma^2\log(6n_i/\delta)) \cdot M(1+\epsilon_2)}{\sqrt{n_i} \cdot \sigma^2(1-\epsilon_1)}$$

where

$$\frac{1+\epsilon_2}{1-\epsilon_1} = \frac {\sqrt {n_i} + \sqrt {M \log (3M / \delta)}} {\sqrt{n_i} - \sqrt {\log (3 / \delta)}}$$. Combining everything together, we get 

$$\Delta_{i, \Theta} \le 8 B_{\theta}M \frac {\sqrt{n_i} + \sqrt {M \log (3M / \delta)}} {\sqrt{n_i} - \sqrt {\log (3 / \delta)}} \frac{\log(6n_i/\delta)}{\sqrt{n_i}}.$$

\qed

\subsection{A Practical Mechanism}
\label{subsec:pure_eps_skat}

The sensitivity above had polynomial dependence on the number $M$ of SNPs. By applying clipping we can remove this dependence and achieve similar performance. Specifically, we privatize the SKAT statistic with the Laplace mechanism. The construction clips each individual's contribution and calibrates the noise accordingly.

Write $\vec u_i = \vec G_i^{\star T} \vec y_i^c \in \Rbb^M$ for the weighted score vector of center $i$, so that the quadratic form $Q_i$ is the weighted squared norm $Q_i = \vec y_i^{cT}\vec K_i \vec y_i^c = \|\vec W^{1/2}\vec u_i\|_2^2$ and the statistic $\Lambda_i$ of \eqref{eq:llr} is an affine function of $Q_i$. The vector $\vec u_i$ decomposes over individuals as $\vec u_i = \sum_{j=1}^{n_i} \vec v_j$ with $\vec v_j = y_{ij}^c \vec g_{ij}^\star$, each term depending only on individual $j$. We clip every contribution in the $L_1$ norm, $\tilde{\vec v}_j = \vec v_j \cdot \min(1, C_1 / \|\vec v_j\|_1)$ for a threshold $C_1 > 0$, form $\tilde{\vec u}_i = \sum_{j=1}^{n_i} \tilde{\vec v}_j$, and release $\vec u_i^{priv} = \tilde{\vec u}_i + \vec\eta$ with independent Laplace coordinates $\eta_m \sim \mathrm{Lap}(2 C_1 K / \eps)$.

\begin{theorem}
\label{thm:l1_laplace_skat}
Let $C_1 > 0$. Releasing $u_i^{priv} = \tilde u_i + \vec\eta$ with $\eta_m \sim \mathrm{Lap}(2 C_1 K/\eps)$ i.i.d.\ satisfies $\eps$-DP over the $K$ rounds. The SKAT statistic $Q_i^{priv} = \|\vec W^{1/2} u_i^{priv}\|_2^2$ and $\Lambda_i^{priv}$ are $\eps$-DP by post-processing.
\end{theorem}

\noindent \textbf{Proof.} Because $v_j$ depends only on individual $j$, replacing one individual leaves the other $n_i - 1$ clipped terms unchanged, so $\|\tilde u_i - \tilde u_i'\|_1 = \|\tilde v_k - \tilde v_{k'}\|_1 \le \|\tilde v_k\|_1 + \|\tilde v_{k'}\|_1 \le 2 C_1$ by the triangle inequality and the clip. The $\ell_1$-sensitivity of $\tilde u_i$ is therefore $\Delta_1 = 2 C_1$, and the Laplace mechanism with the stated scale is $\eps$-DP. Post-processing gives the claim. 

\qed

\subsection{Algorithm Performance for Different Configurations of Parameters}
\label{app:gwas_medium}

We perform sensitivity analysis for the genetic association analysis study described in \cref{sec:gwas} using the same parameters as in the main text. We keep some of these parameters fixed and vary others to show how the performance of our method changes.
 
\begin{figure}[!h]
    \centering
    \includegraphics[width=\linewidth]{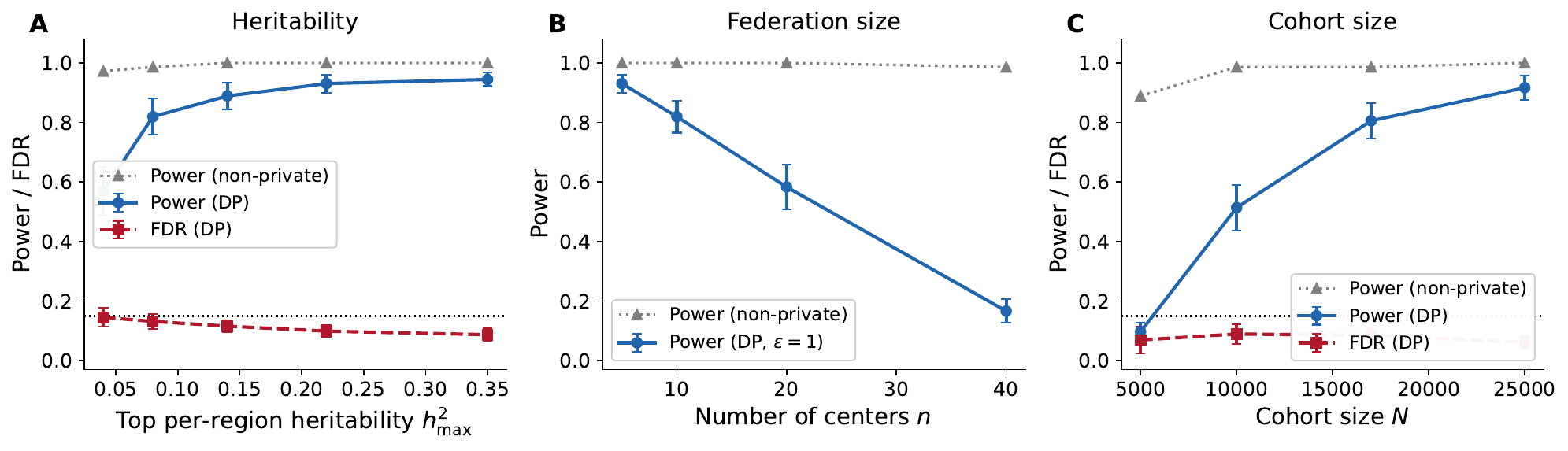}
    \caption{Sensitivity analysis of the genetic association analysis \textbf{(A)} Power and FDR versus per-region heritability. \textbf{(B)} Power versus the number of federating centers $n$ at fixed total cohort. \textbf{(C)} Power and FDR versus cohort size $N$. When not varied, the parameters set as in \cref{sec:gwas}.}
    \label{fig:gwas_2}
\end{figure}

\cref{fig:gwas_2}A shows how power changes under varying heritability for each region, demonstrating that our algorithm achieves higher power with higher heritability, as expected. 

\cref{fig:gwas_2}B fixes the total cohort and varies $n \in \{5, 10, 20, 40\}$ centers. As $n$ grows, each center holds fewer individuals, and the mechanism injects Laplace noise at more sites, so power declines.

\cref{fig:gwas_2}C shows power and FDR as a function of the cohort size $N$ under a complete network. Power rises toward the non-private baseline as $N$ grows, and the FDR is controlled at the nominal level for adequately sized cohorts.

\subsection{Additional Privacy-utility Curves}

\cref{fig:gwas_hospitals_2} verifies that the privacy-utility comparison of \cref{fig:gwas_hospitals} is robust to the choice of network topology: the organizational-level advantage persists when the organizations are arranged in a star network, and when the hospital-level network is taken to be fully connected.

\begin{figure}[!h]
    \centering
    \includegraphics[width=0.75\linewidth]{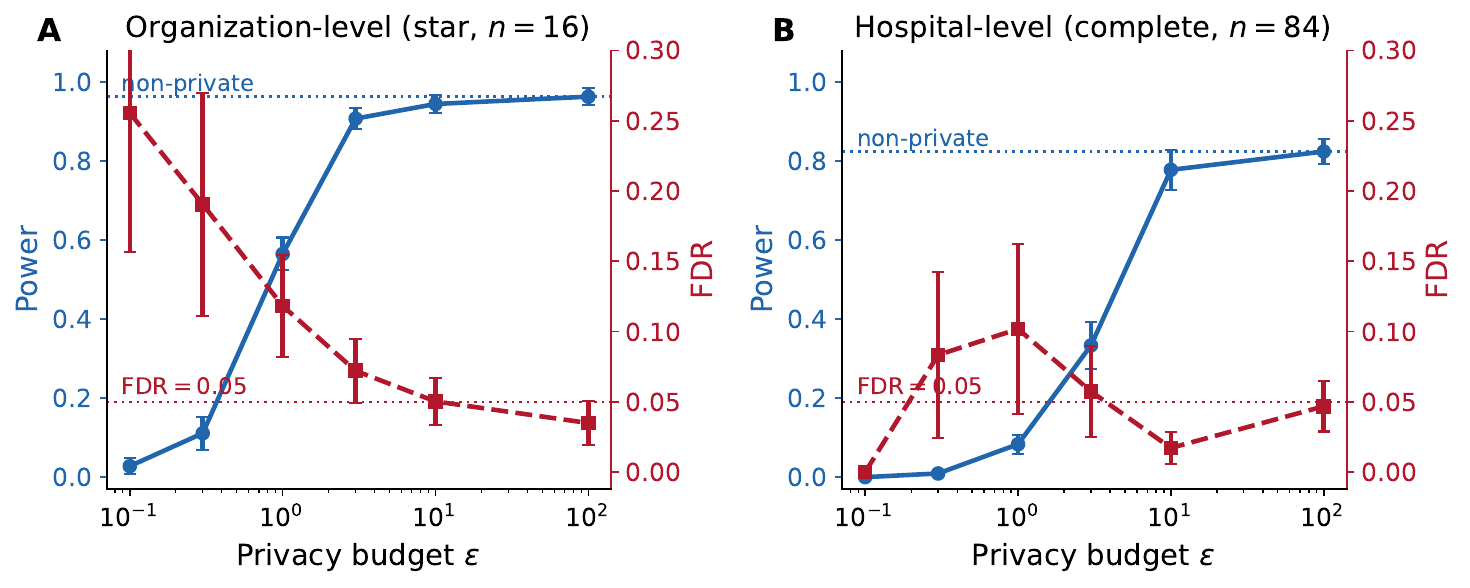}
    \caption{Robustness of the privacy utility curves of \cref{fig:gwas_hospitals} to the network topology. Panel (A) reproduces the same privacy-utility curve with \cref{fig:gwas_hospitals}A when the organizations are arranged in a star network with the central node being the biggest organization. Panel (B) reproduces the same privacy-utility curve with \cref{fig:gwas_hospitals}B when the hospital-level network is fully connected. We have used $T = 10, K = 2$ and 10 independent runs.}
    \label{fig:gwas_hospitals_2}
\end{figure}

\end{APPENDICES}

\end{document}